\documentclass{article}
\usepackage{log_2025}						% for camera-ready version

\usepackage{booktabs}						% professional-quality tables
\usepackage{multirow}						% tabular cells spanning multiple rows
\usepackage{amsfonts}						% blackboard math symbols
\usepackage{graphicx}						% figures
\usepackage{duckuments}						% sample images

\usepackage[numbers,compress,sort]{natbib}	% for numerical citations
\usepackage{url}            % simple URL typesetting
\usepackage{booktabs}       % professional-quality tables
\usepackage{amsfonts}       % blackboard math symbols
\usepackage{nicefrac}       % compact symbols for 1/2, etc.
\usepackage{microtype}      % microtypography
\usepackage{xcolor}         % colors

\usepackage{amsmath}
\usepackage{amssymb}
\usepackage{mathtools}
\usepackage{amsthm}
\usepackage{mdframed}
\usepackage{paracol}
\usepackage{adjustbox}

\usepackage{algorithm}
\usepackage{algorithmic}
\newtheorem{definition}{Definition}
\usepackage{graphicx}
\usepackage{subfigure}
\usepackage{tikz}
\usetikzlibrary{calc}
\usepackage{wrapfig}
\usepackage{resizegather} % Add to preamble
\definecolor{customorange}{HTML}{FF7F50}
\definecolor{custompurple}{HTML}{9370DB}

\newtheorem{theorem}{Theorem}
\newtheorem{proposition}[theorem]{Proposition}

\newcommand{\Loss}{\mathcal{L}} % General individual loss instance (e.g. L(f(x),y))
\newcommand{\cD}{\mathcal{D}} % True underlying data distribution
\newcommand{\cH}{\mathcal{H}}
\newcommand{\cX}{\mathcal{X}}
\newcommand{\cY}{\mathcal{Y}}
\newcommand{\lossell}{\ell}      % Specific metric for equivariance diff e.g. ||.||^2
\newcommand{\EmpiricalL}{\widehat{\mathcal{L}}} % Empirical losses
\newcommand{\E}{\mathbb{E}}
\newcommand{\methodabbr}{\textsc{REMUL}}

\newcommand{\hatLobj}{\EmpiricalL_{\mathrm{obj}}}
\newcommand{\hatLequi}{\EmpiricalL_{\mathrm{equi}}}
\newcommand{\hatLtot}{\EmpiricalL_{\mathrm{total}}}
\newcommand{\phig}{\phi(g)} % Action on input space
\newcommand{\rhog}{\rho(g)} % Action on output/label space
\newcommand{\hatLg}{\EmpiricalL_{\mathrm{g}}}
\newcommand{\fRemul}{f_{\alpha,\beta}} % REMUL empirical minimizer
\newcommand{\fObjEmp}{f^\star_{\mathrm{obj}}} % Objective-only empirical minimizer
\title[Relaxed Equivariance via Multitask Learning]{Relaxed Equivariance via Multitask Learning}

\author[Elhag et al.]{%
\hspace{9ex}Ahmed A. Elhag \\
\hspace{9ex}University of Oxford \\
\hspace{9ex}\email{ahmed.elhag@cs.ox.ac.uk}
\And
T. Konstantin Rusch\\
ELLIS Institute Tübingen \&\\
Max Planck Institute for Intelligent Systems \&\\
Tübingen AI Center\\
\email{tkrusch@tue.ellis.eu}
\AND
Francesco Di Giovanni\\
Valence Labs\\
\email{francesco.di.giovanni@valencelabs.com}
\And
\hspace{-8.4ex}Michael Bronstein\\
\hspace{-5ex}University of Oxford / AITHYRA\\
\hspace{-5ex}\email{michael.bronstein@cs.ox.ac.uk}
}

\begin{document}

\maketitle

\begin{abstract}
Incorporating equivariance as an inductive bias into deep learning architectures to take advantage of the data symmetry has been successful in multiple applications, such as chemistry and dynamical systems. 
In particular, roto-translations are crucial for effectively modeling geometric graphs and molecules, where understanding the 3D structures enhances generalization.
However, strictly equivariant models often pose challenges due  to their higher computational complexity. 
In this paper, we introduce REMUL, a training procedure that learns \emph{approximate} equivariance for unconstrained networks via multitask learning. By formulating equivariance as a tunable objective alongside the primary task loss, REMUL offers a principled way to control the degree of approximate symmetry, relaxing the rigid constraints of traditional equivariant architectures. We show that unconstrained models (which do not build equivariance into the architecture) can learn approximate symmetries by minimizing an additional simple equivariance loss. 
This enables quantitative control over the trade-off between enforcing equivariance constraints and optimizing for task-specific performance.
Our method achieves competitive performance compared to equivariant baselines while being significantly faster (up to 10$\times$ at inference and 2.5$\times$ at training), offering a practical and adaptable approach to leveraging symmetry in unconstrained architectures.
% This allows modulating the level of approximate equivariance based on task requirements -- demanding full equivariance or allowing more flexibility. 
\end{abstract}

\section{Introduction}
\label{section: introduction}
Equivariant machine learning models have achieved notable success across various domains, such as computer vision \cite{weiler2018learningsteerablefiltersrotation, yu2022rotationallyequivariant3dobject}, dynamical systems \cite{ han2022equivariantgraphhierarchybasedneural, xu2024equivariantgraphneuraloperator}, chemistry \cite{satorras2022en, brandstetter2022geometric}, and structural biology \cite{Jumper2021}. For example, incorporating equivariance \textit{w.r.t.} translations and rotations ensures the correct handling of complex structures like graphs and molecules \cite{schütt2021equivariantmessagepassingprediction, bronstein2021geometricdeeplearninggrids, 
thölke2022torchmdnetequivarianttransformersneural,
liao2024equiformerv2improvedequivarianttransformer}.  Equivariant machine learning models benefit from this inductive bias by {\em explicitly} leveraging symmetries of the data during the architecture design. Typically, such architectures have highly constrained layers with restrictions on the form and action of weight matrices and nonlinear activations 
% \textcolor{red}{CITE NEQUIP, MACE} 
\cite{Batzner2022Equivariant, 
batatia2023macehigherorderequivariant}. This may come at the expense of higher computational cost, making it sometimes challenging to scale equivariant architectures, particularly those %that enhance their expressivity by 
relying on spherical harmonics and irreducible representations \cite{thomas2018tensorfieldnetworksrotation, fuchs2020se3transformers, 
liao2023equiformerequivariantgraphattention, luo2024enablingefficientequivariantoperations}. 
% Furthermore, it is not always clear how tractable these models are to train. 
On the other hand, equivariance constraints might limit the expressive power of the network, restricting its ability to act as a universal architecture \cite{pmlr-v202-joshi23a}.

%{\em constraining} 

Equivariant layers are not the only way to incorporate symmetries into deep neural networks. 
%Constraining the layers of an architecture to be equivariant is not the only approach for dealing with group actions. 
Several approaches have been proposed to either offload the equivariance restrictions to faster networks \cite{kaba2022equivariance, mondal2023equivariant, baker2024an, ma2024canonizationperspectiveinvariantequivariant, panigrahi2024improved} 
% (\textcolor{red}{cite works Equivariance with Learned Canonicalization Functions and references therein/follow-ups} 
or simplify the constraints by introducing averaging operations \cite{puny2022frameaveraginginvariantequivariant, pmlr-v202-duval23a, lin2024equivariance, huang2024proteinnucleicacidcomplexmodeling}.
% (\textcolor{red}{cite FAENet: Frame Averaging Equivariant GNN for Materials Modeling and references therein/followups}. 
% Nonetheless, these directions cannot seamlessly leverage {\em unconstrained} architectures that do not bake symmetries into their design by simply altering the training protocol. 
Nonetheless, while these approaches leverage unconstrained architectures, they often require additional networks or averaging techniques to achieve equivariance and may not rely solely on adjustments to the training protocol.
% Nonetheless, while these directions leverage unconstrained architectures, they often require additional networks or modifications to achieve equivariance, rather than relying solely on adjustments to the training protocol.
To this aim, a widely adopted strategy to replace `hard' equivariance (i.e., built into the architecture itself) with a `soft' one, is {\em data augmentation} \cite{Quiroga_2019, benton2020learning, 9710479, bai2021transformersrobustcnns, gerken2022, 
Iglesias_2023,
chatzipantazis2023learning,
Xu2023ImageAugmentation,
Yang2024DataAugmentation}, whereby the training protocol of an arbitrary (unconstrained) network is augmented by assigning the same label to group orbits (e.g., rotated and translated versions of the input). 
%inputs  belonging to the same orbit of the group action.
In fact, recent works 
% \citet{dosovitskiy2021an} 
have shown that unconstrained architectures may offer a valid alternative, provided that enough data
are available \cite{wang2024swallowingbitterpillsimplified, Abramson2024}.

Besides the challenges in computational cost and design, there are also tasks (especially in scientific applications of ML) that do not exhibit full equivariance, such as dynamical phase transitions \cite{Baek_2017, Weidinger_2017}, 
polar fluids \cite{Gibb_2024},
molecular nanocrystals \cite{Yannouleas_2000}, 
and cellular symmetry breaking \cite{Goehring2011Polarization, mietke2019minimalmodelcellularsymmetry}. For such tasks, fully-equivariant networks might be excessively constrained, which further motivates the design of a more flexible approach.

In this work, we present 
\textbf{REMUL}: \textbf{R}elaxed \textbf{E}quivariance via 
\textbf{Mu}ltitask \textbf{L}earning.
% \textbf{CEUM}: \textbf{C}ontrolled \textbf{E}quivariance for \textbf{U}nconstrained \textbf{M}odels.  
REMUL is a training procedure that aims to learn approximate equivariance during training for unconstrained networks using a multitask approach with adaptive weights. 
% We introduce a novel, simple loss function that enables these models to approximate symmetries, which can be optimized during training.
We conduct a comprehensive evaluation of unconstrained models trained with REMUL, comparing their performance and computational efficiency to equivariant models. 
We consider Transformers and Graph Neural Networks (GNNs) and their roto-translational (E(3))-equivariant versions as our main baselines. Our contributions are:
\begin{itemize}
    \item We formulate equivariance as a weighted multitask learning objective for unconstrained models, aiming to simultaneously learn the objective function and approximate the required equivariance associated with the data and the task.
    \item We demonstrate that by adjusting the weighting of the equivariance loss, we can modulate the extent to which a model exhibits equivariance, depending on the task's requirements. Specifically, tasks that demand full equivariance require a higher weight on the equivariance term, whereas tasks that require less strict equivariance can be managed with lower weights.
    \item Empirically, we show that Transformers and Graph Neural Networks trained with our multitask learning approach compete or outperform their equivariant counterparts.
    \item By leveraging the efficiency of Transformers, we achieve up to $10\times$ speed-up at inference and $2.5\times$ speed-up in training compared to equivariant baselines. This finding could provide motivations for the use of unconstrained models, which do not require equivariance in their design, potentially offering a more practical approach.
    \item We point out that the standard Transformer exhibits a more convex loss surface near the local minima compared to the Geometric Algebra Transformer \cite{brehmer2023geometric}, 
    % which is an E(3) equivariant architecture for 3D point clouds  (see \hyperref[fig:loss_surface]{Figure \ref*{fig:loss_surface}}). 
    which can indicate further evidence of the optimization difficulties of equivariant networks.
\end{itemize}
% \textcolor{red}{We need a name and acronym for the training procedure, considering that this is the main contribution of the work. We should mention the gradual penalty applied to equivariance/data augmentation as our contributions, cause to me this is the main factor differentiating us from standard data augmentation procedures. In general I would reduce emphasis on loss landscape and instead emphasize that our approach paves the way for leveraging advanced techniques for handling multiple losses in how we deal with equivariance and symmetries.}

\section{Background}
\label{section: background}
\subsection{Symmetry Groups and Equivariant Models}
\label{section: symmetry_groups_and_equivariant_models}
Symmetry groups, a fundamental concept in abstract algebra and geometry, are a mathematical description of the properties of an object remaining unchanged (invariant) under a set of transformations. 
Formally, a symmetry group \(G\) of a set \(X\) is a group of bijective functions from \(X\) to itself, where the group operation is function composition. 

Equivariant machine learning models are designed to preserve the symmetries associated with the data and the task. In geometric deep learning (GDL), the data is typically assumed to live on some geometric domain (e.g., a graph or a grid) that has an appropriate symmetry group (e.g., permutation or translation) associated with it. 
%
%
%inherent to their data domains. Specifically, a 
Equivariant models implement functions \( f: X \to Y \) from input domain \( X \) to output domain \( Y \) that ensure the actions of a symmetry group \( G \) on data from \( X \) 
 correspond systematically to its actions on 
 \( Y \), 
 %transformations induced by \( G \), as captured 
 through the respective group representations \( \phi \) and \( \rho \). Formally, we say that: 
\begin{definition}
A function \( f \) is {\em equivariant} w.r.t. the group \( G \) if for any transformation \( g \in G \) and any input \( x \in X \), 
\begin{equation}
\label{equation: equi_function}
    f(\phi(g)(x)) = \rho(g)(f(x))
\end{equation}
\end{definition}
The group representations \( \phi \) and \( \rho \) allow us to apply abstract objects (elements of the group \( G \)) on concrete input and output data, in the form of appropriately defined linear transformations. For example, if $G=S_n$ (a permutation group of $n$ elements, arising in learning on graphs with $n$ nodes), its action on $n$-dimensional vectors (e.g., graph node features or labels) can be represented as an $n\times n$ permutation matrix. 

%the input and output spaces, respectively, each mapping \( g \) to a linear transformation in these spaces.
A special case of equivariance is obtained for a trivial output representation $\rho=\mathrm{id}$:
\begin{definition}
A function \(f\) is {\em invariant} w.r.t. the group \(G\) if for all \(g \in G\), \(x \in X\):$
\label{equation: invar_function}
    f(\phi(g)(x)) = f(x).
$
\end{definition}
\subsection{Equivariance as a Constrained Optimization Problem}
\label{section: equivariant_models_as_constraine_optimization_problem}
Consider a class of parametric functions \( f_{\theta} \) from a hypothesis space $\mathcal{H}$, typically implemented as neural networks, whose parameters \( \theta \) are estimated via a general training objective based on data pairs $(x,y)\sim q$:
\begin{equation}
\begin{aligned}\label{eq: minimization_problem}
& \underset{\theta}{\text{minimize}} & & \mathbb{E}_{(x,y)\sim q}\left[\mathcal{L}(f_{\theta}(x), y)\right]
\end{aligned}
\end{equation}
Here, \( \mathcal{L} \) represents the loss function that quantifies the discrepancy between the model's predictions \( f_{\theta}(x) \) and the true labels \( y \). The class of models is considered equivariant with respect to a group \( G \) if it satisfies the constraint in Eq. \ref{equation: equi_function}
% \hyperref[equation: equi_function]{Equation \ref*{equation: equi_function}}
for any input $x \in X$ and for any action $g \in G$.

Equivariance is typically achieved \emph{by design}, by imposing constraints on the form of $f_\theta$. Since $f_\theta$ is usually composed of multiple layers, ensuring equivariance implies restrictions on the operations performed in each layer, a canonical example being message-passing graph neural networks whose local aggregations need to be permutation-equivariant to respect the overall invariance to the action of the symmetric group $S_n$. As such, finding an equivariant solution to the minimization problem in Eq. \ref{eq: minimization_problem}
% \hyperref[eq: minimization_problem]{Equation \ref*{eq: minimization_problem}}
corresponds to solving the following constrained optimization: 
\begin{equation}
\label{equation:equivariant_constraint}
\begin{aligned}
& \underset{\theta}{\text{minimize}} & & \mathbb{E}_{(x,y)\sim q}\left[\mathcal{L}(f_{\theta}(x), y)\right] \\
& \text{subject to} & & f_{\theta}(\phi(g)(x)) = \rho(g) f_{\theta}(x), \, \forall g \in G, \, \forall x \in X
\end{aligned}
\end{equation}
In general, such optimization is challenging, leading to complex design choices to enforce equivariance that could ultimately restrict the class of minimizers and make the training harder. Additionally, for relevant tasks, the optimal solution only needs to be approximate equivariant
\cite{wang2022approximatelyequivariantnetworksimperfectly, 
petrache2023approximationgeneralizationtradeoffsapproximategroup, 
kufel2024approximatelysymmetricneuralnetworksquantum,
ashman2024approximatelyequivariantneuralprocesses}
% (\textcolor{red}{add citations}) 
meaning that the extent to which a model needs to exhibit equivariance can vary significantly based on the specific characteristics of the data and the requirements of the downstream application. In light of these reasons, we require a flexible approach to incorporating equivariance into the learning process. To address this, we propose REMUL, a training procedure that replaces the hard optimization problem with a soft constraint, by % the optimzation problem described in Equation \ref{equation:equivariant_constraint} by 
using a multitask learning approach with adaptive weights.

\section{REMUL Training Procedure}
\label{section: symmetries_through_loss_landscape}
\subsection{Equivariance as a New Learning Objective}
\label{section: equivariance_as_a_learning_objective}
Our main idea is to formulate equivariance as a multitask learning problem for an unconstrained model. We achieve that by {\em relaxing} the optimization problem in Eq. \ref{equation:equivariant_constraint}.
% \hyperref[equation:equivariant_constraint]{Equation \ref*{equation:equivariant_constraint}}.
% Equation \ref{equation:equivariant_constraint}
Namely, once we introduce a functional $\mathcal{F}_{\mathcal{X},G}$ that measures the equivariance of a candidate function $f_\theta$, we replace the constrained variational problem in 
% \hyperref[equation:equivariant_constraint]{Equation \ref*{equation:equivariant_constraint}}
Eq. \ref{equation:equivariant_constraint} with
\begin{equation}\label{eq: relaxed_optimization}
    \underset{\theta}{\text{minimize}}  \,\, \mathbb{E}_{(x,y)\sim q}\left[\alpha\mathcal{L}(f_{\theta}(x), y) + \beta \mathcal{F}_{\mathcal{X},G}(f_\theta(x),y)\right],
\end{equation}
where $\alpha,\beta > 0$. This decomposition allows for tailored learning dynamics where the supervised loss specifically addresses the information from the dataset without constraining the solution $f_\theta$, while the equivariance penalty $\mathcal{F}$ smoothly enforces symmetry preservation. %By adjusting the coefficients $\alpha,\beta$, such objective %equivariant component learns transformation-specific features while the non-equivariant component focuses on capturing the objective function in the given data. Such a configuration 
%offers a balanced approach, that optimizes the model parameters $\theta$ while maintaining the {\em level} of equivariance as dictated by the data and task at hand. 

%weights the importance of the constraint---in practice, we control both the supervised signal and the equivariance loss as discussed below. 
\textbf{Empirical Formulation.}  Let $\cD_n=\{(x_i,y_i)\}_{i=1}^n$ be a training sample of size $n$ drawn i.i.d.\ from an underlying distribution $P_{XY}$ on $\cX\times\cY$. In conventional supervised settings, we define the empirical version of our optimization problem as:
% We note that in conventional supervised settings, one has access to 
% the primary goal is to optimize a predictive function \( f^* \), where the optimization criterion is typically the minimization of a cumulative loss function over 
% a dataset \( \mathcal{X} = \{x_1, x_2, \ldots, x_n\} \) with corresponding labels \( \mathcal{Y} = \{y_1, y_2, \ldots, y_n\} \).  We can then introduce
% \begin{equation}
% \label{equation: supervised_learning}
%     \mathcal{L}_{\text{obj}}(f_{\theta}, \mathcal{X}, \mathcal{Y}) = \sum_{i=1}^n \mathcal{L}(f_{\theta}(x_i), y_i),
% \end{equation}
% and formulate the optimization as:
% Following our definition in \hyperref[eq:function_decompostion]{Equation \ref*{eq:function_decompostion}}, we propose the integration of an \textit{equivariance loss} into the training protocol. % cess of a function \( f_{\theta} \). 
% This approach conceptualizes learning equivariance from the the data leveraging a multitask framework, without requiring equivariance to be explicitly accounted in the architecture design. 
%The augmented loss function is formulated as:
\begin{equation}
    \label{eq_total_loss}
  \mathcal{L}_{\text{total}}(f_{\theta}, \mathcal{X}, \mathcal{Y}, G) = \alpha \hatLobj(f_{\theta}, \mathcal{X}, \mathcal{Y}) + \beta \hatLequi(f_{\theta}, \mathcal{X}, \mathcal{Y}, G),
\end{equation}
% \begin{equation}
%     \label{eq_total_loss}
%     \mathcal{L}_{\text{total}}(f_{\theta}, \mathcal{X}, \mathcal{Y}, G) = \alpha \mathcal{L}_{\text{obj}}(f_{\theta}, \mathcal{X}, \mathcal{Y}) + \beta \sum_{i=1}^{n} \mathcal{L}_{\text{equi}}(f_{\theta}, x_i, y_i, g_i(t)),
% \end{equation}
where $\hatLobj(f_{\theta}, \mathcal{X}, \mathcal{Y})$ is the empirical objective loss given by $\hatLobj(f_{\theta}, \mathcal{X}, \mathcal{Y})= \tfrac1n \sum_{i=1}^n \Loss(f_\theta(x_i),y_i)$, and \( \hatLequi(f_{\theta}, \mathcal{X}, \mathcal{Y}, G) \) represents our \textit{augmented equivariance loss}, specifically designed to enforce the model's adherence to the symmetry action of the group $G$. For a finite number of training samples $n$, we propose an empirical equivariant loss \( \hatLequi \) of the form:
% \[
%   \hatLequi(f_\theta) = \frac1n\sum_{i=1}^n \E_{g\sim\mu_G}\Bigl[ \lossell(f_\theta(\phig x_i),\;\rhog y_i) \Bigr],
% \]
\begin{equation}
 \label{eq:equivariant_loss}
    \hatLequi(f_{\theta}, \mathcal{X}, \mathcal{Y}, G)  = \frac1n\sum_{i=1}^n \E_{g\sim G}\Bigl[ \lossell(f_\theta(\phig x_i),\;\rhog y_i) \Bigr]
\end{equation}

here \( \ell \) is a metric function, typically an \(L_1\) or \(L_2\) norm, that quantifies the discrepancy between \( f(\phi(g)(x_i)) \) and \( \rho(g)(y_i) \), with $g \in G$ randomly-selected group elements drawn from a uniform distribution for each sample. 
% In our implementation, we enhance computational efficiency by selecting a single group element per sample at each training step and found it is effective to give stable results. We also show how performance goes as we increase the number of samples.
In our implementation, we enhance computational efficiency by selecting a single group element per sample at each training step, which we found produces effective results. In addition, we show how the performance varies as we increase the number of group samples.
% In fact, in our implementation, we select as single group-element per sample and change the group elements being sampled in each training ste, to be effective and computationally efficient. We also shwo how the performance goes when increasing the number of samples.
% We also note that using the true label $y$ (instead of $f(x)$) in Eq.~\ref{eq:equivariant_loss} helps in reducing the drift of the function output from the correct output. 

\textbf{Characterizing the \methodabbr{} Trade-off. } % Or a similar title
\label{subsec:remul_tradeoff_theory}
While \methodabbr{} is presented as a practical training procedure, it can be theoretically understood as a regularized optimization problem. 
The parameters \( \alpha \) and \( \beta \) defined in Eq. \ref{eq_total_loss}
% \hyperref[eq_total_loss]{Equation \ref*{eq_total_loss}}
are weighting factors that balance the traditional objective loss with the equivariance loss, enabling practitioners to tailor the training process according to specific requirements of symmetry and generalization. 
% More specifically, a large value of \( \beta \) indicates a more equivariant function while the smaller value of   \( \beta \) indicates a less equivariant function. These parameters allow us to control the trade-off between model generalization and equivariance,  based on the specific requirements of the task.
The following proposition characterizes the properties of the empirical minimizer $f_{\alpha,\beta}$ and the underlying trade-offs between task performance and equivariance. The proof is provided in Appendix \ref{app:proofs}.
\begin{proposition}
\label{prop:empirical_tradeoff} % Changed label for consistency
Let $f_{\alpha,\beta} \in \arg\min_{f\in\cH}\hatLtot(f; \alpha, \beta)$ be an empirical minimizer of the \methodabbr{} objective, and let $f^\star_{\mathrm{obj}} \in \arg\min_{f\in\cH} \hatLobj(f)$ be an empirical minimizer for the objective loss alone.
Then:
\begin{enumerate}
    \item[(a)] $f_{\alpha,\beta}$ is Pareto optimal for the bi-objective problem $(\min \hatLobj(f), \min \hatLequi(f))$.
    \item[(b)] The following trade-off inequality holds:
        \begin{equation}
          \hatLobj(f_{\alpha,\beta}) - \hatLobj(f^\star_{\mathrm{obj}}) \leq \frac{\beta}{\alpha} \left( \hatLequi(f^\star_{\mathrm{obj}}) - \hatLequi(f_{\alpha,\beta}) \right). \label{eq:main_tradeoff_obj_cost_prop11}
        \end{equation}
    % \item[(c)] Symmetrically, for any $\tilde{f} \in \cH$:
    %     \begin{equation}
    %     \hatLequi(f_{\alpha,\beta}) - \hatLequi(\tilde{f}) \leq \frac{\alpha}{\beta} \left( \hatLobj(\tilde{f}) - \hatLobj(f_{\alpha,\beta}) \right). \label{eq:main_tradeoff_equi_gain_prop11}
    %     \end{equation}
\end{enumerate}
\end{proposition}
\paragraph{Controlling Approximate Equivariance via $\beta/\alpha$. }
% Proposition~\ref{prop:empirical_tradeoff} provides key insights into the behavior of \methodabbr{}.
Eq.~\ref{eq:main_tradeoff_obj_cost_prop11} quantifies the empirical cost of enforcing equivariance, showing that any increase in primary task's loss beyond the unconstrained minimum $\hatLobj(f^\star_{\mathrm{obj}})$) is bounded by the product of relative weight $\beta/\alpha$ and the achieved reduction in the equivariance loss (from $\hatLequi(f^\star_{\mathrm{obj}})$ down to $\hatLequi(f_{\alpha,\beta})$). The ratio $\beta/\alpha$ serves as a lever to control the solution's properties: When $\beta/\alpha \to 0$, the objective prioritizes task performance, causing $f_{\alpha,\beta}$ to approximate $f^\star_{\mathrm{obj}}$ (potentially sacrificing equivariance if $f^\star_{\mathrm{obj}}$ lacks natural symmetry). In contrast, when $\beta/\alpha \to \infty$, the objective prioritizes equivariance, driving $\hatLequi(f_{\alpha,\beta})$ toward zero (at the cost of task performance). Finally, at intermediate $\beta/\alpha$ values, 
% the solution $f_{\alpha,\beta}$ represents a specific point on the Pareto frontier, enabling \methodabbr{} to achieve tunable approximate equivariance tailored to task requirements as we demonstrated in Section~\ref{sec: Experiments and Discussion}.
the solution $f_{\alpha,\beta}$ represents a specific balance on the empirical Pareto frontier. 
% \methodabbr{} thus allows learning a tunable degree of approximate equivariance, depending on the requirement of the tasks, as we demonstrated in Section~\ref{sec: Experiments and Discussion}.
\methodabbr{} thus allows learning a tunable degree of approximate equivariance, with larger $\beta$ producing a more equivariant function, while smaller $\beta$ produces a less equivariant function. This flexibility allows us to control the trade-off between model generalization \& equivariance based on the task's requirements, as we demonstrate empirically in Section~\ref{sec: Experiments and Discussion}. 
\subsection{Adapting Penalty Parameters during Training}
\label{section: penalty_settings}
For simultaneously learning the objective and equivariance losses, we consider two distinct approaches to regulate the penalty parameters $\alpha$ and $\beta$: {\em constant} penalty and {\em gradual} penalty.
The constant penalty assigns a fixed weight to each task’s loss throughout the training process. 
In contrast, the gradual penalty dynamically adjusts the weights of each task's loss during training. For gradual penalty, we use the GradNorm algorithm introduced by \cite{chen2018gradnormgradientnormalizationadaptive}, which is particularly suited for tasks that involve simultaneous optimization of multiple loss components, as it dynamically adjusts the weight of each loss during training.  It updates the weights of the loss components based on the magnitudes of their gradients, \textit{w.r.t} the last layer in the network, which is essential for the contribution of each loss. It also has a learning rate parameter \( \eta \), that fine-tunes the speed at
% \switchcolumn
\begin{algorithm}
\caption{GradNorm Algorithm (one step)}
\label{gradnorm}
\begin{algorithmic}[1]
\STATE \textbf{Input:} $\alpha$, $\beta$, $\eta$, $\gamma$, $\hatLobj$, $\hatLequi$, and $\mathcal{W}$ (the weights of the last layer in the network)

% \For{each training step}
    % \State $L_{\text{obj}}(t) = \sum_i \alpha L_{\text{obj}}(f(x_i), y_i)$
    % \State $L_{\text{equi}}(t) = \sum_i \beta L_{\text{equi}}(f, x_i, g_i)$
    % \State  $L(t) = L_{\text{obj}}(t) + L_{\text{equi}}(t)$
    % \State :
    % \State $G_{\text{obj}}^{(i)}(t) = \|\nabla_W L_{\text{obj}}(f, x_i, y_i)\|_2$
    % \State $G_{\text{equi}}^{(i)}(t) = \|\nabla_W L_{\text{equi}}(f, x_i, g_i)\|_2$
    \STATE $\mathcal{G}_{\text{obj}} = \|\nabla_{\mathcal{W}} \alpha \hatLobj\|_2$, $\tilde{\mathcal{L}}_{\text{obj}} = \hatLobj/ \hatLobj(0)$
    \STATE $\mathcal{G}_{\text{equi}} = \|\nabla_{\mathcal{W}} \beta \hatLequi\|_2$, $\tilde{\mathcal{L}}_{\text{equi}} = \hatLequi/ \hatLequi(0)$
    \STATE $\bar{\mathcal{G}} = \frac{\mathcal{G}_{\text{obj}} + \mathcal{G}_{\text{equi}}}{2}$, $r = \frac{\tilde{\mathcal{L}}_{\text{obj}} + \tilde{\mathcal{L}}_{\text{equi}}}{2}$
    \STATE  $r_{\alpha} = \frac{\tilde{\mathcal{L}}_{\text{obj}}}{r}$, $r_{\beta} = \frac{\tilde{\mathcal{L}}_{\text{equi}}}{r}$
    \STATE $ \hatLg = |\mathcal{G}_{\text{obj}} - \bar{\mathcal{G}} \times [r_{\alpha}]^\gamma| + |\mathcal{G}_{\text{equi}} - \bar{\mathcal{G}} \times [r_{\beta}]^\gamma|$
    % \State Update $w_i(t) \mapsto w_i(t+1)$ using $\nabla_{w_i} L_{\text{grad}}$
     \STATE $\alpha = \alpha - \eta \nabla_{\alpha} \hatLg$
    \STATE $\beta = \beta - \eta \nabla_{\beta} \hatLg$
    \STATE \textbf{Return:} $\alpha$, $\beta$
\end{algorithmic}
\end{algorithm}
% \end{paracol}
which the weights are updated, providing precise control over their convergence rates (see Algorithm \ref{gradnorm} for details).
\paragraph{Equivariance with Data Augmentation.}
\label{section: data_augmentation}
Standard data augmentation for enforcing equivariance typically involves augmenting the training data with pairs $(\phi(g)(x_i), \rho(g)(y_i))$ and training the model $f_\theta$ using only the original task loss $\mathcal{L}_{obj}$, i.e., minimizing $\sum_i \mathcal{L}(f_\theta(\phi(g)(x_i)), \rho(g)(y_i))$ over the augmented dataset. This implicitly encourages the network to learn symmetries by penalizing predictions on transformed data using the standard task objective. REMUL differs by introducing a separate, explicit equivariance loss term $\mathcal{L}_{\text{equi}}$ alongside the standard objective loss $\mathcal{L}_{\text{obj}}$  on the original data, as indicated in Eq.~\ref{eq_total_loss}.  The multitask framework with weights $\alpha, \beta$ allows \textit{explicit control} over the balance between fitting the original data and enforcing the equivariance constraint. 
% Data augmentation is a widely recognized technique that enhances the performance of machine learning models by including different transformations in the training process. It involved creating a transformed input and measuring the original loss between the model prediction and the transformed target.
% In contrast, our method utilizes an additional \textit{controlled} equivariance loss to incorporate symmetrical considerations simultaneously with the objective loss during training. In fact, traditional data augmentation techniques can be interpreted as special cases of Eq. \ref{eq_total_loss} 
% % \hyperref[eq_total_loss]{Equation \ref*{eq_total_loss}}
% where $\alpha = 0$ and $\beta = 1$.

\section{Quantifying Learned Equivariance}
\label{measure_equivariant}
Using group transformations to measure and assess the symmetries of ML models has been studied in several domains \citep{ 
lyle2020benefitsinvarianceneuralnetworks,
kvinge2022waysdeepneuralnetworks, 
moskalev2023genuineinvariancelearningweighttying,
gruver2024liederivativemeasuringlearned, speicher2024understandingroleinvariancetransfer}. Inspired by the idea of frame-averaging \citep{puny2022frameaveraginginvariantequivariant, pmlr-v202-duval23a, lin2024equivariance}, we introduce a metric to quantify the degree of equivariance exhibited by a function $f$, defined as:
\begin{equation}
\label{eq:equi_measure_1}
    E(f, G) = \frac{1}{|\mathcal{D}|} \sum_{x \in \mathcal{D}} \left\| \frac{1}{M} \sum_{i=1}^M \rho(g_i)(f(x)) - \frac{1}{M} \sum_{i=1}^M f(\phi(g_i)(x)) \right\|_2
\end{equation}
where $\| \cdot \|_2$ denotes an $L_2$ norm (for non-scalar function), and $M$ is a large number of samples from $G$. (Proof in Appendix~\ref{app:proofs}). This error indicates the average deviation of a function \( f \) from perfect equivariance across the data distribution \( \mathcal{D} \) (lower value means more equivariant function). 
% In practice, we use $M = 100$ samples from the group and noticed this was sufficient to obtain stable results.
We also compare to the standard measure that takes the average over the group of differences between  $f(\phi(g)(x))$ and $\rho(g)(f(x))$,
\begin{equation}
\label{eq:equi_measure_2}
    E'(f, G) = \frac{1}{|\mathcal{D}|} \sum_{x \in \mathcal{D}} \frac{1}{M} \sum_{i=1}^M \left\| f(\phi(g_i)(x)) - \rho(g_i)(f(x)) \right\|_2
\end{equation}
%\hyperref[eq:equi_measure_1]{Equation \ref*{eq:equi_measure_1}} \& \hyperref[eq:equi_measure_2]{Equation \ref*{eq:equi_measure_2}}
% Equations \ref{eq:equi_measure_1} \& \ref{eq:equi_measure_2}
%indicate a practical metric for evaluating how closely the function $f$ approximates perfect equivariance throughout a data distribution $D$ (which should be zero for a perfect equivariance function). 
We observed that both measures have very similar behavior in our experiments, where $E$ and $E'$ are near zero for equivariant models.
% We also demonstrate that increasing the value of $\beta$ in Eq. \ref{eq_total_loss} results in a less equivariant error for $E$ and $E'$.
Furthermore, as we discussed in Section~\ref{section: symmetries_through_loss_landscape},
we demonstrate empirically that increasing the \methodabbr{} penalty weight $\beta$ (Eq.~\ref{eq_total_loss}) results in a lower equivariant error for $E$ and $E'$. 

\section{Related Work}
\label{related work}
\textbf{Equivariant ML Models.}
In the vision domain, group convolutions have proven to be a powerful tool for handling rotation equivariance for images and enhanced model generalization \cite{cohen2016group, 
cohen2019gaugeequivariantconvolutionalnetworks, weiler2021generale2equivariantsteerablecnns, qiao2023scalerotationequivariantliegroupconvolution}. Similarly, the development of equivariant architectures with respect to roto-translations for geometric data has been an active area of research \cite{chen2021equivariantpointnetwork3d, satorras2022en, han2022equivariantgraphhierarchybasedneural, xu2024equivariantgraphneuraloperator}. Techniques that use spherical harmonics and irreducible representations have shown a large success in modeling graph-structured data, such as SE(3)-Transformers \cite{fuchs2020se3transformers}, Tensor Field Networks \cite{thomas2018tensorfieldnetworksrotation}, and DimeNet \cite{gasteiger2022directionalmessagepassingmolecular}. More recently, \cite{brehmer2023geometric} introduced an E(3) equivariant Transformer that employs geometric algebra for processing 3D point clouds.

\textbf{Data Augmentation and Unconstrained Models.}
Alternatively, integrating transformations through data augmentation is a widely used strategy across multiple vision tasks, enhancing performance in image classification \cite{perez2017effectivenessdataaugmentationimage, inoue2018dataaugmentationpairingsamples,rahat2024dataaugmentationimageclassification}, object detection \cite{zoph2019learningdataaugmentationstrategies, wang2019dataaugmentationobjectdetection,kisantal2019augmentationsmallobjectdetection}, and segmentation \cite{negassi2021smartsamplingaugmentoptimalefficientdata, chen2021diverse, yu2024diffusionbaseddataaugmentationnuclei}. For geometric data, \cite{hu2021forcenetgraphneuralnetwork} has adapted a Graph Neural Network architecture with data augmentation to process 3D molecular structures. In parallel, \cite{dosovitskiy2021an} introduced that Vision Transformers (ViTs) with a large amount of training data can achieve comparable performance with Convolutional Neural Networks (CNNs), obviating the need for explicit translation equivariance within the architecture. Recently, this has shown to be effective for handling geometric data \cite{wang2024swallowingbitterpillsimplified, Abramson2024}. 

\textbf{Learning Symmetries and Approximate Equivariance. } 
Several studies have shown that the layers of CNN architectures can be approximated for a soft constraint \cite{wang2022approximatelyequivariantnetworksimperfectly, 
vanderouderaa2022relaxingequivarianceconstraintsnonstationary, romero2022learning, 
veefkind2024probabilisticapproachlearningdegree,
wu2024sbdetsymmetrybreakingobjectdetector, mcneela2023almost}. Conversely, \cite{vanderouderaa2023learninglayerwiseequivariancesautomatically} extends the Bayesian model selection approach to learning symmetries in image datasets. \cite{pmlr-v151-yeh22b} introduced a parameter-sharing scheme to achieve permutations and shifts equivariances in Gaussian distributions.  
Recent works have relaxed the hard constrained models to a soft constraint by adding unconstrained layers in the architecture design \cite{NEURIPS2021_fc394e99, 
pertigkiozoglou2024improvingequivariantmodeltraining}, canonicalization network \cite{lawrence2024improving}, or explicit relaxation \cite{kaba2023symmetry}. Additionally, 
\cite{8898722} modified the loss of CNN for segmentation task. \cite{shakerinava2022structuring} introduced a method to learn equivariant representation using the group invariants, while \cite{bhardwaj2023steerableequivariantrepresentationlearning} defined a regularizer that injects the equivariance in the latent space of the network by explicitly modeling transformations with additional learnable maps. 
In contrast, several works have started from pre-trained models \cite{Basu_Sattigeri_Natesan, kim2023learning}.
Furthermore, 
% For geometric data, recent works have modified 
the EGNN framework \cite{satorras2022en} has been modified using an invariant function \cite{zheng2024relaxingcontinuousconstraintsequivariant} or adversarial training procedure \cite{yang2023generative}.
% However, in our work, we start completely from unconstrained models without assuming any equivariance over the space of functions in the architecture design. 
However, in our work, we start from completely \textit{unconstrained} models, without imposing any equivariance constraints on the space of functions within the architecture. 
Moreover, we did not assume a specific class of models or introduce additional parameters, which increases the applicability of our method to various domains and makes it computationally efficient.

\section{Experiments and Discussion}
\label{sec: Experiments and Discussion}
In this section, we aim to compare constrained equivariant models with unconstrained models trained with REMUL, our multitask approach.  We are targeting three main questions: Can unconstrained models learn the approximate equivariance, how does that affect the performance \& generalization, %trade-off, 
and what are their computational costs.
\begin{figure}[t!]
    \centering
    \subfigure{\label{fig:id_grad}
        \includegraphics[width=0.3\textwidth]{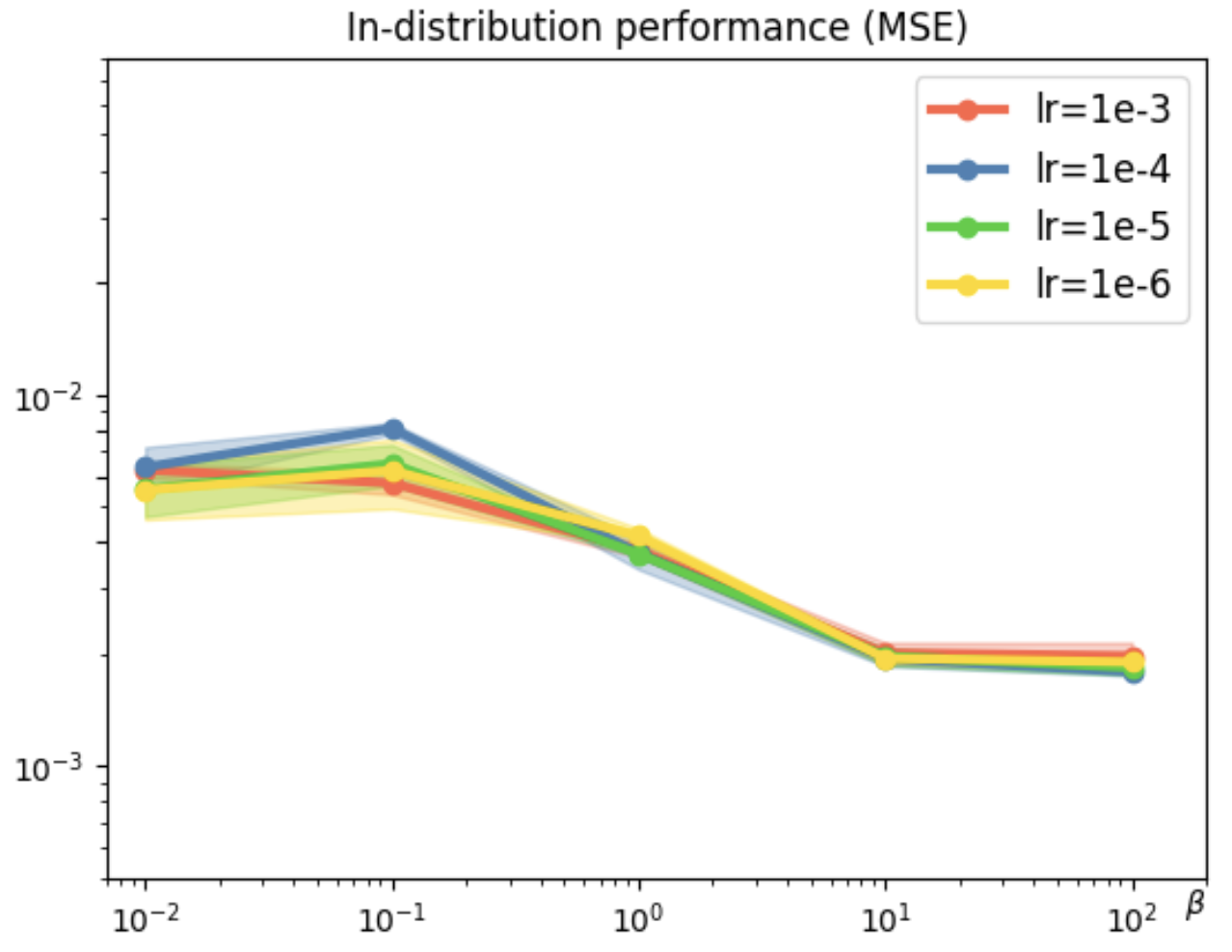}
    }
    % \hspace{-0.01cm} 
    \subfigure{\label{fig:id_constant}

    \begin{tikzpicture}[baseline={(0,0)}, scale=0.95]
      \node[inner sep=0, anchor=base] {\includegraphics[width=0.3\textwidth]{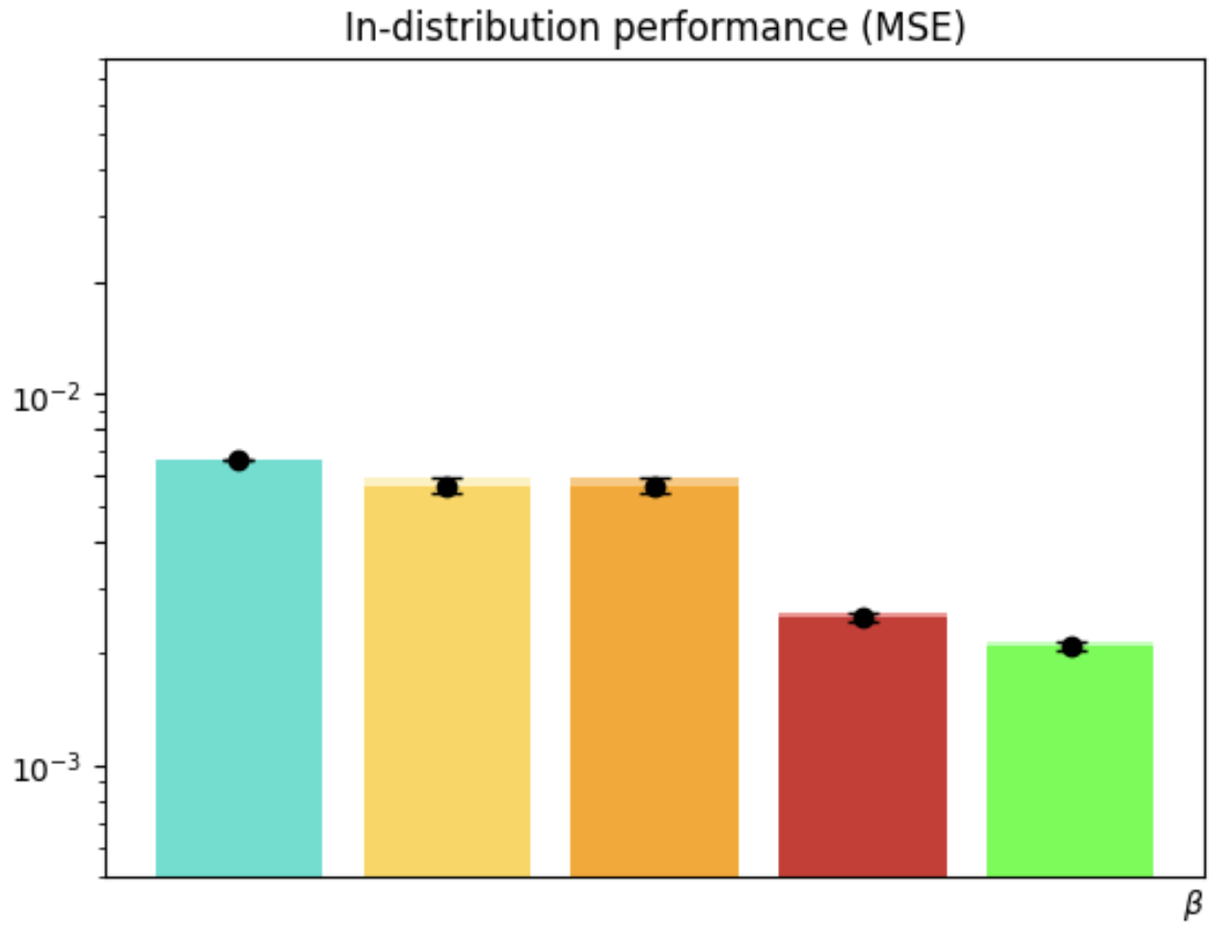}};
      % \node[font=\tiny] at (0.0, 0.2) {$0.1$};
      % \node[font=\fontsize{4pt}{5pt}\selectfont] at (0.8, 0.3) {$0.1$};
      \node at (-1.35, 0.13) {\scalebox{0.46}{$0.01$}};
      \node at (-0.6, 0.13) {\scalebox{0.46}{$0.1$}};
      \node at (0.13, 0.13) {\scalebox{0.46}{$1.0$}};
      \node at (0.86, 0.13) {\scalebox{0.46}{$10.0$}};
      \node at (1.63, 0.13) {\scalebox{0.46}{$100.0$}};
    \end{tikzpicture}%

    }
    % \hspace{-0.01cm} 
    \subfigure{\label{fig:id_baseline}
        \includegraphics[width=0.3\textwidth]{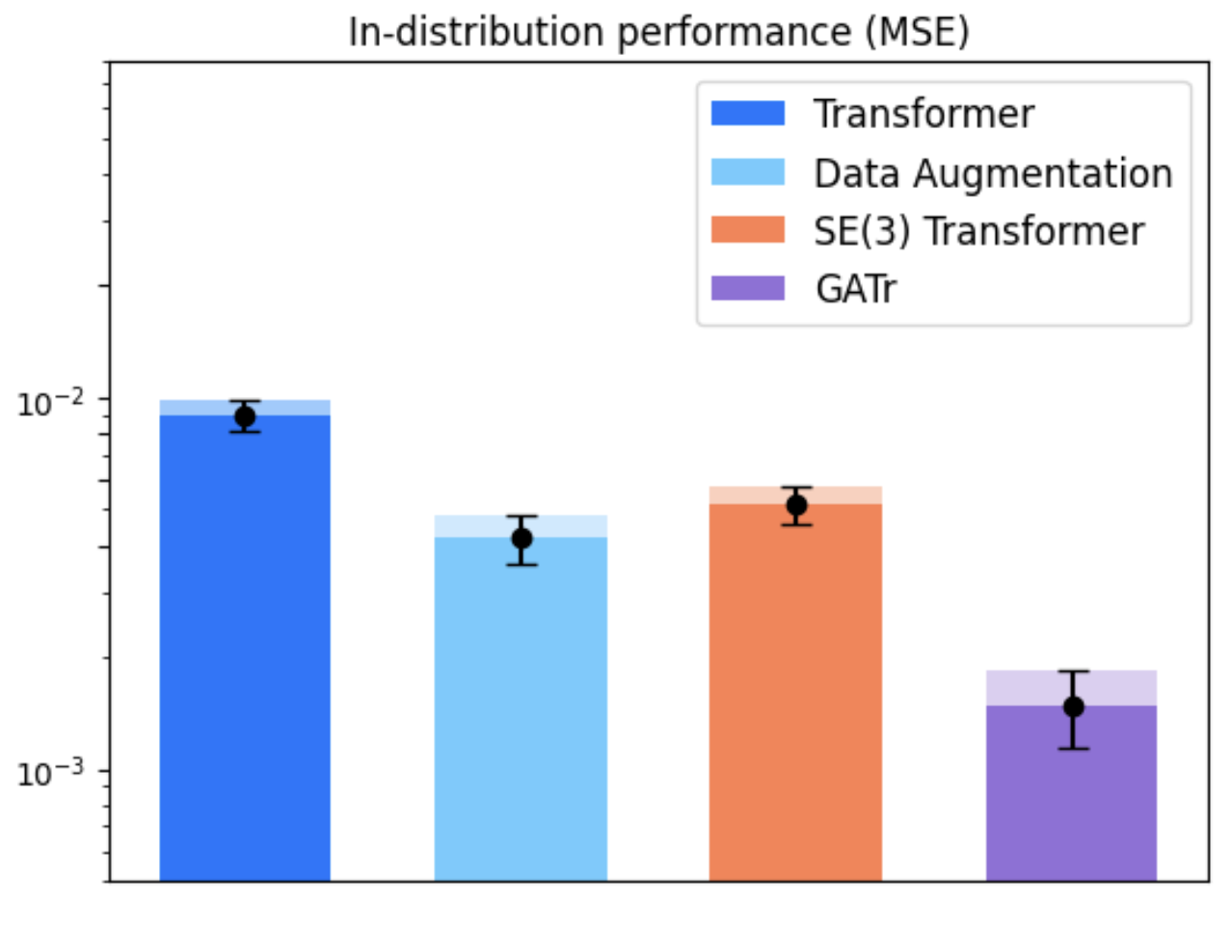}
    }
    \vspace{-0.2cm} 
    % Second row
    \subfigure{\label{fig:ood_grad}
        \includegraphics[width=0.3\textwidth]{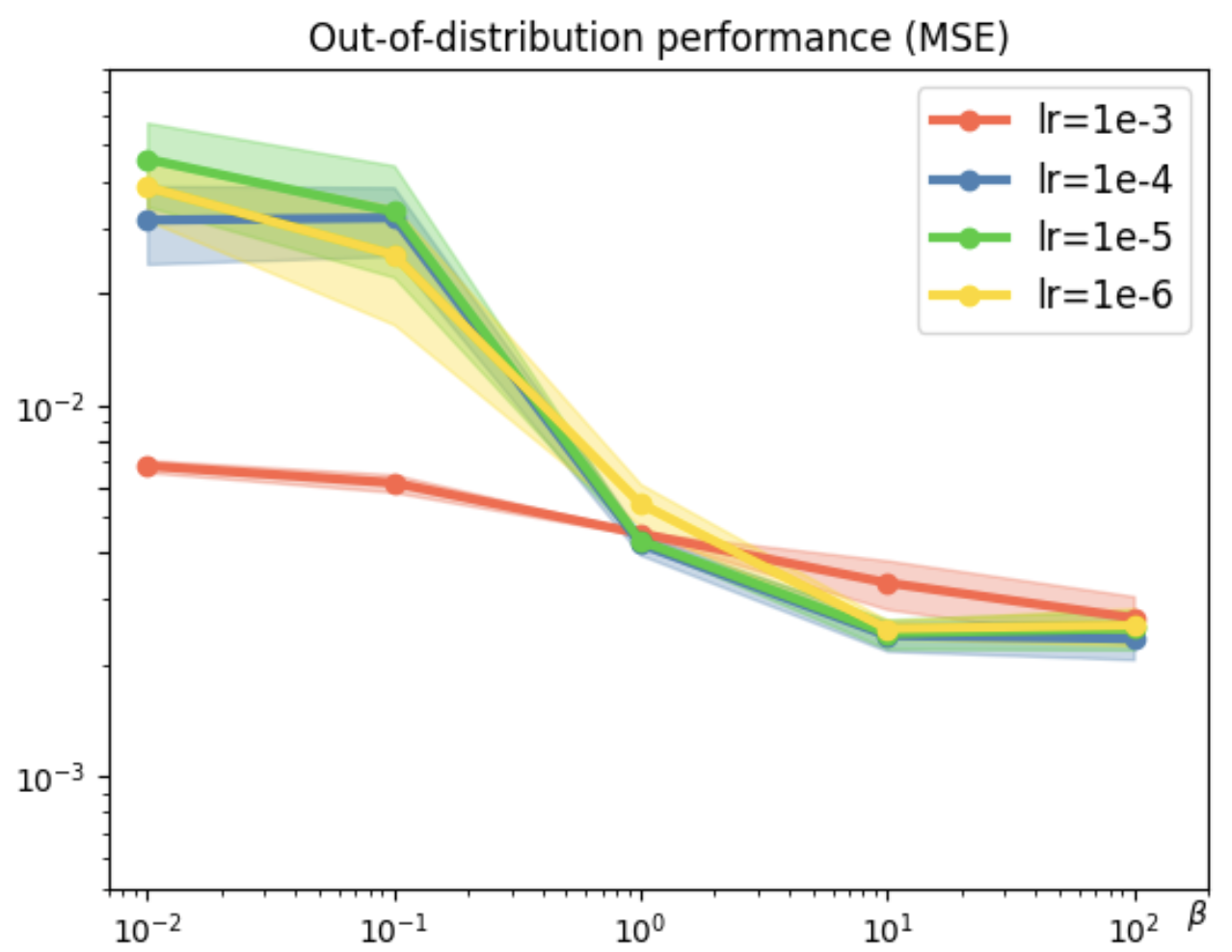}
    }
    % \hspace{-0.01cm} 
    \subfigure{\label{fig:ood_constant}
    
    \begin{tikzpicture}[baseline={(0,0)}, scale=0.95]
      \node[inner sep=0, anchor=base] {\includegraphics[width=0.3\textwidth]{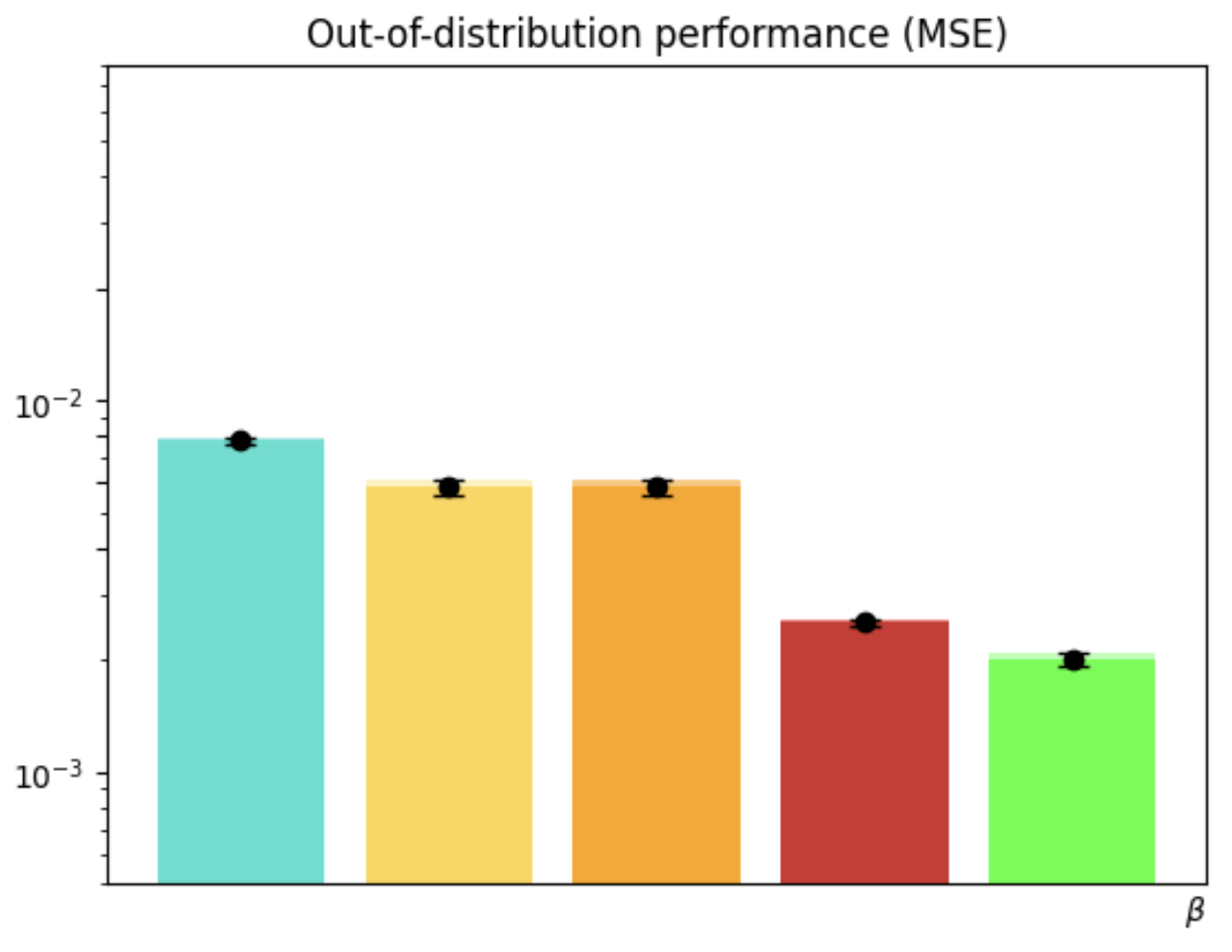}};
      % \node[font=\tiny] at (0.0, 0.2) {$0.1$};
      % \node[font=\fontsize{4pt}{5pt}\selectfont] at (0.8, 0.3) {$0.1$};
      \node at (-1.35, 0.13) {\scalebox{0.46}{$0.01$}};
      \node at (-0.6, 0.13) {\scalebox{0.46}{$0.1$}};
      \node at (0.13, 0.13) {\scalebox{0.46}{$1.0$}};
      \node at (0.86, 0.13) {\scalebox{0.46}{$10.0$}};
      \node at (1.63, 0.13) {\scalebox{0.46}{$100.0$}};
    \end{tikzpicture}%
    }
    % \hspace{-0.01cm} 
    \subfigure{\label{fig:ood_baseline}
        \includegraphics[width=0.3\textwidth]{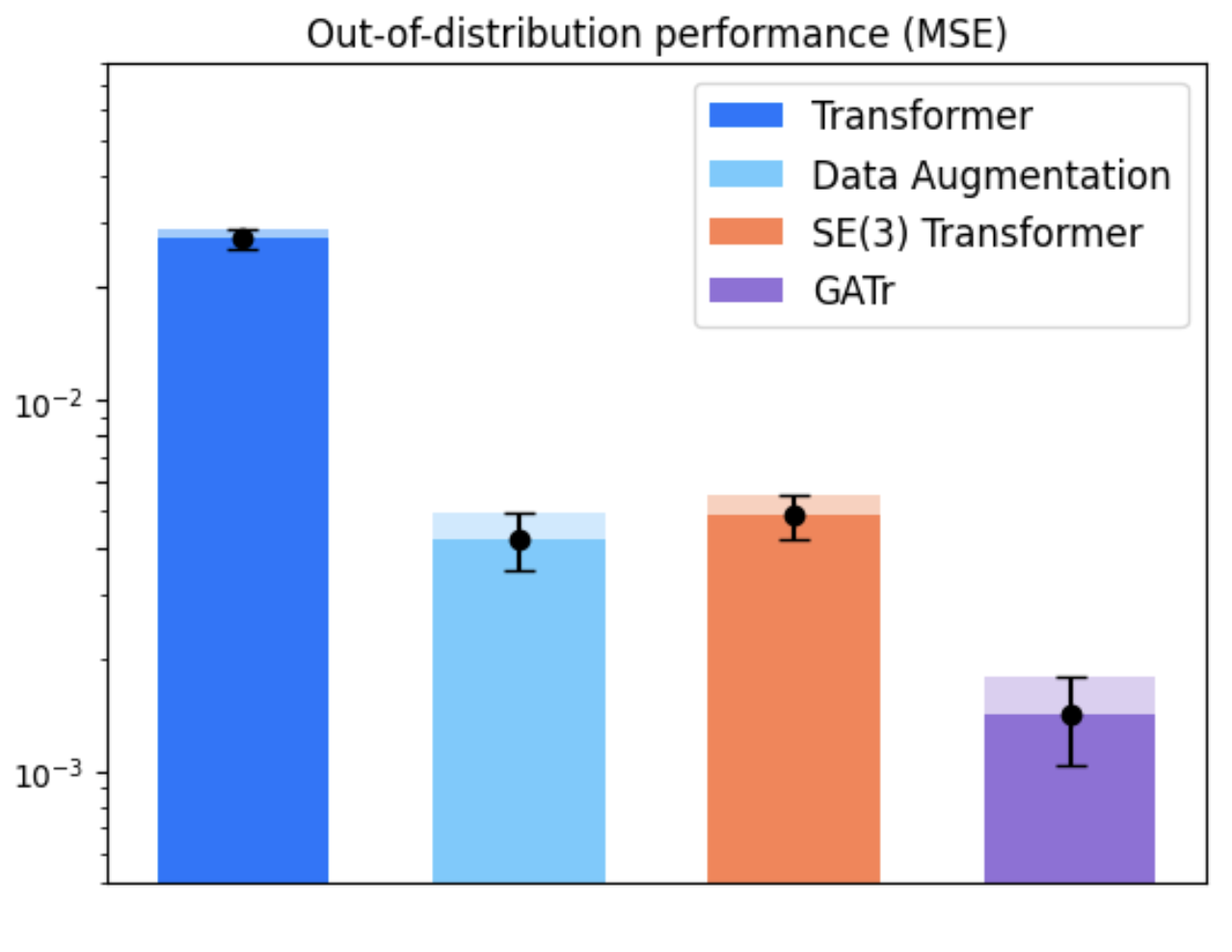}
    }
    % Third row
    \vspace{-0.4cm} 
    \subfigure{
    \label{fig:m1_grad}
        \begin{tikzpicture}[baseline={(0,0)}, scale=0.95]
      \node[inner sep=0, anchor=base] {\includegraphics[width=0.3\textwidth]{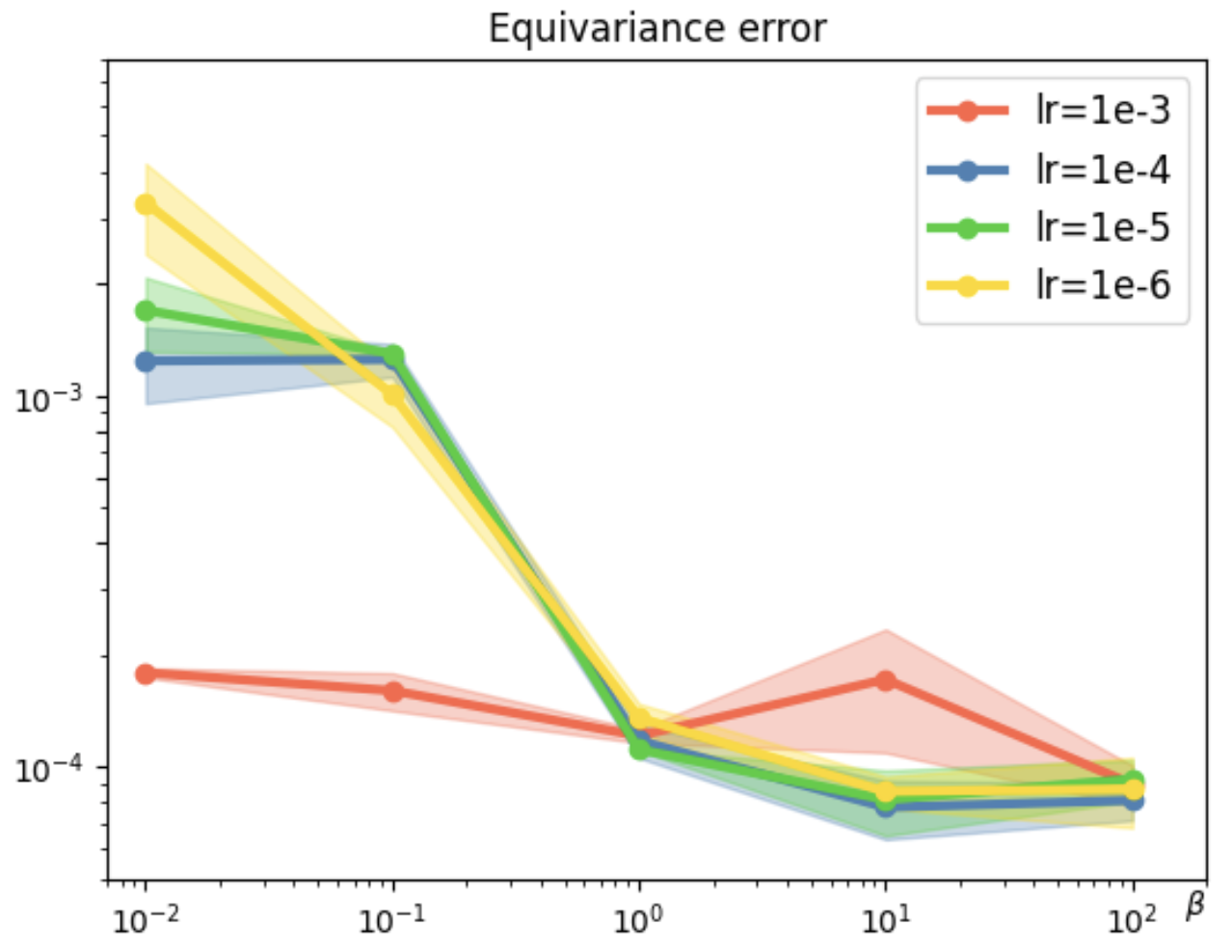}};
      \node at (0.15, -0.23) {\scalebox{0.9}{(a) REMUL: Gradual penalty}};
    \end{tikzpicture}%
    }
    % \hspace{-0.01cm} 
    \subfigure{\label{fig:m1_constant}
    \begin{tikzpicture}[baseline={(0,0)}, scale=0.95]
      \node[inner sep=0, anchor=base] {\includegraphics[width=0.3\textwidth]{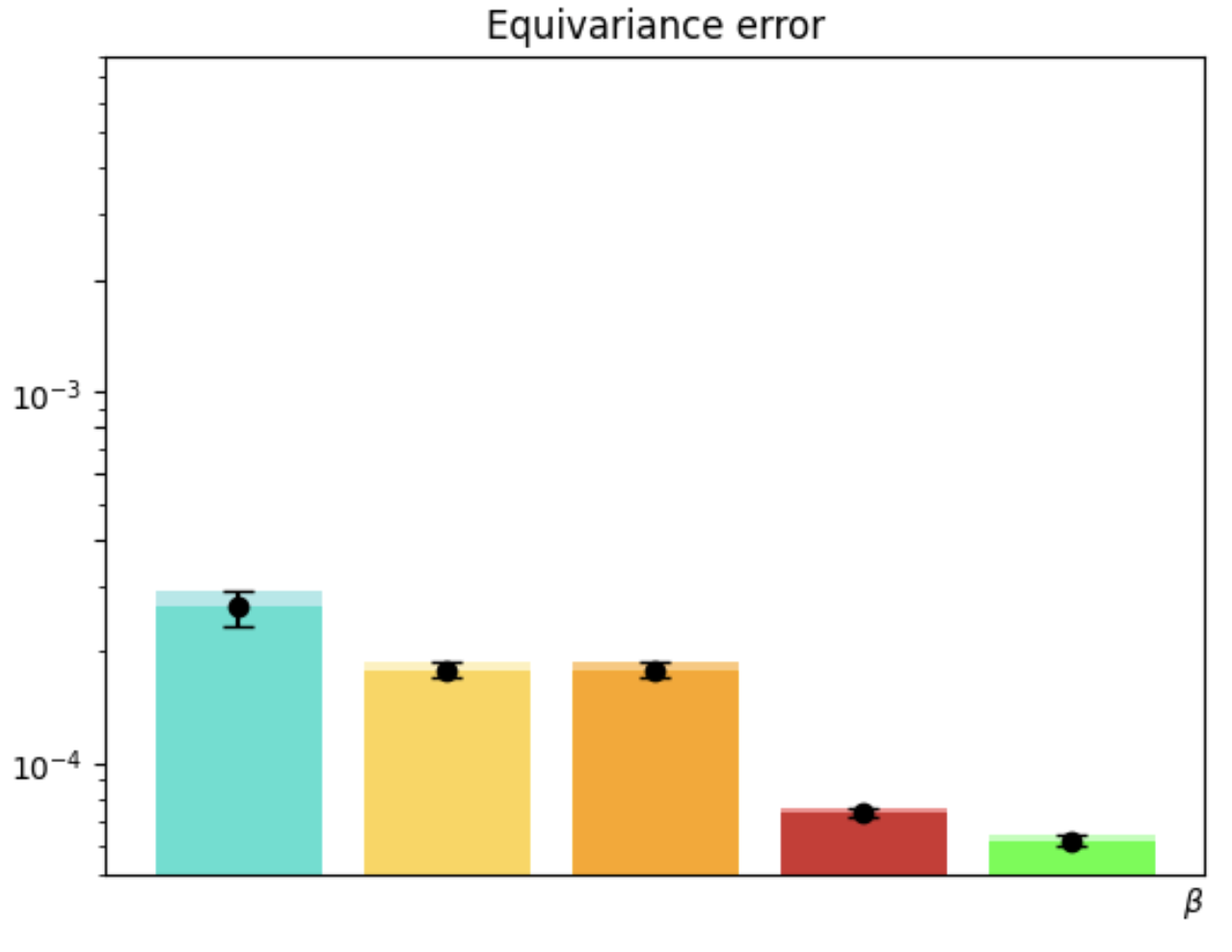}};
      % \node[font=\tiny] at (0.0, 0.2) {$0.1$};
      % \node[font=\fontsize{4pt}{5pt}\selectfont] at (0.8, 0.3) {$0.1$};
      \node at (-1.35, 0.15) {\scalebox{0.46}{$0.01$}};
      \node at (-0.6, 0.15) {\scalebox{0.46}{$0.1$}};
      \node at (0.13, 0.15) {\scalebox{0.46}{$1.0$}};
      \node at (0.88, 0.15) {\scalebox{0.46}{$10.0$}};
      \node at (1.64, 0.15) {\scalebox{0.46}{$100.0$}};
      \node at (0.15, -0.23) {\scalebox{0.9}{(b) REMUL: Constant penalty}};
    \end{tikzpicture}%
    }
    % \hspace{-0.01cm} 
\subfigure{\label{fig:m1_baseline}
  \raisebox{-0.2ex}{%  % Adjust this value to move up/down (1ex ≈ height of 'x')
    \begin{tikzpicture}[baseline={(0,0)}, scale=0.95]
      \node[inner sep=0, anchor=base] {\includegraphics[width=0.3\textwidth]{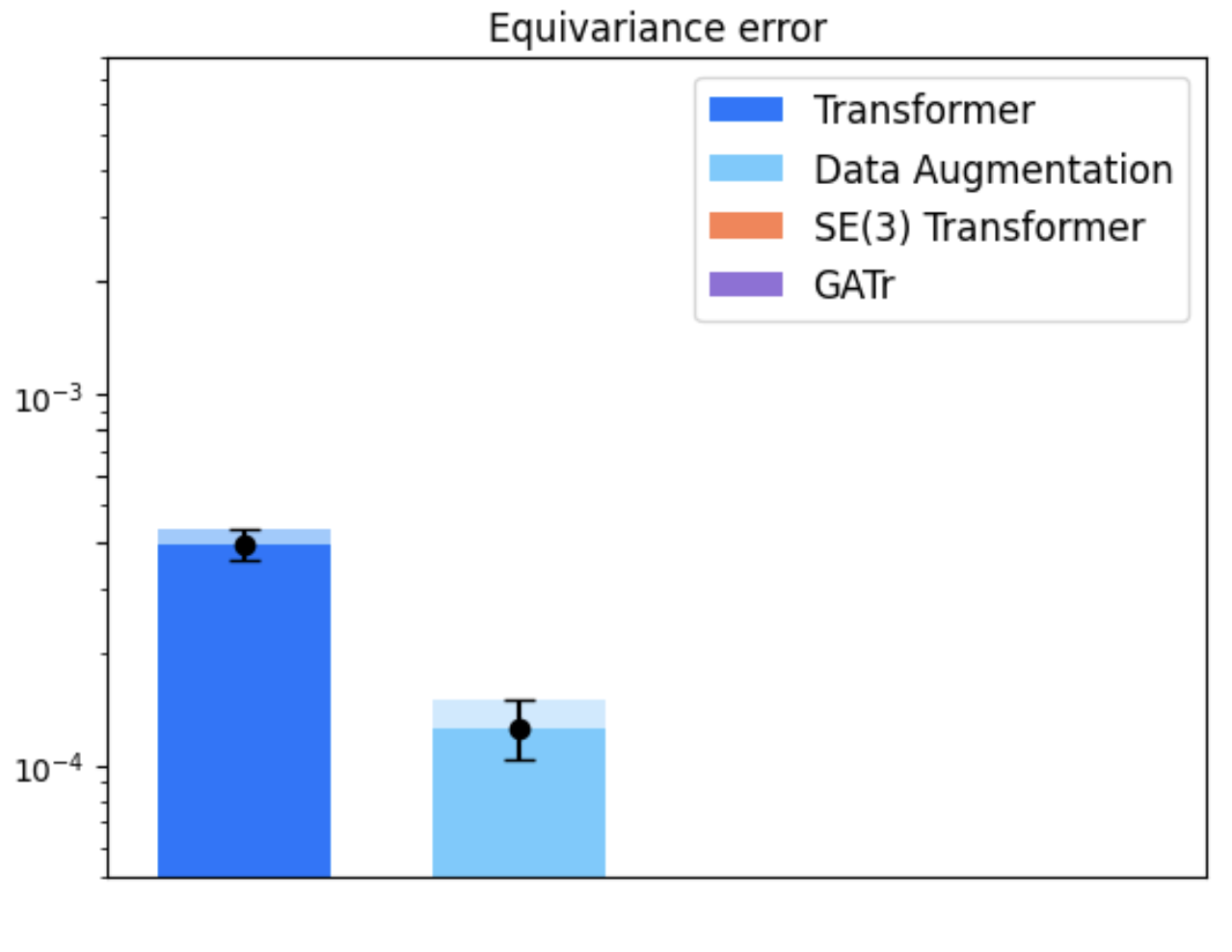}};
      % \fill[customorange] (0.34,-1.5) rectangle (0.9,-1.6);
      % \fill[custompurple] (1.27,-1.5) rectangle (1.8,-1.6);
      \fill[customorange] (0.45,0.09) rectangle (1.05,0.19);
      \fill[custompurple] (1.4,0.09) rectangle (2.0,0.19);
      % \node[font=\tiny] at (-3.8, 0.3) {$0.1$};
      \node at (0.15, -0.2) {\scalebox{0.9}{(c) Baselines}};
    \end{tikzpicture}%
  }
}
\vspace{2ex}
    \caption{N-body dynamical system. Each row represents a different evaluation scenario. Top: in-distribution performance, Middle: out-of-distribution performance, Bottom: equivariance error. The columns correspond to different architectures/ model conditions. (a) Transformer trained with REMUL (gradual penalty), (b) Transformer trained with a constant penalty, (c) Baselines (equivariant models, standard Transformer, and data augmentation). We conclude that Transformer architecture with high $\beta$ reduces the equivariance error and improves the performance. 
    % SE(3)-Transformer and GATr have a small equivariance error below the range of the plots ($2.8e^{-10}$ and $1.13e^{-15}$ respectively).
    }
    \label{fig: performance on N body system}
% \vspace{-2ex}
\end{figure}
We evaluate our method on different tasks for geometric data: N-body dynamical system (Section \ref{sec: N Body dynamical system}),
% (\hyperref[sec: N Body dynamical system]{Section \ref*{sec: N Body dynamical system}}), 
motion capture (Section \ref{sec: motion capture}), and molecular dynamics (Section \ref{sec: Molecular Dynamics}). 
% We apply our equivariance loss to unconstrained models that do not consider roto-translation equivariance in their design: Transformer and Graph Neural Networks.
For unconstrained models, we apply
REMUL to Transformers and Graph Neural Networks. 
We then compare against their equivariant counterparts: SE(3)-Transformer \citep{fuchs2020se3transformers}, Geometric Algebra Transformer \citep{brehmer2023geometric}, and  Equivariant Graph Neural Networks \citep{satorras2022en} as well as unconstrained models with data augmentation. We consider learning the rotation group $SO(3)$ for REMUL and data augmentation and we subtract the center of mass for translation. We use the equivariance metric defined in Eq. \ref{eq:equi_measure_1} to analyze our results, and include the second metric in Appendix \ref{appendix: Additional Experiments}. 
We also conduct a comparative analysis for the computational requirements of unconstrained models and equivariant models in Section \ref{sec: time complexity}. Lastly, we discuss the loss surfaces in Appendix \ref{sec: Loss surface}. 
Implementation details and additional experiments can be found in 
Appendix \ref{appendix: implementation_details}
\&
Appendix \ref{appendix: Additional Experiments}.
% Appendices \ref{appendix: implementation_details} \& \ref{appendix: Additional Experiments}.
\begin{figure}[t!]
    \centering
    \subfigure[MSE: Walking]{
        \includegraphics[width=0.22\textwidth]{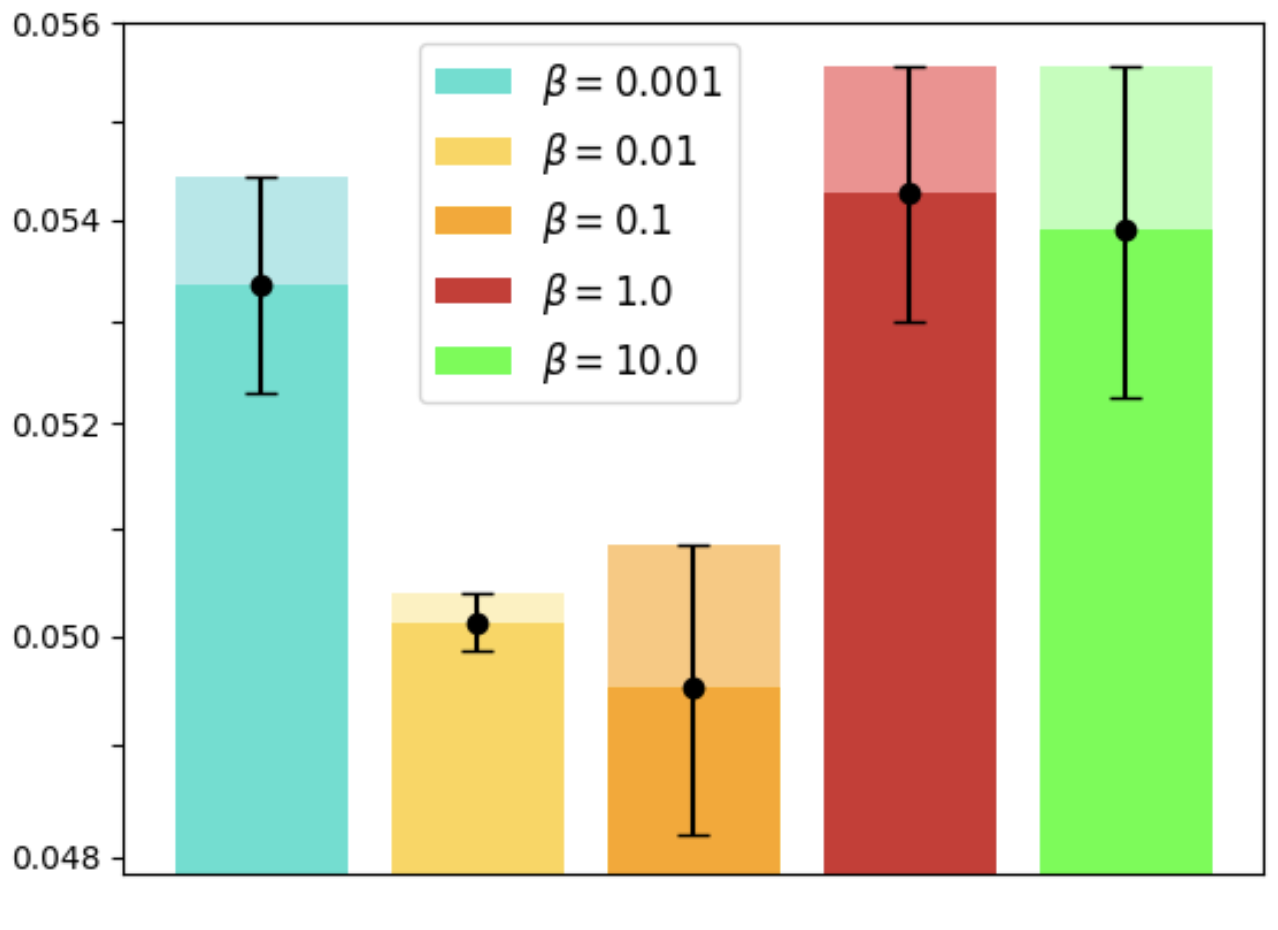}
}
% \hspace{-0.01cm} 
    \subfigure[Equiv. error: Walking]{
        \includegraphics[width=0.22\textwidth]{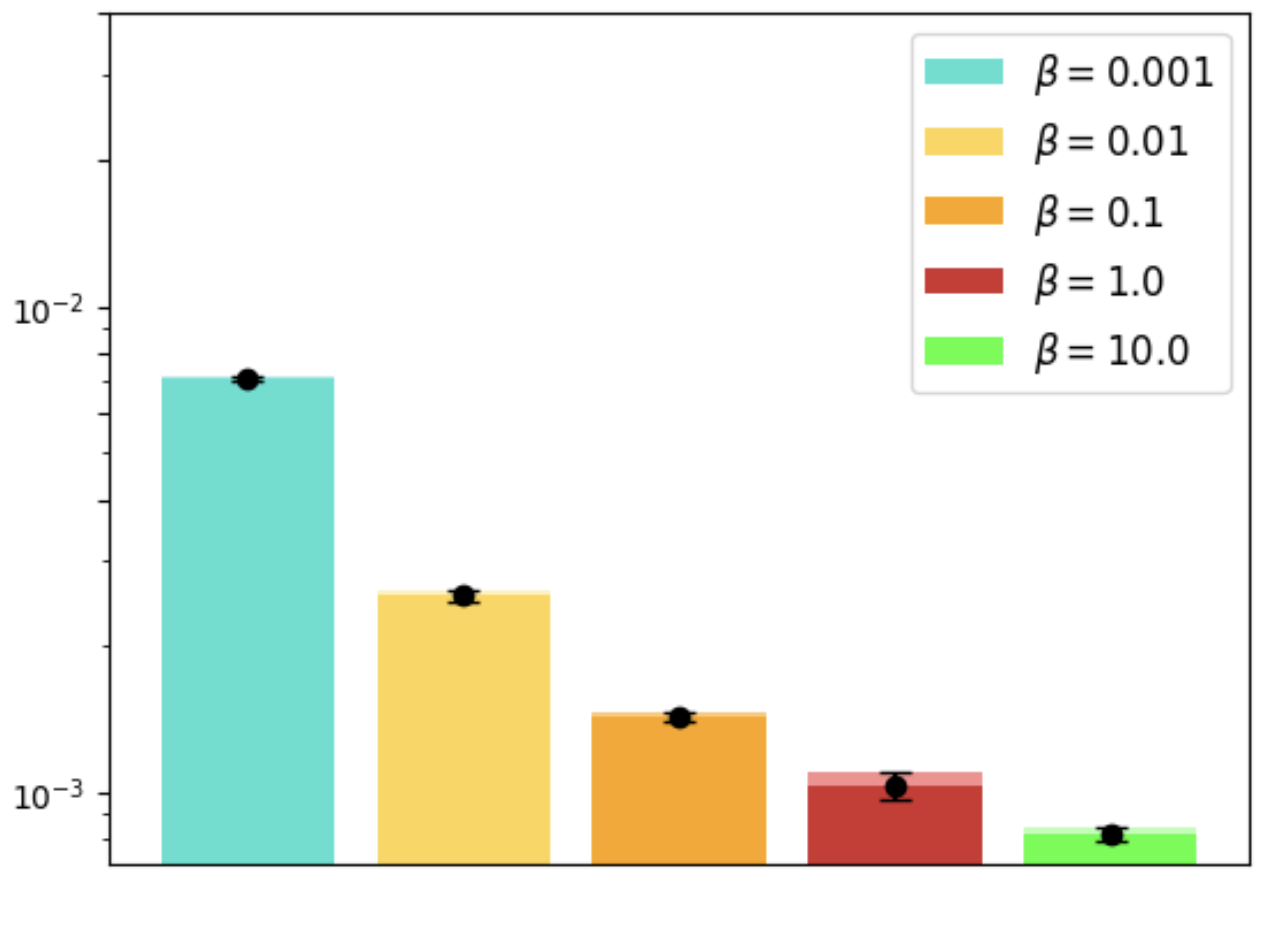}
}
% \hspace{-0.01cm} 
    \subfigure[MSE: Running]{
        \includegraphics[width=0.22\textwidth]{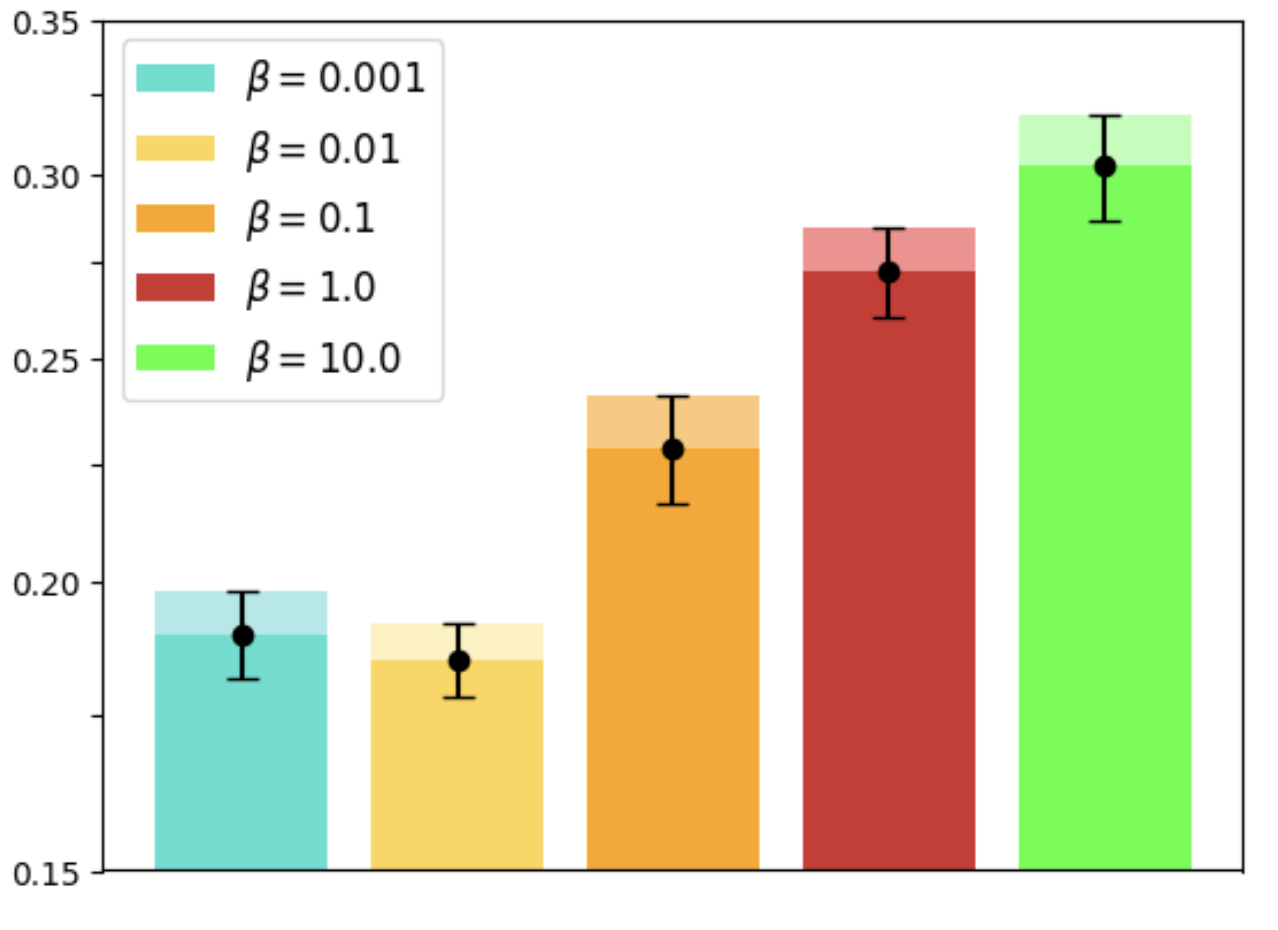}
}
\vspace{-0.3cm} 
     \subfigure[Equiv. error: Running]{
        \includegraphics[width=0.22\textwidth]{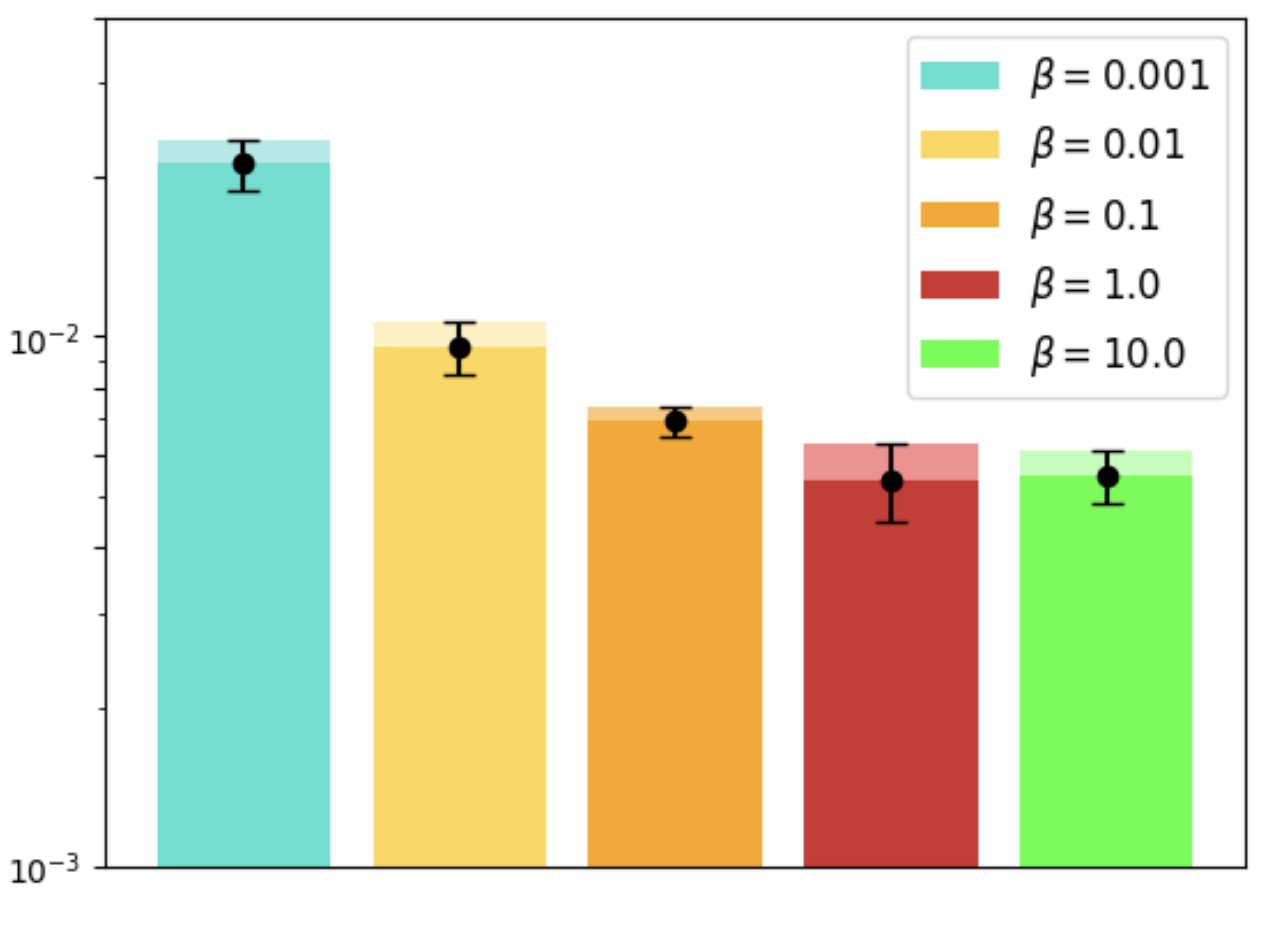}
}
\vspace{2ex}
    \caption{Motion Capture dataset: Transformer trained with REMUL.  
    % Two figures on the left: Performance (MSE) and equivariance error for walking task (Subject \#$35$), respectively. Two figures on the right: Performance (MSE) and equivariance error for running task (Subject \#$9$), respectively. 
    We show a trade-off between model performance and equiv. error, where high penalty $\beta$ gives less equiv. error (more equivariant model) but the best performance comes at an intermediate level of equivariance for both tasks.}
    \label{fig: performance Motion capture dataset}
\end{figure}
\subsection{N-Body Dynamical System}
\label{sec: N Body dynamical system}
\begin{wraptable}{r}[0pt]{0.35\textwidth}
\caption{N-body dynamical system: MSE ($\times 10^{-3}$).
{\color{red} First}, {\color{orange} Second}.}
\label{tab:performance:nbody}
\vskip 0.15in
\centering
\begin{adjustbox}{scale=0.77}
\begin{tabular}{lcc}
\hline
 & In-dist. & OOD \\
\hline
SE(3)-Tr     & $5.16$\tiny$\pm0.70$      & $4.85$\tiny$\pm0.78$ \\
GATr         & {\color{red}$1.49$\tiny$\pm0.43$} & {\color{red}$1.41$\tiny$\pm0.46$} \\
Transformer  & $8.99$\tiny$\pm1.04$      & $27.06$\tiny$\pm2.01$ \\
DA-Tr        & $4.20$\tiny$\pm0.79$      & $4.21$\tiny$\pm0.91$ \\
REMUL-Tr     & {\color{orange}$1.94$\tiny$\pm0.01$} & {\color{orange}$1.83$\tiny$\pm0.04$} \\
\hline
\end{tabular}
\end{adjustbox}
\end{wraptable}

To conduct ablation studies of our method, we utilized the dynamical system problem described by \cite{brehmer2023geometric}. The task involves predicting the positions of particles after $100$ Euler time steps of Newton's motion equation, given initial positions, masses, and velocities. This problem is equivariant under rotation and translation groups, implying that any rotation/translation of the initial states should rotate/translate the final states of the particles by the same amount.
We conduct comparisons between Transformer trained with REMUL against two equivariant architectures: SE(3)-Transformer and Geometric Algebra Transformer (GATr). 
We use the same Transformer version and hyperparameters specified by 
% with the same hyperparameters specified by 
\cite{brehmer2023geometric}. 
% Additional implementation details are provided in 
Implementation details, including in-distribution and out-of-distribution settings, in
Appendix \ref{appendix: n_body_system}.
Our results are presented in Figure \ref{fig: performance on N body system} and Table \ref{tab:performance:nbody}. 

From Figure \ref{fig: performance on N body system}, we noticed that increasing the penalty parameter $\beta$ of the equivariance loss significantly reduces the equivariance error in both constant and gradual settings (which results in a more equivariant model). Equivariant architectures demonstrate an equivariance error near zero, which is expected by their design. 
% The performance behaves similarly, with a higher penalty enhancing generalization capabilities for in and out-of distributions. 
The performance behaves similarly; a higher penalty enhances model generalization for both in-distribution and out-of-distribution.
Transformer with high $\beta$ outperforms both data augmentation and SE(3)-Transformer across in-distribution and out-of-distribution and competes with GATr. We also observe that despite SE(3)-Transformer having a substantially lower equivariance error, its performance is slightly worse than Transformer trained with data augmentation. 
This highlights that equivariance, although improving generalization in this task, is only one aspect of understanding model performance.
Lastly, the standard Transformer (without REMUL and data augmentation) exhibits the highest equivariance error and the lowest overall performance.
\subsection{Motion Capture}
\label{sec: motion capture}
We further illustrate a comparison on a real-world task, the Motion Capture dataset from \cite{cmumocap2003}. This dataset features 3D trajectory data that records a range of human motions, and the task involves predicting the final trajectory based on initial positions and velocities. We have reported results for two types of motion: Walking (Subject \#$35$) and Running (Subject \#$9$). We adhered to the standard experimental setup found in the literature \citep{han2022equivariantgraphhierarchybasedneural, huang2022equivariantgraphmechanicsnetworks, xu2024equivariantgraphneuraloperator}, employing a train/validation/test split of $200$/$600$/$600$ for Walking and $200$/$240$/$240$ for Running (additional details in Appendix \ref{appendix: Motion Capture}). 

We apply our training procedure REMUL to the Transformer architecture and compare it with SE(3)-Transformer, Equivariant Graph Neural Operator (EGNO) \cite{xu2024equivariantgraphneuraloperator}, Geometric Algebra Transformer (GATr), standard Transformer, and Transformer trained with data augmentation.
We also compare with Equivariant MLP \citep{finzi2021practicalmethodconstructingequivariant}, as well as two approximate equivariance architectures: Residual Pathway Priors (RPP) \citep{NEURIPS2021_fc394e99}, and Projection-Based Equivariance Regularizer (PER) \citep{kim2023regularizingsoftequivariancemixed}. As these architectures are designed specifically on MLP and linear layers, we apply our method to a standard MLP with a similar number of parameters. Our results are presented 
% \begin{paracol}{2}
% \sloppy  
in Table \ref{tab:performance:Motion Capture dataset}. For REMUL, we provide plots on how the performance and equivariance error change \textit{w.r.t.} the penalty parameter $\beta$ in Figure \ref{fig: performance Motion capture dataset}.
% \switchcolumn
\begin{wraptable}{r}[0pt]{0.37\textwidth}
\caption{Motion Capture dataset: MSE ($\times 10^{-2}$). REMUL procedure and data augmentation (DA) were applied to standard Transformer and MLP. {\color{red} First}, {\color{orange} Second}. REMUL comes best in both tasks.}
\label{tab:performance:Motion Capture dataset}
\vskip 0.15in
\centering
% First comparison table (Transformers)
\begin{adjustbox}{scale=0.77}
\begin{tabular}{lcc}
\hline
 & Walking & Running \\
\hline
SE(3)-Tr & $10.85$\tiny$\pm1.3$ & $42.13$\tiny$\pm3.4$ \\
GATr & $10.06$\tiny$\pm1.3$ & $32.38$\tiny$\pm3.9$ \\
% EGNN         & $28.7$\tiny$\pm1.6$   & $50.9$\tiny$\pm0.9$ \\
% EGNN-R       & $90.7$\tiny$\pm2.4$   & $816.7$\tiny$\pm2.7$ \\
% EGNN-S       & $26.4$\tiny$\pm1.5$   & $54.2$\tiny$\pm1.9$ \\
EGNO         & $8.1$\tiny$\pm1.6$    & $33.9$\tiny$\pm1.7$ \\
Transformer & {\color{orange}$5.21$\tiny$\pm0.08$} & {\color{orange}$20.78$\tiny$\pm1.5$} \\
DA-Tr  & $5.3$\tiny$\pm0.18$ & $29.83$\tiny$\pm1.4$ \\
REMUL-Tr & {\color{red}$4.95$\tiny$\pm0.1$} & {\color{red}$18.5$\tiny$\pm0.7$} \\
\hline
% \hline
% \vspace{1em}
EMLP & $7.01$\tiny$\pm0.46$ & $57.38$\tiny$\pm8.39$ \\
RPP & $6.99$\tiny$\pm0.21$ & $34.18$\tiny$\pm2.00$ \\
PER & $7.48$\tiny$\pm0.39$ & {\color{orange}$33.03$\tiny$\pm0.37$} \\
MLP & $6.80$\tiny$\pm0.18$ & $39.56$\tiny$\pm2.25$ \\
DA-MLP & {\color{orange}$6.37$\tiny$\pm0.04$} & $40.23$\tiny$\pm0.94$ \\
REMUL-MLP & {\color{red}$6.04$\tiny$\pm0.09$} & {\color{red}$32.57$\tiny$\pm1.47$} \\
\hline
\end{tabular}
\end{adjustbox}
\end{wraptable}
% \end{paracol}

Table \ref{tab:performance:Motion Capture dataset} indicates that when processing 3D positions related to human motions, both SE(3)-Transformer and GATr perform worse than the standard Transformer. This outcome is noteworthy because human motion often lacks full rotational symmetry, particularly along the vertical or gravity axis. 
% Indeed, as detailed in Appendix~\ref{app:axis_equivariance_motion} (Table~\ref{tab:appendix_axis_equivariance}), our analysis of axis-specific equivariance errors for \methodabbr{}-Transformer confirms that the error is highest for rotations around the Z-axis (typically aligned with gravity).
In fact, as detailed in the Appendix~\ref{Additional Experiments: Motion Capture} (Table~\ref{tab:seperate equi error motion_capture}), our analysis of axis-specific equivariance errors for \methodabbr{}-Transformer confirms that the error is highest for rotations around the $Z$-axis. Consequently, imposing strict $SO(3)$ equivariance across all axes may not be beneficial and can be detrimental to performance. 
% Consequently, imposing strict $SO(3)$ equivariance across all axes may not be beneficial or even detrimental. 
In contrast, a standard Transformer trained with REMUL has the best performance in both tasks. 
% Following Figure \ref{fig: performance Motion capture dataset}, there is a noticeable trade-off in model performance with different values of penalty parameter $\beta$. Best performance is observed at an intermediate level of equivariance, where the model balances between being too rigid (fully equivariant) and too flexible (non-equivariant).
% This finding underscores the importance of carefully considering the specific characteristics of the data and the task when designing equivariant architectures.
Following Figure~\ref{fig: performance Motion capture dataset}, there is a noticeable trade-off: while higher $\beta$ values reduce overall equivariance error, optimal task performance is often observed at an intermediate level of learned equivariance, where the model balances between being too rigid (fully equivariant) and too flexible (non-equivariant). This underscores that the optimal degree of symmetry is task-dependent and that \methodabbr{}'s flexibility in learning approximate equivariance is advantageous for such real-world scenarios. 
\vspace{-1ex}
\subsection{Molecular Dynamics}
\label{sec: Molecular Dynamics}
We also present a comparative analysis between constrained equivariant models and unconstrained models focusing on molecular dynamics, specifically predicting 3D molecule structures. We utilize the MD17 dataset \citep{Chmiela_2017}, which comprises trajectories of eight small molecules. We use the same dataset split in \cite{huang2022equivariantgraphmechanicsnetworks, xu2024equivariantgraphneuraloperator}, allocating $500$ samples for train, $2000$ for validation, and $2000$ for test. For this task, 
% we selected the equivariant graph neural network (EGNN) architecture introduced by \citet{satorras2022en} and its non-equivariant version GNN. 
we selected the Equivariant Graph Neural Network (EGNN) architecture and its non-equivariant GNN counterpart, as presented in \cite{satorras2022en}.
We then apply REMUL procedure as well as data augmentation to the GNN architecture. 
Both architectures have the same hyperparameters (more information is indicated in Appendix \ref{appendix: Molecular Dynamics}).
We also compare with GMN \citep{huang2022equivariantgraphmechanicsnetworks}, EGNO \citep{xu2024equivariantgraphneuraloperator}, and HEGNN \cite{cen2024high}. 
Our results are provided in Table \ref{tab:performance:MD17}.
We illustrate how the performance and equivariance error of a GNN trained with REMUL vary across different molecules as a function of $\beta$ in Figure \ref{fig:md17} and Figure \ref{fig:Additional MD17}.
\begin{table}[h]
% \caption{Performance on MD17 dataset: MSE ($\times 10^{-2}$). Top-1 in {\color{red}red}, top-2 in {\color{orange}orange}.}
\caption{MD17 dataset: MSE ($\times 10^{-2}$). REMUL procedure and data augmentation (DA) were applied to GNN. {\color{red} First}, {\color{orange} Second}.}
\label{tab:performance:MD17}
\vskip 0.15in
\centering
\begin{adjustbox}{scale=0.72}
\begin{tabular}{lcccccccc}
\toprule
 & Aspirin & Benzene & Ethanol & Malonaldehyde & Naphthalene & Salicylic & Toluene & Uracil \\
\midrule
EGNN               & $14.41_{\pm0.15}$ & $62.40_{\pm0.53}$ & $4.64_{\pm0.01}$ & $13.64_{\pm0.01}$ & $0.47_{\pm0.02}$ & $1.02_{\pm0.02}$ & $11.78_{\pm0.07}$ & $0.64_{\pm0.01}$ \\
GMN                & $10.14_{\pm0.03}$ & $48.12_{\pm0.4}$  & $4.83_{\pm0.01}$ & $13.11_{\pm0.03}$ & {\color{orange}$0.40_{\pm0.01}$} & $0.91_{\pm0.01}$ & $10.22_{\pm0.08}$ & $0.59_{\pm0.01}$ \\
EGNO               & {\color{red}$9.18_{\pm0.06}$} & $48.85_{\pm0.55}$ & $4.62_{\pm0.01}$ & {\color{red}$12.80_{\pm0.02}$} & {\color{red}$0.37_{\pm0.01}$} & {\color{red}$0.86_{\pm0.02}$} & $10.21_{\pm0.05}$ & {\color{red}$0.52_{\pm0.02}$} \\
HEGNN              & $9.94_{\pm0.07}$ & $59.93_{\pm5.21}$ & $4.62_{\pm0.01}$ & {\color{orange}$12.85_{\pm0.01}$} & {\color{red}$0.37_{\pm0.02}$} & {\color{orange}$0.88_{\pm0.02}$} & $10.56_{\pm0.33}$ & {\color{orange}$0.54_{\pm0.01}$} \\
GNN                & {\color{orange}$9.26_{\pm0.40}$} & {\color{orange}$26.13_{\pm0.11}$} & {\color{orange}$4.26_{\pm0.03}$} & $18.45_{\pm0.54}$ & $0.54_{\pm0.001}$ & $1.02_{\pm0.02}$ & {\color{orange}$9.93_{\pm0.82}$} & $0.70_{\pm0.001}$ \\
DA-GNN   & $13.7_{\pm0.04}$  & $110.93_{\pm5.3}$ & $5.74_{\pm0.02}$ & $13.65_{\pm0.02}$ & $0.69_{\pm0.001}$ & $1.33_{\pm0.04}$ & $19.14_{\pm0.001}$ & $0.73_{\pm0.002}$ \\
REMUL-GNN              & {\color{orange}$9.28_{\pm0.40}$}  & {\color{red}$25.95_{\pm0.18}$} & {\color{red}$4.02_{\pm0.16}$} & $13.59_{\pm0.03}$ & $0.54_{\pm0.001}$ & $0.99_{\pm0.001}$ & {\color{red}$9.38_{\pm0.20}$} & $0.67_{\pm0.001}$ \\
\bottomrule
\end{tabular}
\end{adjustbox}
\end{table}

From the results presented in Table \ref{tab:performance:MD17}, GNN trained with REMUL outperforms EGNN in six out of eight molecules.  Interestingly, a standard GNN, without data augmentation or REMUL, surpasses the  performance of EGNN on multiple molecules, such as Aspirin and Toluene. In Figure \ref{fig:md17} \& Figure \ref{fig:Additional MD17},
% In Figures \ref{fig:md17} \& \ref{fig:Additional MD17},
% we observe that each molecule has a different response to the penalty parameter $\beta$ in terms of the model performance. 
we observe that the optimal performance of each molecule is attained at different values of the penalty parameter $\beta$.
For instance, Malonaldehyde exhibits a direct correlation between model performance and equivariance, where a higher $\beta$ yields better performance. Conversely, for most other molecules, there appears to be a pronounced trade-off where the best performance is achieved at a lower value of $\beta$. This is particularly evident with molecules like Aspirin, where a standard GNN architecture outperforms EGNN.
We also plot the 3D structures of the eight molecules in Figure \ref{fig:MD17 molecules structures}. Molecules such as Malonaldehyde, characterized by their symmetric components, might be ideally suited for equivariant design. However, this advantage does not apply to all molecules. Aspirin on the other side, might have an asymmetric structure and exhibit a range of interactions and dynamic states that equivariant models might simplify. Consequently, for such molecules, less equivariant models could potentially offer more accurate predictions.

Finally, while REMUL also achieves competitive performance relative to other equivariant models such as EGNO and HEGNN, it is important to note that these models incorporate architectural elements that enhance geometric properties in distinct ways and might not be directly comparable to a simple GNN. For example, EGNO employs additional Fourier features, while HEGNN leverages high-degree steerable features.

\subsection{Computational Complexity}
\label{sec: time complexity}
\begin{figure}[h]
    \centering
    \subfigure[Combined forward pass]{
        \includegraphics[width=0.3\textwidth]{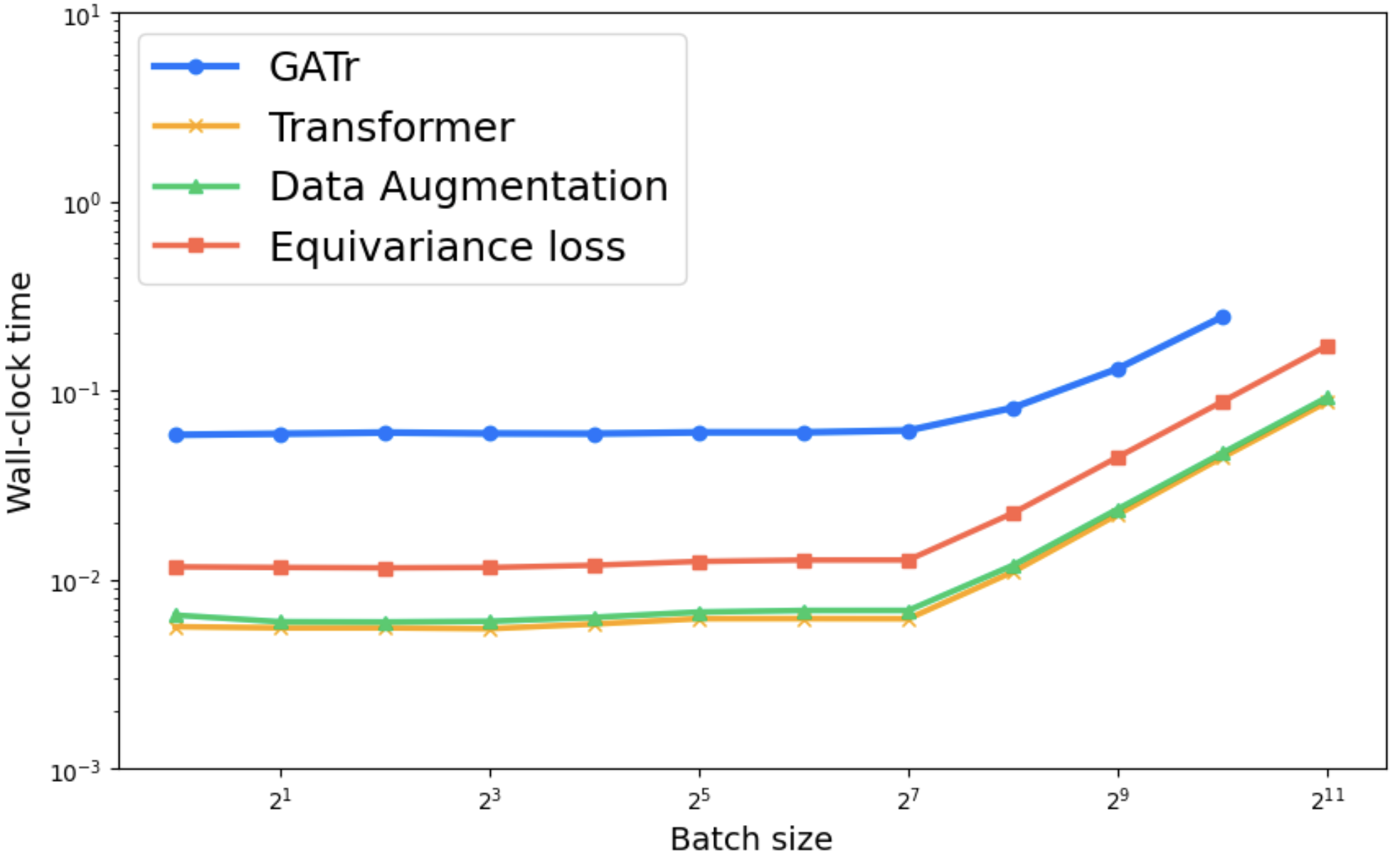}
    }
    % \hspace{-0.01cm} 
    \subfigure[Backward pass]{
        \includegraphics[width=0.3\textwidth]{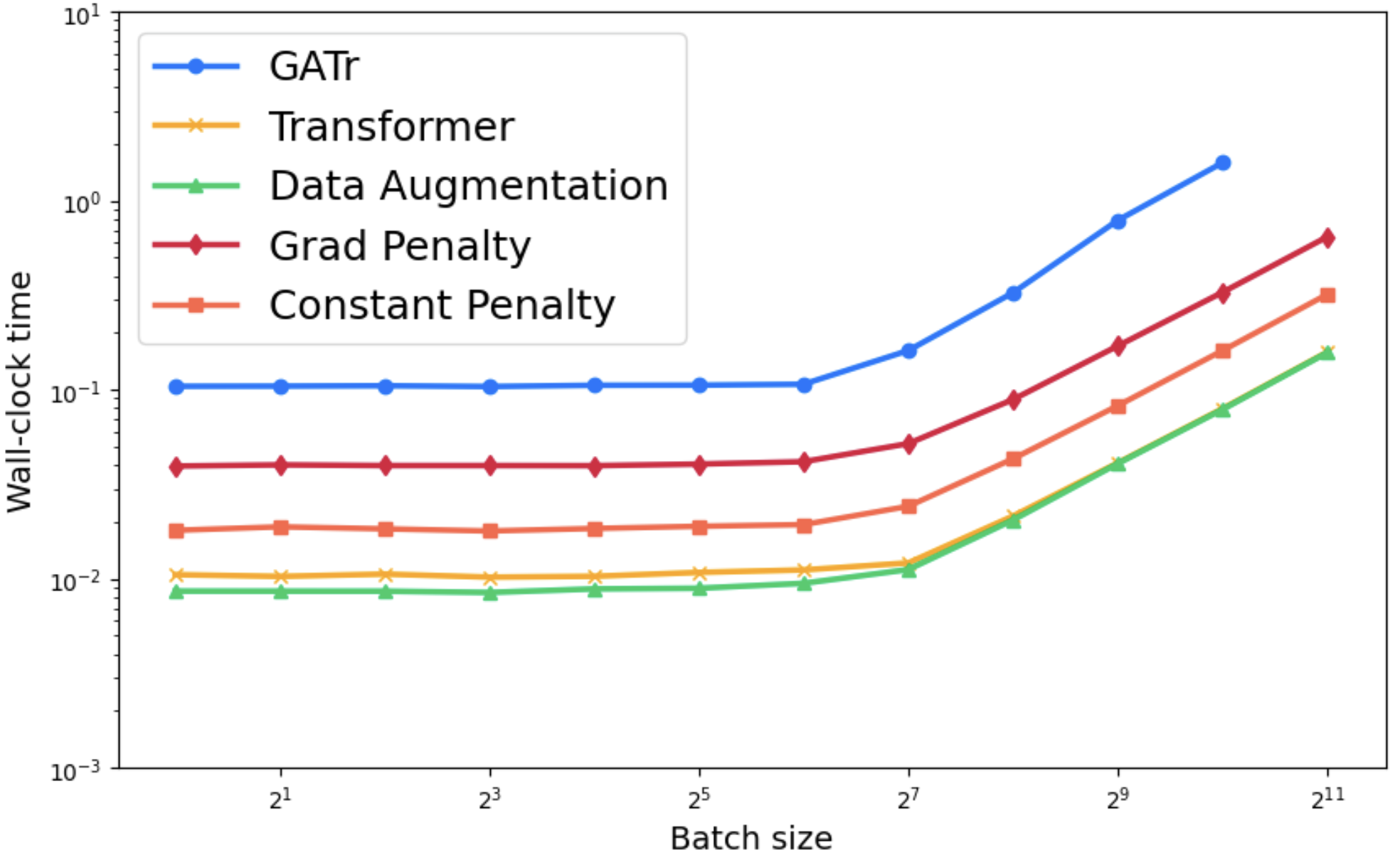}
    }
    % \hspace{-0.01cm} 
    \vspace{-0.2cm} 
    \subfigure[Inference time]{
        \includegraphics[width=0.3\textwidth]{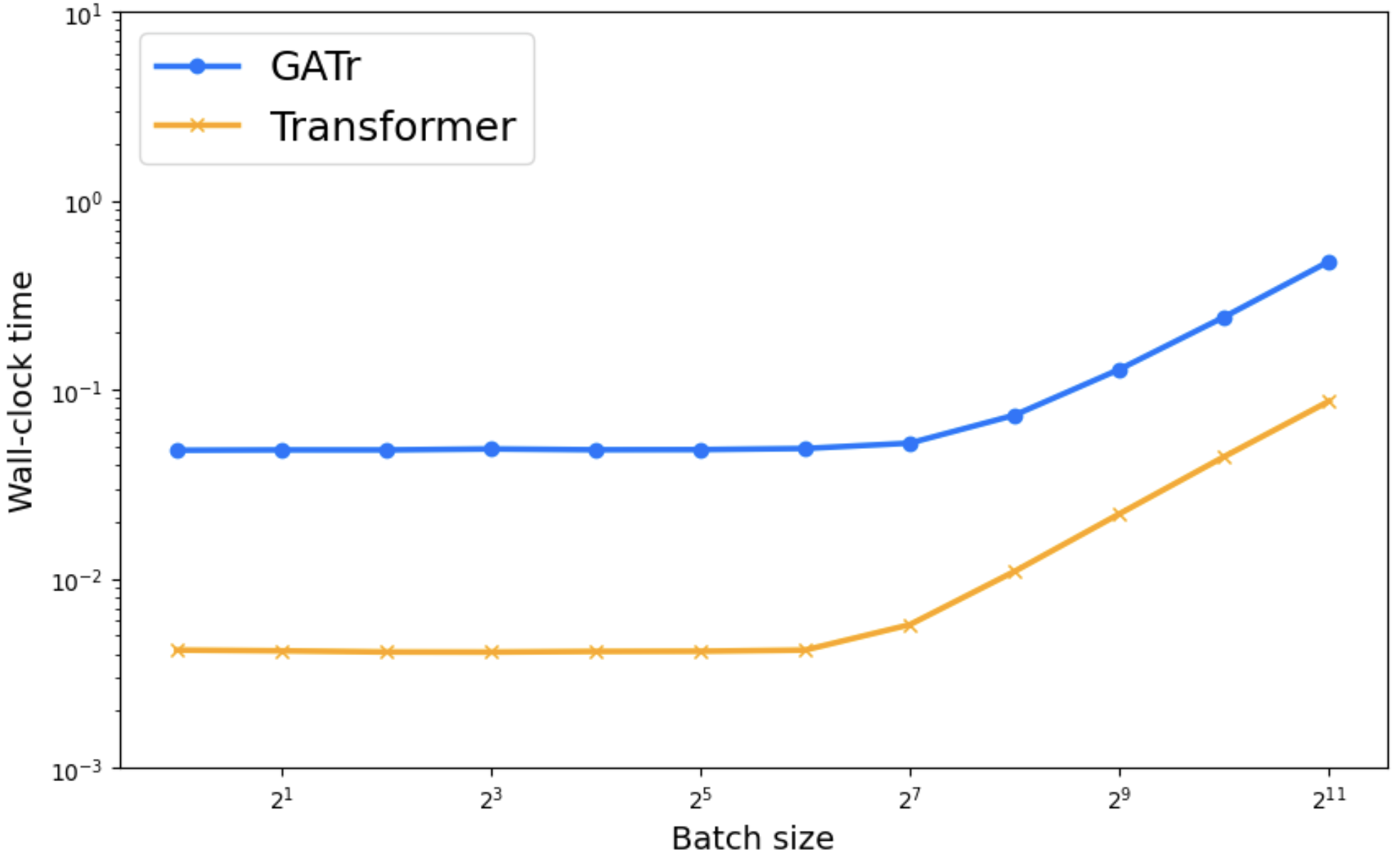}
    } 
    \vspace{1ex}
    \caption{Computational time for GATr and Transformer architectures. 
    % Plots from left to right: Combined forward pass, backward pass, and inference time, respectively.
    GATr has the highest time in all scenarios. Inference times for all versions of the Transformer (standard and trained with equivariance loss and data augmentation) are the same.}
    \vspace{-0.4cm}
%     % GATr has highest time in all scenarios. At inference, augmented Transformer matches standard Transformer}
    \label{fig:Time complexity}
\end{figure}

In this section, we report the computational time for the Geometric Algebra Transformer (GATr) and Transformer architectures. We selected models with an equivalent number of blocks and parameters for a fair comparison. 
% GATr incorporates a unique design that includes a multivector parameter, while we adjusted the Transformer to match the parameter count of GATr. Both models have around $2.6$M parameters. 
Detailed configurations are provided in Appendix \ref{appendix: Time Complexity}.
% SE(3)-Transformer gives out of memory for this setting.
We measured the computational efficiency of each model by recording the time taken for both forward and backward passes during training, as well as inference time. For the Transformer's computations, we also considered all the cases of data augmentation and our training procedure with the equivariance loss. 
% (constant and gradual penalties). 
Figure \ref{fig:Time complexity} includes the wall-clock time as a function of batch size with a fixed number of nodes.\\
% Add to preamble if not already included:
% Solution 2: Using \parbox for the caption
% \begin{wrapfigure}{r}{0.32\textwidth}
%     \centering
%     \vspace{-0.02cm}
%     \subfigure{
%         \includegraphics[width=0.3\textwidth]{images/time_complex_forward.png}
%     }
%     \vspace{-0.02cm}
%     \subfigure{
%         \includegraphics[width=0.3\textwidth]{images/time_complex_backward.png}
%     }
%     % \vspace{-0.5cm}
%     \subfigure{
%         \includegraphics[width=0.3\textwidth]{images/time_complex_inf.png}
%     }
%     \caption{Computational time for GATr and Transformer architectures.}
%     % GATr has highest time in all scenarios. At inference, augmented Transformer matches standard Transformer.}}
%     \label{fig:Time complexity}
% \end{wrapfigure}

In all comparisons, GATr architecture consistently required the highest time, being approximately ten times slower than Transformer architecture. This significant difference can be attributed to the calculations of multivectors in GATr's design. In the combined forward and backward passes, the addition of the equivariance loss increases the computation time of the standard Transformer as we calculate two model outputs at each step. However, it's still around $2.5 \times$ faster than GATr, in the worst case of a gradual penalty. Furthermore, GATr reached its memory capacity earlier, hitting an out-of-memory issue at a batch size of $2^{11}$. During inference, the computational speed for the Transformer trained with equivariance loss or data augmentation matches the standard Transformer, which results in an inference speed that is $10 \times$ faster than GATr.
Notably, while we include GATr as our equivariant baseline, GATr itself is computationally more efficient than many equivariant architectures such as SE(3)-Transformer and SEGNN, as indicated in \cite{brehmer2023geometric}.
\section{Conclusion}
\label{conclusion}
We introduced REMUL, a simple and effective method for learning \emph{approximately} equivariant functions using unconstrained architectures. By formulating equivariance as an explicit, tunable objective within a multitask learning framework, REMUL relaxes the often costly and rigid constraints of traditional equivariant models. We demonstrated empirically that unconstrained networks trained with REMUL can learn appropriate levels of symmetry, controlled by a hyperparameter $\beta$. This allows us to balance the benefits of the equivariance inductive bias against task-specific requirements and computational costs. Our method achieves competitive performance compared to constrained baselines on various geometric tasks, while offering significant speed advantages (up to $10 \times$ faster inference, 2.5$\times$ faster training).

\textbf{Limitations and Future Directions.}
This work introduces a simple approach for understanding and analyzing unconstrained versus equivariant models, which significantly impact the field by enabling broader applicability and easier integration into existing frameworks. Building on these foundations, numerous additional ideas for extending our study present exciting opportunities for future research.
% Numerous additional ideas for extending our study offer exciting opportunities for future research. 
For instance, as we noted earlier,  $\alpha$ and $\beta$ serve as additional hyperparameters that could be constant or automatically updated with GradNorm algorithm, we could explore more efficient learnable weights, such as \cite{crawshaw2021slawscaledlossapproximate, bohn2024taskweightinggradientprojection}. Another promising avenue is applying our method during the fine-tuning phase when leveraging pre-trained models for tasks that require equivariance \cite{zaidi2023pretraining, ni2024pretrainingfractionaldenoisingenhance}. 
On the other side, further analysis is required to understand the theoretical guarantees of approximate equivariance offered by REMUL, such as how relaxing equivariance constraints affects the model’s generalization bounds \cite{petrache2023approximationgeneralizationtradeoffsapproximategroup}.

\subsubsection*{Acknowledgments}
The authors would like to thank Joey Bose, Jacob Bamberger, Xingyue Huang, Emily Jin, Katarina Petrovic, and Scott Le Roux for their valuable comments on the early versions of the manuscript.  
This research is partially supported by EPSRC Turing AI World-Leading Research Fellowship No. EP/X040062/1, EPSRC AI Hub on Mathematical Foundations of Intelligence: An ``Erlangen Programme'' for AI No. EP/Y028872/1, and the Postdoc.Mobility grant P500PT-217915 from the Swiss National Science Foundation.

\bibliographystyle{unsrtnat}
\bibliography{reference}

\begin{thebibliography}{101}
\providecommand{\natexlab}[1]{#1}
\providecommand{\url}[1]{\texttt{#1}}
\expandafter\ifx\csname urlstyle\endcsname\relax
  \providecommand{\doi}[1]{doi: #1}\else
  \providecommand{\doi}{doi: \begingroup \urlstyle{rm}\Url}\fi

\bibitem[Weiler et~al.(2018)Weiler, Hamprecht, and Storath]{weiler2018learningsteerablefiltersrotation}
Maurice Weiler, Fred~A. Hamprecht, and Martin Storath.
\newblock Learning steerable filters for rotation equivariant cnns.
\newblock In \emph{2018 IEEE/CVF Conference on Computer Vision and Pattern Recognition}, pages 849--858, 2018.
\newblock \doi{10.1109/CVPR.2018.00095}.

\bibitem[Yu et~al.(2022)Yu, Wu, and Yi]{yu2022rotationallyequivariant3dobject}
Hong-Xing Yu, Jiajun Wu, and Li~Yi.
\newblock Rotationally equivariant 3d object detection.
\newblock In \emph{2022 IEEE/CVF Conference on Computer Vision and Pattern Recognition (CVPR)}, pages 1446--1454, 2022.
\newblock \doi{10.1109/CVPR52688.2022.00151}.

\bibitem[Han et~al.(2022)Han, Huang, Xu, and Rong]{han2022equivariantgraphhierarchybasedneural}
Jiaqi Han, Wenbing Huang, Tingyang Xu, and Yu~Rong.
\newblock Equivariant graph hierarchy-based neural networks.
\newblock In Alice~H. Oh, Alekh Agarwal, Danielle Belgrave, and Kyunghyun Cho, editors, \emph{Advances in Neural Information Processing Systems}, 2022.
\newblock URL \url{https://openreview.net/forum?id=ywxtmG1nU_6}.

\bibitem[Xu et~al.(2024)Xu, Han, Lou, Kossaifi, Ramanathan, Azizzadenesheli, Leskovec, Ermon, and Anandkumar]{xu2024equivariantgraphneuraloperator}
Minkai Xu, Jiaqi Han, Aaron Lou, Jean Kossaifi, Arvind Ramanathan, Kamyar Azizzadenesheli, Jure Leskovec, Stefano Ermon, and Anima Anandkumar.
\newblock Equivariant graph neural operator for modeling 3d dynamics.
\newblock In \emph{Proceedings of the 41st International Conference on Machine Learning}, 2024.

\bibitem[Satorras et~al.(2021)Satorras, Hoogeboom, and Welling]{satorras2022en}
Victor~Garcia Satorras, Emiel Hoogeboom, and Max Welling.
\newblock E(n) equivariant graph neural networks.
\newblock In \emph{Proceedings of the 38rd International Conference on Machine Learning}, 2021.

\bibitem[Brandstetter et~al.(2022)Brandstetter, Hesselink, van~der Pol, Bekkers, and Welling]{brandstetter2022geometric}
Johannes Brandstetter, Rob Hesselink, Elise van~der Pol, Erik~J Bekkers, and Max Welling.
\newblock Geometric and physical quantities improve e(3) equivariant message passing.
\newblock In \emph{International Conference on Learning Representations}, 2022.
\newblock URL \url{https://openreview.net/forum?id=_xwr8gOBeV1}.

\bibitem[Jumper et~al.(2021)Jumper, Evans, Pritzel, Green, Figurnov, Ronneberger, Tunyasuvunakool, Bates, Žídek, Potapenko, Bridgland, Meyer, Kohl, Ballard, Cowie, Romera-Paredes, Nikolov, Jain, Adler, Back, Petersen, Reiman, Clancy, Zielinski, Steinegger, Pacholska, Berghammer, Bodenstein, Silver, Vinyals, Senior, Kavukcuoglu, Kohli, and Hassabis]{Jumper2021}
John Jumper, Richard Evans, Alexander Pritzel, Tim Green, Michael Figurnov, Olaf Ronneberger, Kathryn Tunyasuvunakool, Russ Bates, Augustin Žídek, Anna Potapenko, Alex Bridgland, Clemens Meyer, Simon A.~A. Kohl, Andrew~J. Ballard, Andrew Cowie, Bernardino Romera-Paredes, Stanislav Nikolov, Rishub Jain, Jonas Adler, Trevor Back, Stig Petersen, David Reiman, Ellen Clancy, Michal Zielinski, Martin Steinegger, Michalina Pacholska, Tamas Berghammer, Sebastian Bodenstein, David Silver, Oriol Vinyals, Andrew~W. Senior, Koray Kavukcuoglu, Pushmeet Kohli, and Demis Hassabis.
\newblock Highly accurate protein structure prediction with alphafold.
\newblock \emph{Nature}, 596\penalty0 (7873):\penalty0 583--589, 2021.
\newblock \doi{10.1038/s41586-021-03819-2}.
\newblock URL \url{https://doi.org/10.1038/s41586-021-03819-2}.

\bibitem[Sch{\"u}tt et~al.(2021)Sch{\"u}tt, Unke, and Gastegger]{schütt2021equivariantmessagepassingprediction}
Kristof Sch{\"u}tt, Oliver Unke, and Michael Gastegger.
\newblock Equivariant message passing for the prediction of tensorial properties and molecular spectra.
\newblock In \emph{Proceedings of the 38th International Conference on Machine Learning}, 2021.
\newblock URL \url{https://proceedings.mlr.press/v139/schutt21a.html}.

\bibitem[Bronstein et~al.(2021)Bronstein, Bruna, Cohen, and Veličković]{bronstein2021geometricdeeplearninggrids}
Michael~M. Bronstein, Joan Bruna, Taco Cohen, and Petar Veličković.
\newblock Geometric deep learning: Grids, groups, graphs, geodesics, and gauges, 2021.
\newblock URL \url{https://arxiv.org/abs/2104.13478}.

\bibitem[Th{\"o}lke and Fabritiis(2022)]{thölke2022torchmdnetequivarianttransformersneural}
Philipp Th{\"o}lke and Gianni~De Fabritiis.
\newblock Equivariant transformers for neural network based molecular potentials.
\newblock In \emph{International Conference on Learning Representations}, 2022.
\newblock URL \url{https://openreview.net/forum?id=zNHzqZ9wrRB}.

\bibitem[Liao et~al.(2024)Liao, Wood, Das*, and Smidt*]{liao2024equiformerv2improvedequivarianttransformer}
Yi-Lun Liao, Brandon Wood, Abhishek Das*, and Tess Smidt*.
\newblock {EquiformerV2: Improved Equivariant Transformer for Scaling to Higher-Degree Representations}.
\newblock In \emph{International Conference on Learning Representations (ICLR)}, 2024.
\newblock URL \url{https://openreview.net/forum?id=mCOBKZmrzD}.

\bibitem[Batzner et~al.(2022)Batzner, Musaelian, Sun, et~al.]{Batzner2022Equivariant}
Simon Batzner, Albert Musaelian, Lixin Sun, et~al.
\newblock E(3)-equivariant graph neural networks for data-efficient and accurate interatomic potentials.
\newblock \emph{Nature Communications}, 13:\penalty0 2453, 2022.
\newblock \doi{10.1038/s41467-022-29939-5}.
\newblock URL \url{https://doi.org/10.1038/s41467-022-29939-5}.

\bibitem[Batatia et~al.(2022)Batatia, Kovacs, Simm, Ortner, and Csanyi]{batatia2023macehigherorderequivariant}
Ilyes Batatia, David~Peter Kovacs, Gregor N.~C. Simm, Christoph Ortner, and Gabor Csanyi.
\newblock {MACE}: Higher order equivariant message passing neural networks for fast and accurate force fields.
\newblock In Alice~H. Oh, Alekh Agarwal, Danielle Belgrave, and Kyunghyun Cho, editors, \emph{Advances in Neural Information Processing Systems}, 2022.
\newblock URL \url{https://openreview.net/forum?id=YPpSngE-ZU}.

\bibitem[Thomas et~al.(2018)Thomas, Smidt, Kearnes, Yang, Li, Kohlhoff, and Riley]{thomas2018tensorfieldnetworksrotation}
Nathaniel Thomas, Tess Smidt, Steven Kearnes, Lusann Yang, Li~Li, Kai Kohlhoff, and Patrick Riley.
\newblock Tensor field networks: Rotation- and translation-equivariant neural networks for 3d point clouds, 2018.
\newblock URL \url{https://arxiv.org/abs/1802.08219}.

\bibitem[Fuchs et~al.(2020)Fuchs, Worrall, Fischer, and Welling]{fuchs2020se3transformers}
Fabian~B. Fuchs, Daniel~E. Worrall, Volker Fischer, and Max Welling.
\newblock Se(3)-transformers: 3d roto-translation equivariant attention networks.
\newblock In \emph{Advances in Neural Information Processing Systems 34 (NeurIPS)}, 2020.

\bibitem[Liao and Smidt(2023)]{liao2023equiformerequivariantgraphattention}
Yi-Lun Liao and Tess Smidt.
\newblock Equiformer: Equivariant graph attention transformer for 3d atomistic graphs.
\newblock In \emph{International Conference on Learning Representations}, 2023.
\newblock URL \url{https://openreview.net/forum?id=KwmPfARgOTD}.

\bibitem[Luo et~al.(2024)Luo, Chen, and Krishnapriyan]{luo2024enablingefficientequivariantoperations}
Shengjie Luo, Tianlang Chen, and Aditi~S. Krishnapriyan.
\newblock Enabling efficient equivariant operations in the fourier basis via gaunt tensor products, 2024.
\newblock URL \url{https://arxiv.org/abs/2401.10216}.

\bibitem[Joshi et~al.(2023)Joshi, Bodnar, Mathis, Cohen, and Lio]{pmlr-v202-joshi23a}
Chaitanya~K. Joshi, Cristian Bodnar, Simon~V Mathis, Taco Cohen, and Pietro Lio.
\newblock On the expressive power of geometric graph neural networks.
\newblock In \emph{Proceedings of the 40th International Conference on Machine Learning}, 2023.
\newblock URL \url{https://proceedings.mlr.press/v202/joshi23a.html}.

\bibitem[Kaba et~al.(2022)Kaba, Mondal, Zhang, Bengio, and Ravanbakhsh]{kaba2022equivariance}
S{\'e}kou-Oumar Kaba, Arnab~Kumar Mondal, Yan Zhang, Yoshua Bengio, and Siamak Ravanbakhsh.
\newblock Equivariance with learned canonicalization functions.
\newblock In \emph{NeurIPS 2022 Workshop on Symmetry and Geometry in Neural Representations}, 2022.
\newblock URL \url{https://openreview.net/forum?id=pVD1k8ge25a}.

\bibitem[Mondal et~al.(2023)Mondal, Panigrahi, Kaba, Rajeswar, and Ravanbakhsh]{mondal2023equivariant}
Arnab~Kumar Mondal, Siba~Smarak Panigrahi, S{\'e}kou-Oumar Kaba, Sai Rajeswar, and Siamak Ravanbakhsh.
\newblock Equivariant adaptation of large pretrained models.
\newblock In \emph{Thirty-seventh Conference on Neural Information Processing Systems}, 2023.
\newblock URL \url{https://openreview.net/forum?id=m6dRQJw280}.

\bibitem[Baker et~al.(2024)Baker, Wang, de~Fernex, and Wang]{baker2024an}
Justin Baker, Shih-Hsin Wang, Tommaso de~Fernex, and Bao Wang.
\newblock An explicit frame construction for normalizing 3d point clouds.
\newblock In \emph{Forty-first International Conference on Machine Learning}, 2024.
\newblock URL \url{https://openreview.net/forum?id=SZ0JnRxi0x}.

\bibitem[Ma et~al.(2024)Ma, Wang, Lim, Jegelka, and Wang]{ma2024canonizationperspectiveinvariantequivariant}
George Ma, Yifei Wang, Derek Lim, Stefanie Jegelka, and Yisen Wang.
\newblock A canonicalization perspective on invariant and equivariant learning.
\newblock In \emph{The Thirty-eighth Annual Conference on Neural Information Processing Systems}, 2024.
\newblock URL \url{https://openreview.net/forum?id=jjcY92FX4R}.

\bibitem[Panigrahi and Mondal(2024)]{panigrahi2024improved}
Siba~Smarak Panigrahi and Arnab~Kumar Mondal.
\newblock Improved canonicalization for model agnostic equivariance.
\newblock In \emph{CVPR 2024 Workshop on Equivariant Vision: From Theory to Practice}, 2024.
\newblock URL \url{https://arxiv.org/abs/2405.14089}.

\bibitem[Puny et~al.(2022)Puny, Atzmon, Smith, Misra, Grover, Ben-Hamu, and Lipman]{puny2022frameaveraginginvariantequivariant}
Omri Puny, Matan Atzmon, Edward~J. Smith, Ishan Misra, Aditya Grover, Heli Ben-Hamu, and Yaron Lipman.
\newblock Frame averaging for invariant and equivariant network design.
\newblock In \emph{International Conference on Learning Representations}, 2022.
\newblock URL \url{https://openreview.net/forum?id=zIUyj55nXR}.

\bibitem[Duval et~al.(2023)Duval, Schmidt, Hern\'{a}ndez-Garc\'{\i}a, Miret, Malliaros, Bengio, and Rolnick]{pmlr-v202-duval23a}
Alexandre~Agm Duval, Victor Schmidt, Alex Hern\'{a}ndez-Garc\'{\i}a, Santiago Miret, Fragkiskos~D. Malliaros, Yoshua Bengio, and David Rolnick.
\newblock {FAEN}et: Frame averaging equivariant {GNN} for materials modeling.
\newblock In \emph{Proceedings of the 40th International Conference on Machine Learning}, 2023.
\newblock URL \url{https://proceedings.mlr.press/v202/duval23a.html}.

\bibitem[Lin et~al.(2024)Lin, Helwig, Gui, and Ji]{lin2024equivariance}
Yuchao Lin, Jacob Helwig, Shurui Gui, and Shuiwang Ji.
\newblock Equivariance via minimal frame averaging for more symmetries and efficiency.
\newblock In \emph{Forty-first International Conference on Machine Learning}, 2024.
\newblock URL \url{https://openreview.net/forum?id=guFsTBXsov}.

\bibitem[Huang et~al.(2024)Huang, Song, Ying, and Jin]{huang2024proteinnucleicacidcomplexmodeling}
Tinglin Huang, Zhenqiao Song, Rex Ying, and Wengong Jin.
\newblock Protein-nucleic acid complex modeling with frame averaging transformer.
\newblock In \emph{The Thirty-eighth Annual Conference on Neural Information Processing Systems}, 2024.
\newblock URL \url{https://openreview.net/forum?id=Xngi3Z3wkN}.

\bibitem[Quiroga et~al.(2019)Quiroga, Ronchetti, Lanzarini, and Bariviera]{Quiroga_2019}
Facundo Quiroga, Franco Ronchetti, Laura Lanzarini, and Aurelio~F. Bariviera.
\newblock \emph{Revisiting Data Augmentation for Rotational Invariance in Convolutional Neural Networks}, page 127–141.
\newblock Springer International Publishing, March 2019.
\newblock ISBN 9783030154134.
\newblock \doi{10.1007/978-3-030-15413-4_10}.
\newblock URL \url{http://dx.doi.org/10.1007/978-3-030-15413-4_10}.

\bibitem[Benton et~al.(2020)Benton, Finzi, Izmailov, and Wilson]{benton2020learning}
Gregory Benton, Marc Finzi, Pavel Izmailov, and Andrew~Gordon Wilson.
\newblock Learning invariances in neural networks.
\newblock \emph{arXiv preprint arXiv:2010.11882}, 2020.

\bibitem[Liu et~al.(2021)Liu, Huang, Huang, and Wang]{9710479}
Aoming Liu, Zehao Huang, Zhiwu Huang, and Naiyan Wang.
\newblock Direct differentiable augmentation search.
\newblock In \emph{2021 IEEE/CVF International Conference on Computer Vision (ICCV)}, pages 12199--12208, 2021.
\newblock \doi{10.1109/ICCV48922.2021.01200}.

\bibitem[Bai et~al.(2021)Bai, Mei, Yuille, and Xie]{bai2021transformersrobustcnns}
Yutong Bai, Jieru Mei, Alan Yuille, and Cihang Xie.
\newblock Are transformers more robust than cnns?
\newblock In \emph{Thirty-Fifth Conference on Neural Information Processing Systems}, 2021.

\bibitem[Gerken et~al.(2022)Gerken, Carlsson, Linander, Ohlsson, Petersson, and Persson]{gerken2022}
Jan~E. Gerken, Oscar Carlsson, Hampus Linander, Fredrik Ohlsson, Christoffer Petersson, and Daniel Persson.
\newblock Equivariance versus {{Augmentation}} for {{Spherical Images}}.
\newblock In \emph{Proceedings of the 39th {{International Conference}} on {{Machine Learning}}}, pages 7404--7421. {PMLR}, 2022.
\newblock \doi{10.48550/arXiv.2202.03990}.

\bibitem[Iglesias et~al.(2023)Iglesias, Talavera, González-Prieto, Mozo, and Gómez-Canaval]{Iglesias_2023}
Guillermo Iglesias, Edgar Talavera, Ángel González-Prieto, Alberto Mozo, and Sandra Gómez-Canaval.
\newblock Data augmentation techniques in time series domain: a survey and taxonomy.
\newblock \emph{Neural Computing and Applications}, 35\penalty0 (14):\penalty0 10123–10145, March 2023.
\newblock ISSN 1433-3058.
\newblock \doi{10.1007/s00521-023-08459-3}.
\newblock URL \url{http://dx.doi.org/10.1007/s00521-023-08459-3}.

\bibitem[Chatzipantazis et~al.(2023)Chatzipantazis, Pertigkiozoglou, Daniilidis, and Dobriban]{chatzipantazis2023learning}
Evangelos Chatzipantazis, Stefanos Pertigkiozoglou, Kostas Daniilidis, and Edgar Dobriban.
\newblock Learning augmentation distributions using transformed risk minimization.
\newblock \emph{Transactions on Machine Learning Research}, 2023.
\newblock ISSN 2835-8856.
\newblock URL \url{https://openreview.net/forum?id=LRYtNj8Xw0}.

\bibitem[Xu et~al.(2023)Xu, Yoon, Fuentes, and Park]{Xu2023ImageAugmentation}
Mingle Xu, Sook Yoon, Alvaro Fuentes, and Dong~Sun Park.
\newblock A comprehensive survey of image augmentation techniques for deep learning.
\newblock \emph{Pattern Recognition}, 137:\penalty0 109347, 2023.
\newblock ISSN 0031-3203.
\newblock \doi{10.1016/j.patcog.2023.109347}.
\newblock URL \url{https://www.sciencedirect.com/science/article/pii/S0031320323000481}.

\bibitem[Yang et~al.(2024)Yang, Guo, Zhao, and Shen]{Yang2024DataAugmentation}
Suorong Yang, Suhan Guo, Jian Zhao, and Furao Shen.
\newblock Investigating the effectiveness of data augmentation from similarity and diversity: An empirical study.
\newblock \emph{Pattern Recognition}, 148:\penalty0 110204, 2024.
\newblock ISSN 0031-3203.
\newblock \doi{10.1016/j.patcog.2023.110204}.
\newblock URL \url{https://www.sciencedirect.com/science/article/pii/S0031320323009019}.

\bibitem[Wang et~al.(2024)Wang, Elhag, Jaitly, Susskind, and Bautista]{wang2024swallowingbitterpillsimplified}
Yuyang Wang, Ahmed~A. Elhag, Navdeep Jaitly, Joshua~M. Susskind, and Miguel~{\'A}ngel Bautista.
\newblock Swallowing the bitter pill: Simplified scalable conformer generation.
\newblock In \emph{Forty-first International Conference on Machine Learning}, 2024.

\bibitem[Abramson et~al.(2024)Abramson, Adler, Dunger, Evans, Green, Pritzel, Ronneberger, Willmore, Ballard, Bambrick, Bodenstein, Evans, Hung, O’Neill, Reiman, Tunyasuvunakool, Wu, Žemgulytė, Arvaniti, Beattie, Bertolli, Bridgland, Cherepanov, Congreve, Cowen-Rivers, Cowie, Figurnov, Fuchs, Gladman, Jain, Khan, Low, Perlin, Potapenko, Savy, Singh, Stecula, Thillaisundaram, Tong, Yakneen, Zhong, Zielinski, Žídek, Bapst, Kohli, Jaderberg, Hassabis, and Jumper]{Abramson2024}
Josh Abramson, Jonas Adler, Jack Dunger, Richard Evans, Tim Green, Alexander Pritzel, Olaf Ronneberger, Lindsay Willmore, Andrew~J. Ballard, Joshua Bambrick, Sebastian~W. Bodenstein, David~A. Evans, Chia-Chun Hung, Michael O’Neill, David Reiman, Kathryn Tunyasuvunakool, Zachary Wu, Akvilė Žemgulytė, Eirini Arvaniti, Charles Beattie, Ottavia Bertolli, Alex Bridgland, Alexey Cherepanov, Miles Congreve, Alexander~I. Cowen-Rivers, Andrew Cowie, Michael Figurnov, Fabian~B. Fuchs, Hannah Gladman, Rishub Jain, Yousuf~A. Khan, Caroline M.~R. Low, Kuba Perlin, Anna Potapenko, Pascal Savy, Sukhdeep Singh, Adrian Stecula, Ashok Thillaisundaram, Catherine Tong, Sergei Yakneen, Ellen~D. Zhong, Michal Zielinski, Augustin Žídek, Victor Bapst, Pushmeet Kohli, Max Jaderberg, Demis Hassabis, and John~M. Jumper.
\newblock Accurate structure prediction of biomolecular interactions with alphafold 3.
\newblock \emph{Nature}, 630\penalty0 (8016):\penalty0 493--500, 2024.
\newblock \doi{10.1038/s41586-024-07487-w}.

\bibitem[Baek et~al.(2017)Baek, Kafri, and Lecomte]{Baek_2017}
Yongjoo Baek, Yariv Kafri, and Vivien Lecomte.
\newblock Dynamical symmetry breaking and phase transitions in driven diffusive systems.
\newblock \emph{Physical Review Letters}, 118\penalty0 (3), January 2017.
\newblock ISSN 1079-7114.
\newblock \doi{10.1103/physrevlett.118.030604}.
\newblock URL \url{http://dx.doi.org/10.1103/PhysRevLett.118.030604}.

\bibitem[Weidinger et~al.(2017)Weidinger, Heyl, Silva, and Knap]{Weidinger_2017}
Simon~A. Weidinger, Markus Heyl, Alessandro Silva, and Michael Knap.
\newblock Dynamical quantum phase transitions in systems with continuous symmetry breaking.
\newblock \emph{Physical Review B}, 96\penalty0 (13), October 2017.
\newblock ISSN 2469-9969.
\newblock \doi{10.1103/physrevb.96.134313}.
\newblock URL \url{http://dx.doi.org/10.1103/PhysRevB.96.134313}.

\bibitem[Gibb et~al.(2024)Gibb, Hobbs, Nikolova, Raistrick, Berrow, Mertelj, Osterman, Sebastián, Gleeson, and Mandle]{Gibb_2024}
Calum~J. Gibb, Jordan Hobbs, Diana~I. Nikolova, Thomas Raistrick, Stuart~R. Berrow, Alenka Mertelj, Natan Osterman, Nerea Sebastián, Helen~F. Gleeson, and Richard.~J. Mandle.
\newblock Spontaneous symmetry breaking in polar fluids.
\newblock \emph{Nature Communications}, 15\penalty0 (1), July 2024.
\newblock ISSN 2041-1723.
\newblock \doi{10.1038/s41467-024-50230-2}.
\newblock URL \url{http://dx.doi.org/10.1038/s41467-024-50230-2}.

\bibitem[Yannouleas and Landman(2000)]{Yannouleas_2000}
Constantine Yannouleas and Uzi Landman.
\newblock Erratum: Spontaneous symmetry breaking in single and molecular quantum dots [phys. rev. lett. 82, 5325 (1999)].
\newblock \emph{Physical Review Letters}, 85\penalty0 (10):\penalty0 2220–2220, September 2000.
\newblock ISSN 1079-7114.
\newblock \doi{10.1103/physrevlett.85.2220}.
\newblock URL \url{http://dx.doi.org/10.1103/PhysRevLett.85.2220}.

\bibitem[Goehring et~al.(2011)Goehring, Trong, Bois, Chowdhury, Nicola, Hyman, and Grill]{Goehring2011Polarization}
Nathan~W. Goehring, Philipp~Khuc Trong, Justin~S. Bois, Debanjan Chowdhury, Ernesto~M. Nicola, Anthony~A. Hyman, and Stephan~W. Grill.
\newblock Polarization of {PAR} proteins by advective triggering of a pattern-forming system.
\newblock \emph{Science}, 334:\penalty0 1137--1141, 2011.
\newblock \doi{10.1126/science.1208619}.
\newblock Epub 2011 Oct 20.

\bibitem[Mietke et~al.(2019)Mietke, Jemseena, Kumar, Sbalzarini, and J\"ulicher]{mietke2019minimalmodelcellularsymmetry}
Alexander Mietke, V.~Jemseena, K.~Vijay Kumar, Ivo~F. Sbalzarini, and Frank J\"ulicher.
\newblock Minimal model of cellular symmetry breaking.
\newblock \emph{Phys. Rev. Lett.}, 123:\penalty0 188101, Oct 2019.
\newblock \doi{10.1103/PhysRevLett.123.188101}.
\newblock URL \url{https://link.aps.org/doi/10.1103/PhysRevLett.123.188101}.

\bibitem[Brehmer et~al.(2023)Brehmer, Haan, Behrends, and Cohen]{brehmer2023geometric}
Johann Brehmer, Pim~De Haan, S{\"o}nke Behrends, and Taco Cohen.
\newblock Geometric algebra transformer.
\newblock In \emph{Thirty-seventh Conference on Neural Information Processing Systems}, 2023.
\newblock URL \url{https://openreview.net/forum?id=M7r2CO4tJC}.

\bibitem[Wang et~al.(2022)Wang, Walters, and Yu]{wang2022approximatelyequivariantnetworksimperfectly}
Rui Wang, Robin Walters, and Rose Yu.
\newblock Approximately equivariant networks for imperfectly symmetric dynamics.
\newblock In \emph{Proceedings of the 39th International Conference on Machine Learning}, 2022.

\bibitem[Petrache and Trivedi(2023)]{petrache2023approximationgeneralizationtradeoffsapproximategroup}
Mircea Petrache and Shubhendu Trivedi.
\newblock Approximation-generalization trade-offs under (approximate) group equivariance.
\newblock In \emph{Advances in Neural Information Processing Systems}, 2023.

\bibitem[Kufel et~al.(2024)Kufel, Kemp, Linsel, Laumann, and Yao]{kufel2024approximatelysymmetricneuralnetworksquantum}
Dominik~S. Kufel, Jack Kemp, Simon~M. Linsel, Chris~R. Laumann, and Norman~Y. Yao.
\newblock Approximately-symmetric neural networks for quantum spin liquids, 2024.
\newblock URL \url{https://arxiv.org/abs/2405.17541}.

\bibitem[Ashman et~al.(2024)Ashman, Diaconu, Weller, Bruinsma, and Turner]{ashman2024approximatelyequivariantneuralprocesses}
Matthew Ashman, Cristiana Diaconu, Adrian Weller, Wessel Bruinsma, and Richard~E. Turner.
\newblock Approximately equivariant neural processes, 2024.
\newblock URL \url{https://arxiv.org/abs/2406.13488}.

\bibitem[Chen et~al.(2018)Chen, Badrinarayanan, Lee, and Rabinovich]{chen2018gradnormgradientnormalizationadaptive}
Zhao Chen, Vijay Badrinarayanan, Chen-Yu Lee, and Andrew Rabinovich.
\newblock {G}rad{N}orm: Gradient normalization for adaptive loss balancing in deep multitask networks.
\newblock In \emph{Proceedings of the 35th International Conference on Machine Learning}, 2018.
\newblock URL \url{https://proceedings.mlr.press/v80/chen18a.html}.

\bibitem[Lyle et~al.(2020)Lyle, van~der Wilk, Kwiatkowska, Gal, and Bloem-Reddy]{lyle2020benefitsinvarianceneuralnetworks}
Clare Lyle, Mark van~der Wilk, Marta Kwiatkowska, Yarin Gal, and Benjamin Bloem-Reddy.
\newblock On the benefits of invariance in neural networks, 2020.
\newblock URL \url{https://arxiv.org/abs/2005.00178}.

\bibitem[Kvinge et~al.(2022)Kvinge, Emerson, Jorgenson, Vasquez, Doster, and Lew]{kvinge2022waysdeepneuralnetworks}
Henry Kvinge, Tegan Emerson, Grayson Jorgenson, Scott Vasquez, Timothy Doster, and Jesse Lew.
\newblock In what ways are deep neural networks invariant and how should we measure this?
\newblock In Alice~H. Oh, Alekh Agarwal, Danielle Belgrave, and Kyunghyun Cho, editors, \emph{Advances in Neural Information Processing Systems}, 2022.
\newblock URL \url{https://openreview.net/forum?id=SCD0hn3kMHw}.

\bibitem[Moskalev et~al.(2023)Moskalev, Sepliarskaia, Bekkers, and Smeulders]{moskalev2023genuineinvariancelearningweighttying}
Artem Moskalev, Anna Sepliarskaia, Erik~J. Bekkers, and Arnold Smeulders.
\newblock On genuine invariance learning without weight-tying.
\newblock In \emph{Proceedings of the 40th International Conference on Machine Learning}, 2023.

\bibitem[Gruver et~al.(2023)Gruver, Finzi, Goldblum, and Wilson]{gruver2024liederivativemeasuringlearned}
Nate Gruver, Marc~Anton Finzi, Micah Goldblum, and Andrew~Gordon Wilson.
\newblock The lie derivative for measuring learned equivariance.
\newblock In \emph{The Eleventh International Conference on Learning Representations}, 2023.
\newblock URL \url{https://openreview.net/forum?id=JL7Va5Vy15J}.

\bibitem[Speicher et~al.(2024)Speicher, Nanda, and Gummadi]{speicher2024understandingroleinvariancetransfer}
Till Speicher, Vedant Nanda, and Krishna~P. Gummadi.
\newblock Understanding the role of invariance in transfer learning.
\newblock \emph{Transactions on Machine Learning Research}, 2024.
\newblock ISSN 2835-8856.
\newblock URL \url{https://openreview.net/forum?id=spJI4LSPIU}.

\bibitem[Cohen and Welling(2016)]{cohen2016group}
Taco Cohen and Max Welling.
\newblock Group equivariant convolutional networks.
\newblock In \emph{Proceedings of the 33rd International Conference on Machine Learning}, 2016.
\newblock URL \url{https://proceedings.mlr.press/v48/cohenc16.html}.

\bibitem[Cohen et~al.(2019)Cohen, Weiler, Kicanaoglu, and Welling]{cohen2019gaugeequivariantconvolutionalnetworks}
Taco~S. Cohen, Maurice Weiler, Berkay Kicanaoglu, and Max Welling.
\newblock Gauge equivariant convolutional networks and the icosahedral cnn.
\newblock In \emph{Proceedings of the 36th International Conference on Machine Learning}, 2019.

\bibitem[Weiler and Cesa(2019)]{weiler2021generale2equivariantsteerablecnns}
Maurice Weiler and Gabriele Cesa.
\newblock {General E(2)-Equivariant Steerable CNNs}.
\newblock In \emph{Conference on Neural Information Processing Systems (NeurIPS)}, 2019.

\bibitem[Qiao et~al.(2023)Qiao, Xu, and Li]{qiao2023scalerotationequivariantliegroupconvolution}
Wei-Dong Qiao, Yang Xu, and Hui Li.
\newblock Scale-rotation-equivariant lie group convolution neural networks (lie group-cnns), 2023.
\newblock URL \url{https://arxiv.org/abs/2306.06934}.

\bibitem[Chen et~al.(2021{\natexlab{a}})Chen, Liu, Chen, Li, and Hill]{chen2021equivariantpointnetwork3d}
Haiwei Chen, Shichen Liu, Weikai Chen, Hao Li, and Randall Hill.
\newblock Equivariant point network for 3d point cloud analysis.
\newblock In \emph{Proceedings of the IEEE/CVF Conference on Computer Vision and Pattern Recognition}, 2021{\natexlab{a}}.

\bibitem[Gasteiger et~al.(2020)Gasteiger, Groß, and Günnemann]{gasteiger2022directionalmessagepassingmolecular}
Johannes Gasteiger, Janek Groß, and Stephan Günnemann.
\newblock Directional message passing for molecular graphs.
\newblock In \emph{International Conference on Learning Representations}, 2020.
\newblock URL \url{https://openreview.net/forum?id=B1eWbxStPH}.

\bibitem[Perez and Wang(2017)]{perez2017effectivenessdataaugmentationimage}
Luis Perez and Jason Wang.
\newblock The effectiveness of data augmentation in image classification using deep learning, 2017.
\newblock URL \url{https://arxiv.org/abs/1712.04621}.

\bibitem[Inoue(2018)]{inoue2018dataaugmentationpairingsamples}
Hiroshi Inoue.
\newblock Data augmentation by pairing samples for images classification, 2018.
\newblock URL \url{https://arxiv.org/abs/1801.02929}.

\bibitem[Rahat et~al.(2024)Rahat, Hossain, Ahmed, Jha, and Ewetz]{rahat2024dataaugmentationimageclassification}
Fazle Rahat, M~Shifat Hossain, Md~Rubel Ahmed, Sumit~Kumar Jha, and Rickard Ewetz.
\newblock Data augmentation for image classification using generative ai, 2024.
\newblock URL \url{https://arxiv.org/abs/2409.00547}.

\bibitem[Zoph et~al.(2020)Zoph, Cubuk, Ghiasi, Lin, Shlens, and Le]{zoph2019learningdataaugmentationstrategies}
Barret Zoph, Ekin~D. Cubuk, Golnaz Ghiasi, Tsung-Yi Lin, Jonathon Shlens, and Quoc~V. Le.
\newblock Learning data augmentation strategies for object detection.
\newblock In \emph{Computer Vision -- ECCV 2020}, 2020.

\bibitem[Wang et~al.(2019)Wang, Wang, Yang, Zhang, and Zuo]{wang2019dataaugmentationobjectdetection}
Hao Wang, Qilong Wang, Fan Yang, Weiqi Zhang, and Wangmeng Zuo.
\newblock Data augmentation for object detection via progressive and selective instance-switching, 2019.
\newblock URL \url{https://arxiv.org/abs/1906.00358}.

\bibitem[Kisantal et~al.(2019)Kisantal, Wojna, Murawski, Naruniec, and Cho]{kisantal2019augmentationsmallobjectdetection}
Mate Kisantal, Zbigniew Wojna, Jakub Murawski, Jacek Naruniec, and Kyunghyun Cho.
\newblock Augmentation for small object detection, 2019.
\newblock URL \url{https://arxiv.org/abs/1902.07296}.

\bibitem[Negassi et~al.(2022)Negassi, Wagner, and Reiterer]{negassi2021smartsamplingaugmentoptimalefficientdata}
Misgana Negassi, Diane Wagner, and Alexander Reiterer.
\newblock Smart(sampling)augment: Optimal and efficient data augmentation for semantic segmentation.
\newblock \emph{Algorithms}, 15\penalty0 (5), 2022.
\newblock ISSN 1999-4893.
\newblock \doi{10.3390/a15050165}.
\newblock URL \url{https://www.mdpi.com/1999-4893/15/5/165}.

\bibitem[Chen et~al.(2021{\natexlab{b}})Chen, Lian, Wang, Deng, Kuang, Fung, Gateno, Shen, Xia, and Yap]{chen2021diverse}
X~Chen, C~Lian, L~Wang, H~Deng, T~Kuang, SH~Fung, J~Gateno, D~Shen, JJ~Xia, and PT~Yap.
\newblock Diverse data augmentation for learning image segmentation with cross-modality annotations.
\newblock \emph{Medical Image Analysis}, 71:\penalty0 102060, 2021{\natexlab{b}}.
\newblock \doi{10.1016/j.media.2021.102060}.
\newblock Epub 2021 Apr 20. PMID: 33957558; PMCID: PMC8184609.

\bibitem[Yu et~al.(2023)Yu, Li, Lou, Liu, Wan, Chen, and Li]{yu2024diffusionbaseddataaugmentationnuclei}
Xinyi Yu, Guanbin Li, Wei Lou, Siqi Liu, Xiang Wan, Yan Chen, and Haofeng Li.
\newblock Diffusion-based data augmentation for nuclei image segmentation.
\newblock In \emph{Medical Image Computing and Computer Assisted Intervention -- MICCAI 2023}, 2023.

\bibitem[Hu et~al.(2021)Hu, Shuaibi, Das, Goyal, Sriram, Leskovec, Parikh, and Zitnick]{hu2021forcenetgraphneuralnetwork}
Weihua Hu, Muhammed Shuaibi, Abhishek Das, Siddharth Goyal, Anuroop Sriram, Jure Leskovec, Devi Parikh, and C.~Lawrence Zitnick.
\newblock Forcenet: A graph neural network for large-scale quantum calculations, 2021.
\newblock URL \url{https://arxiv.org/abs/2103.01436}.

\bibitem[Dosovitskiy et~al.(2021)Dosovitskiy, Beyer, Kolesnikov, Weissenborn, Zhai, Unterthiner, Dehghani, Minderer, Heigold, Gelly, Uszkoreit, and Houlsby]{dosovitskiy2021an}
Alexey Dosovitskiy, Lucas Beyer, Alexander Kolesnikov, Dirk Weissenborn, Xiaohua Zhai, Thomas Unterthiner, Mostafa Dehghani, Matthias Minderer, Georg Heigold, Sylvain Gelly, Jakob Uszkoreit, and Neil Houlsby.
\newblock An image is worth 16x16 words: Transformers for image recognition at scale.
\newblock In \emph{International Conference on Learning Representations}, 2021.
\newblock URL \url{https://openreview.net/forum?id=YicbFdNTTy}.

\bibitem[van~der Ouderaa et~al.(2022)van~der Ouderaa, Romero, and van~der Wilk]{vanderouderaa2022relaxingequivarianceconstraintsnonstationary}
Tycho~F.A. van~der Ouderaa, David~W. Romero, and Mark van~der Wilk.
\newblock Relaxing equivariance constraints with non-stationary continuous filters.
\newblock In \emph{Advances in Neural Information Processing Systems}, 2022.
\newblock URL \url{https://openreview.net/forum?id=5oEk8fvJxny}.

\bibitem[Romero and Lohit(2022)]{romero2022learning}
David~W. Romero and Suhas Lohit.
\newblock Learning partial equivariances from data.
\newblock In Alice~H. Oh, Alekh Agarwal, Danielle Belgrave, and Kyunghyun Cho, editors, \emph{Advances in Neural Information Processing Systems}, 2022.
\newblock URL \url{https://openreview.net/forum?id=pNHT6oBaPr8}.

\bibitem[Veefkind and Cesa(2024)]{veefkind2024probabilisticapproachlearningdegree}
Lars Veefkind and Gabriele Cesa.
\newblock A probabilistic approach to learning the degree of equivariance in steerable cnns, 2024.
\newblock URL \url{https://arxiv.org/abs/2406.03946}.

\bibitem[Wu et~al.(2024)Wu, Liu, Dong, Tang, Yang, Jin, Chen, and Wei]{wu2024sbdetsymmetrybreakingobjectdetector}
Zhiqiang Wu, Yingjie Liu, Hanlin Dong, Xuan Tang, Jian Yang, Bo~Jin, Mingsong Chen, and Xian Wei.
\newblock Sbdet: A symmetry-breaking object detector via relaxed rotation-equivariance, 2024.
\newblock URL \url{https://arxiv.org/abs/2408.11760}.

\bibitem[McNeela(2023)]{mcneela2023almost}
Daniel McNeela.
\newblock Almost equivariance via lie algebra convolutions.
\newblock In \emph{NeurIPS 2023 Workshop on Symmetry and Geometry in Neural Representations}, 2023.
\newblock URL \url{https://openreview.net/forum?id=2sLBXyVsPE}.

\bibitem[van~der Ouderaa et~al.(2023)van~der Ouderaa, Immer, and van~der Wilk]{vanderouderaa2023learninglayerwiseequivariancesautomatically}
Tycho~F.A. van~der Ouderaa, Alexander Immer, and Mark van~der Wilk.
\newblock Learning layer-wise equivariances automatically using gradients.
\newblock In \emph{Thirty-seventh Conference on Neural Information Processing Systems}, 2023.
\newblock URL \url{https://openreview.net/forum?id=bNIHdyunFC}.

\bibitem[Yeh et~al.(2022)Yeh, Hu, Hasegawa-Johnson, and Schwing]{pmlr-v151-yeh22b}
Raymond~A. Yeh, Yuan-Ting Hu, Mark Hasegawa-Johnson, and Alexander Schwing.
\newblock Equivariance discovery by learned parameter-sharing.
\newblock In \emph{Proceedings of The 25th International Conference on Artificial Intelligence and Statistics}, 2022.
\newblock URL \url{https://proceedings.mlr.press/v151/yeh22b.html}.

\bibitem[Finzi et~al.(2021{\natexlab{a}})Finzi, Benton, and Wilson]{NEURIPS2021_fc394e99}
Marc Finzi, Gregory Benton, and Andrew~G Wilson.
\newblock Residual pathway priors for soft equivariance constraints.
\newblock In \emph{Advances in Neural Information Processing Systems}, 2021{\natexlab{a}}.
\newblock URL \url{https://proceedings.neurips.cc/paper_files/paper/2021/file/fc394e9935fbd62c8aedc372464e1965-Paper.pdf}.

\bibitem[Pertigkiozoglou et~al.(2024)Pertigkiozoglou, Chatzipantazis, Trivedi, and Daniilidis]{pertigkiozoglou2024improvingequivariantmodeltraining}
Stefanos Pertigkiozoglou, Evangelos Chatzipantazis, Shubhendu Trivedi, and Kostas Daniilidis.
\newblock Improving equivariant model training via constraint relaxation.
\newblock In \emph{The Thirty-eighth Annual Conference on Neural Information Processing Systems}, 2024.
\newblock URL \url{https://openreview.net/forum?id=tWkL7k1u5v}.

\bibitem[Lawrence et~al.(2024)Lawrence, Portilheiro, Zhang, and Kaba]{lawrence2024improving}
Hannah Lawrence, Vasco Portilheiro, Yan Zhang, and S{\'e}kou-Oumar Kaba.
\newblock Improving equivariant networks with probabilistic symmetry breaking.
\newblock In \emph{ICML 2024 Workshop on Geometry-grounded Representation Learning and Generative Modeling}, 2024.
\newblock URL \url{https://openreview.net/forum?id=1VlRaXNMWO}.

\bibitem[Kaba and Ravanbakhsh(2023)]{kaba2023symmetry}
S{\'e}kou-Oumar Kaba and Siamak Ravanbakhsh.
\newblock Symmetry breaking and equivariant neural networks.
\newblock In \emph{NeurIPS 2023 Workshop on Symmetry and Geometry in Neural Representations}, 2023.
\newblock URL \url{https://openreview.net/forum?id=d55JaRL9wh}.

\bibitem[Lin et~al.(2019)Lin, Huang, Collins, Bradbury, and Malof]{8898722}
Kangcheng Lin, Bohao Huang, Leslie~M. Collins, Kyle Bradbury, and Jordan~M. Malof.
\newblock A simple rotational equivariance loss for generic convolutional segmentation networks: preliminary results.
\newblock In \emph{IGARSS 2019 - 2019 IEEE International Geoscience and Remote Sensing Symposium}, pages 3876--3879, 2019.
\newblock \doi{10.1109/IGARSS.2019.8898722}.

\bibitem[Shakerinava et~al.(2022)Shakerinava, Mondal, and Ravanbakhsh]{shakerinava2022structuring}
Mehran Shakerinava, Arnab~Kumar Mondal, and Siamak Ravanbakhsh.
\newblock Structuring representations using group invariants.
\newblock In Alice~H. Oh, Alekh Agarwal, Danielle Belgrave, and Kyunghyun Cho, editors, \emph{Advances in Neural Information Processing Systems}, 2022.
\newblock URL \url{https://openreview.net/forum?id=vWUmBjin_-o}.

\bibitem[Bhardwaj et~al.(2023)Bhardwaj, McClinton, Wang, Lajoie, Sun, Isola, and Krishnan]{bhardwaj2023steerableequivariantrepresentationlearning}
Sangnie Bhardwaj, Willie McClinton, Tongzhou Wang, Guillaume Lajoie, Chen Sun, Phillip Isola, and Dilip Krishnan.
\newblock Steerable equivariant representation learning, 2023.
\newblock URL \url{https://arxiv.org/abs/2302.11349}.

\bibitem[Basu et~al.(2023)Basu, Sattigeri, Natesan~Ramamurthy, Chenthamarakshan, Varshney, Varshney, and Das]{Basu_Sattigeri_Natesan}
Sourya Basu, Prasanna Sattigeri, Karthikeyan Natesan~Ramamurthy, Vijil Chenthamarakshan, Kush~R. Varshney, Lav~R. Varshney, and Payel Das.
\newblock Equi-tuning: Group equivariant fine-tuning of pretrained models.
\newblock \emph{Proceedings of the AAAI Conference on Artificial Intelligence}, 2023.
\newblock URL \url{https://ojs.aaai.org/index.php/AAAI/article/view/25832}.

\bibitem[Kim et~al.(2023{\natexlab{a}})Kim, Nguyen, Suleymanzade, An, and Hong]{kim2023learning}
Jinwoo Kim, Dat~Tien Nguyen, Ayhan Suleymanzade, Hyeokjun An, and Seunghoon Hong.
\newblock Learning probabilistic symmetrization for architecture agnostic equivariance.
\newblock In \emph{Thirty-seventh Conference on Neural Information Processing Systems}, 2023{\natexlab{a}}.
\newblock URL \url{https://openreview.net/forum?id=phnN1eu5AX}.

\bibitem[Zheng et~al.(2024)Zheng, Liu, Li, Yao, and Rong]{zheng2024relaxingcontinuousconstraintsequivariant}
Zinan Zheng, Yang Liu, Jia Li, Jianhua Yao, and Yu~Rong.
\newblock Relaxing continuous constraints of equivariant graph neural networks for physical dynamics learning, 2024.
\newblock URL \url{https://arxiv.org/abs/2406.16295}.

\bibitem[Yang et~al.(2023)Yang, Walters, Dehmamy, and Yu]{yang2023generative}
Jianke Yang, Robin Walters, Nima Dehmamy, and Rose Yu.
\newblock Generative adversarial symmetry discovery.
\newblock In \emph{Proceedings of the 40th International Conference on Machine Learning}, 2023.

\bibitem[{CMU}(2003)]{cmumocap2003}
{CMU}.
\newblock Carnegie mellon motion capture database.
\newblock \url{http://mocap.cs.cmu.edu}, 2003.

\bibitem[Huang et~al.(2022)Huang, Han, Rong, Xu, Sun, and Huang]{huang2022equivariantgraphmechanicsnetworks}
Wenbing Huang, Jiaqi Han, Yu~Rong, Tingyang Xu, Fuchun Sun, and Junzhou Huang.
\newblock Equivariant graph mechanics networks with constraints.
\newblock In \emph{International Conference on Learning Representations}, 2022.
\newblock URL \url{https://openreview.net/forum?id=SHbhHHfePhP}.

\bibitem[Finzi et~al.(2021{\natexlab{b}})Finzi, Welling, and Wilson]{finzi2021practicalmethodconstructingequivariant}
Marc Finzi, Max Welling, and Andrew~Gordon Wilson.
\newblock A practical method for constructing equivariant multilayer perceptrons for arbitrary matrix groups.
\newblock In \emph{Proceedings of the 38th International Conference on Machine Learning}, 2021{\natexlab{b}}.

\bibitem[Kim et~al.(2023{\natexlab{b}})Kim, Lee, Yang, and Lee]{kim2023regularizingsoftequivariancemixed}
Hyunsu Kim, Hyungi Lee, Hongseok Yang, and Juho Lee.
\newblock Regularizing towards soft equivariance under mixed symmetries.
\newblock In \emph{Proceedings of the 40th International Conference on Machine Learning}, 2023{\natexlab{b}}.

\bibitem[Chmiela et~al.(2017)Chmiela, Tkatchenko, Sauceda, Poltavsky, Schütt, and Müller]{Chmiela_2017}
Stefan Chmiela, Alexandre Tkatchenko, Huziel~E. Sauceda, Igor Poltavsky, Kristof~T. Schütt, and Klaus-Robert Müller.
\newblock Machine learning of accurate energy-conserving molecular force fields.
\newblock \emph{Science Advances}, 3\penalty0 (5), May 2017.
\newblock ISSN 2375-2548.
\newblock \doi{10.1126/sciadv.1603015}.
\newblock URL \url{http://dx.doi.org/10.1126/sciadv.1603015}.

\bibitem[Cen et~al.(2024)Cen, Li, Lin, Ren, Wang, and Huang]{cen2024high}
Jiacheng Cen, Anyi Li, Ning Lin, Yuxiang Ren, Zihe Wang, and Wenbing Huang.
\newblock Are high-degree representations really unnecessary in equivariant graph neural networks?
\newblock In \emph{The Thirty-eighth Annual Conference on Neural Information Processing Systems}, 2024.
\newblock URL \url{https://openreview.net/forum?id=M0ncNVuGYN}.

\bibitem[Crawshaw and Košecká(2021)]{crawshaw2021slawscaledlossapproximate}
Michael Crawshaw and Jana Košecká.
\newblock Slaw: Scaled loss approximate weighting for efficient multi-task learning, 2021.
\newblock URL \url{https://arxiv.org/abs/2109.08218}.

\bibitem[Bohn et~al.(2024)Bohn, Freeman, Tercan, and Meisen]{bohn2024taskweightinggradientprojection}
Christian Bohn, Ido Freeman, Hasan Tercan, and Tobias Meisen.
\newblock Task weighting through gradient projection for multitask learning, 2024.
\newblock URL \url{https://arxiv.org/abs/2409.01793}.

\bibitem[Zaidi et~al.(2023)Zaidi, Schaarschmidt, Martens, Kim, Teh, Sanchez-Gonzalez, Battaglia, Pascanu, and Godwin]{zaidi2023pretraining}
Sheheryar Zaidi, Michael Schaarschmidt, James Martens, Hyunjik Kim, Yee~Whye Teh, Alvaro Sanchez-Gonzalez, Peter Battaglia, Razvan Pascanu, and Jonathan Godwin.
\newblock Pre-training via denoising for molecular property prediction.
\newblock In \emph{The Eleventh International Conference on Learning Representations}, 2023.
\newblock URL \url{https://openreview.net/forum?id=tYIMtogyee}.

\bibitem[Ni et~al.(2024)Ni, Feng, Hong, Sun, Ma, Ma, Ye, and Lan]{ni2024pretrainingfractionaldenoisingenhance}
Yuyan Ni, Shikun Feng, Xin Hong, Yuancheng Sun, Wei-Ying Ma, Zhi-Ming Ma, Qiwei Ye, and Yanyan Lan.
\newblock Pre-training with fractional denoising to enhance molecular property prediction, 2024.
\newblock URL \url{https://arxiv.org/abs/2407.11086}.

\bibitem[Li et~al.(2018)Li, Xu, Taylor, Studer, and Goldstein]{li2018visualizinglosslandscapeneural}
Hao Li, Zheng Xu, Gavin Taylor, Christoph Studer, and Tom Goldstein.
\newblock Visualizing the loss landscape of neural nets.
\newblock In \emph{Neural Information Processing Systems}, 2018.

\end{thebibliography}
% For bibLaTeX users:
% \printbibliography

\newpage
\appendix

\section{Loss Surface}
\label{sec: Loss surface}
In this section, we analyze the relative ease of training equivariant models compared to non-equivariant models by examining the loss surface around the achieved local minima for each model. We explore how each architecture influences the loss landscape when trained on the same task. 
% However, given the high dimensionality of neural network parameter spaces, visualizing their loss functions in three dimensions can be challenging. 
However, due to the high dimensionality of parameter spaces in neural networks, visualizing their loss functions in three dimensions might be a significant challenge.
We use the filter normalization method introduced by \cite{li2018visualizinglosslandscapeneural}, which calculates the loss function along two randomly selected Gaussian directions in the parameters space, starting from the optimal parameters $\theta^{*}$
achieved at the end of training. 

\begin{figure*}[h]
  \centering
  \subfigure[Geometric Algebra Transformer]{\label{fig:gat_loss_surface}
    \includegraphics[width=0.4\textwidth]{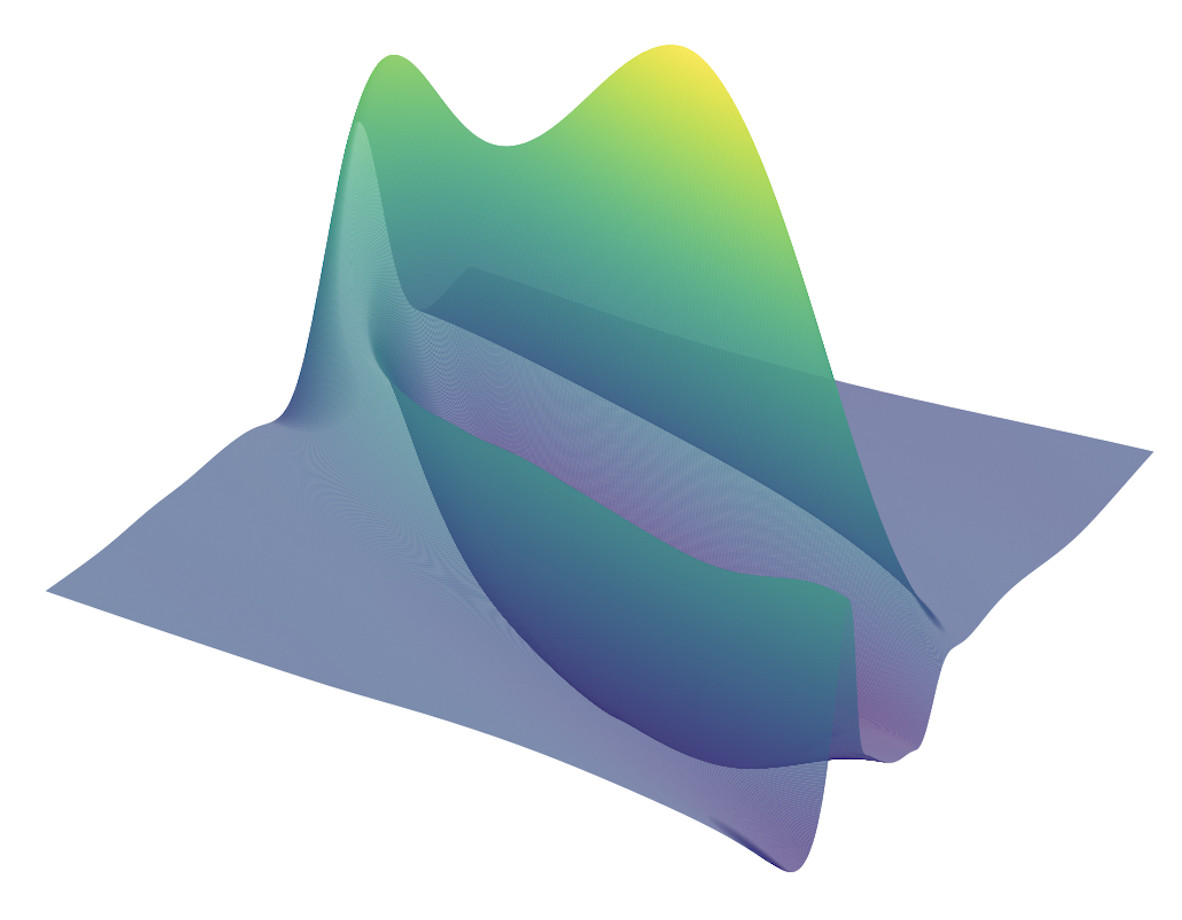}
  }\hspace{-0.01cm} 
  \subfigure[Transformer]{\label{fig:transformer_loss_surface}
    \includegraphics[width=0.4\textwidth]{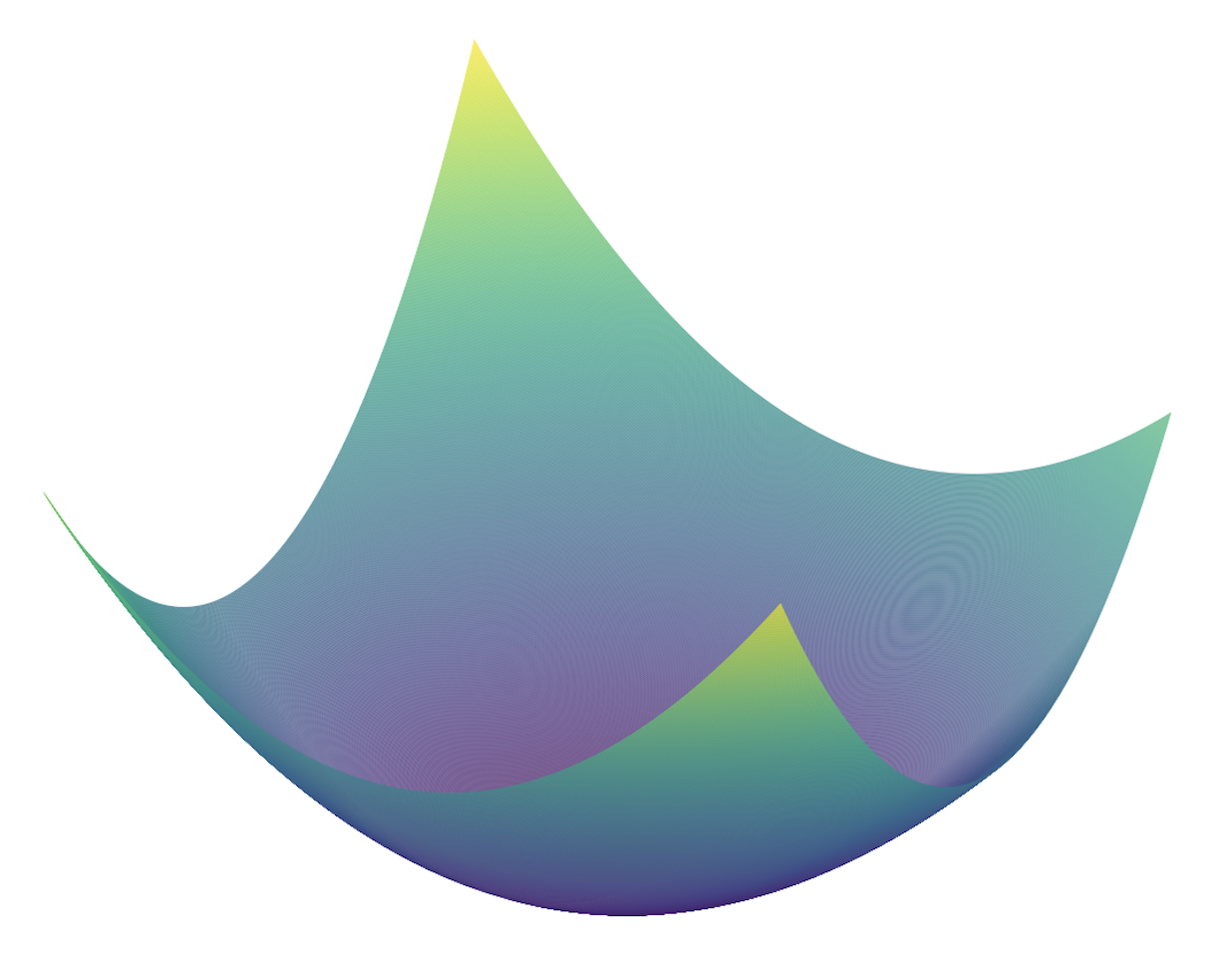}
  }
  \caption{Loss surface around local minima of trained models on N-body dynamical system.}
  \label{fig:loss_surface}
\end{figure*}

We visualize the loss surface of the Geometric Algebra Transformer (GATr) and Transformer in Figure \ref{fig:loss_surface}, trained on the N-body dynamical system. We observe that the Transformer architecture exhibits a more favorable loss shape around its local minima, characterized by a convex structure. This might suggest that the optimization path for the Transformer is smoother and potentially easier to navigate during training, leading to more stable convergence.
In contrast, the loss surface of GATr appears more erratic and rugged. This complexity in the loss landscape can indicate multiple local minima and a higher sensitivity to initial conditions or parameter settings. Such characteristics might complicate the training process, requiring more careful tuning of hyperparameters. We leave this for future work to analyze how the optimization path for each model behaves during training.
% Finally, it is important to note that, although we observed a more convex loss landscape near local minima for unconstrained models compared to their equivariant counterparts, this conclusion is subject to certain limitations. Specifically, our analysis did not account for the trajectories these models follow during optimization. Understanding the optimization paths and the influence of different initialization settings on these paths remain an open question. In future work, we plan to examine more closely the optimization process of each model and how it evolves during training.

\section{Proofs}
\label{app:proofs}
\subsection{Propositions}
\label{app:proof_tradeoff_detail}
\textbf{Proposition~\ref{prop:empirical_tradeoff}.}
Let $f_{\alpha,\beta} \in \arg\min_{f\in\cH}\hatLtot(f; \alpha, \beta)$ be an empirical minimizer of the \methodabbr{} objective, and let $f^\star_{\mathrm{obj}} \in \arg\min_{f\in\cH} \hatLobj(f)$ be an empirical minimizer for the objective loss alone.
Then:
\begin{enumerate}
    \item[(a)] $f_{\alpha,\beta}$ is Pareto optimal for the bi-objective problem $(\min \hatLobj(f), \min \hatLequi(f))$.
    \item[(b)] The following trade-off inequality holds:
        \begin{equation}
          \hatLobj(f_{\alpha,\beta}) - \hatLobj(f^\star_{\mathrm{obj}}) \leq \frac{\beta}{\alpha} \left( \hatLequi(f^\star_{\mathrm{obj}}) - \hatLequi(f_{\alpha,\beta}) \right). \label{eq:main_tradeoff_obj_cost_prop}
        \end{equation}
    % \item[(c)] Symmetrically, for any $\tilde{f} \in \cH$:
    %     \begin{equation}
    %     \hatLequi(f_{\alpha,\beta}) - \hatLequi(\tilde{f}) \leq \frac{\alpha}{\beta} \left( \hatLobj(\tilde{f}) - \hatLobj(f_{\alpha,\beta}) \right). \label{eq:main_tradeoff_equi_gain_prop}
    %     \end{equation}
\end{enumerate}
\begin{proof}
Let $\fRemul$ be an empirical minimizer of $\hatLtot(f;\alpha,\beta)$. By definition, for any $\tilde{f} \in \cH$:
\[ \alpha\,\hatLobj(\fRemul) + \beta\,\hatLequi(\fRemul) \le \alpha\,\hatLobj(\tilde{f}) + \beta\,\hatLequi(\tilde{f}). \]
To show Eq. \ref{eq:main_tradeoff_obj_cost_prop}: Rearrange the optimality condition:
\[ \alpha(\hatLobj(\fRemul) - \hatLobj(\tilde{f})) \le \beta(\hatLequi(f') - \hatLequi(\fRemul)). \]
Dividing by $\alpha > 0$:
\[ \hatLobj(\fRemul) - \hatLobj(\tilde{f}) \le \frac{\beta}{\alpha}(\hatLequi(\tilde{f}) - \hatLequi(\fRemul)). \]
Setting $\tilde{f} = \fObjEmp$ yields Eq. \ref{eq:main_tradeoff_obj_cost_prop}.

% To show the symmetrical inequality (\hyperref[eq:main_tradeoff_equi_gain_prop]{Equation \ref*{eq:main_tradeoff_equi_gain_prop}}):
% Rearrange the initial optimality condition as:
% \[ \beta(\hatLequi(\fRemul) - \hatLequi(\tilde{f})) \le \alpha(\hatLobj(\tilde{f}) - \hatLobj(\fRemul)). \]
% Dividing by $\beta > 0$ yields the stated result.

For Pareto Optimality: Assume, for contradiction, that $\fRemul$ is not Pareto optimal. Then there exists an $\widetilde{\widetilde{f}} \in \cH$ such that $\hatLobj(\widetilde{\widetilde{f}}) \le \hatLobj(\fRemul)$ and $\hatLequi(\widetilde{\widetilde{f}}) \le \hatLequi(\fRemul)$, with at least one of these inequalities being strict. Since $\alpha > 0$ and $\beta > 0$, this would imply $\alpha\hatLobj(\widetilde{\widetilde{f}}) + \beta\hatLequi(\widetilde{\widetilde{f}}) < \alpha\hatLobj(\fRemul) + \beta\hatLequi(\fRemul)$. This contradicts the assumption that $\fRemul$ is a minimizer of $\hatLtot(f; \alpha, \beta)$. Therefore, $\fRemul$ must be Pareto optimal.
\end{proof}

\subsection{Equivariance Measure}
\label{app:measure_equivariant}
We define the equivariance metric $E$ to quantify the degree of equivariance exhibited by a function $f$, as:
\begin{equation}
\label{eq:equi_measure_app}
    E(f, G) = \frac{1}{|\mathcal{D}|} \sum_{x \in \mathcal{D}} \left\| \frac{1}{M} \sum_{i=1}^M \rho(g_i)(f(x)) - \frac{1}{M} \sum_{i=1}^M f(\phi(g_i)(x)) \right\|_2
\end{equation}
\begin{proof}
Starting from Eq. \ref{equation: equi_function}: $f(\phi(g)(x)) = \rho(g)(f(x)$, 
% and assume that \( G \) is a finite or compact group with normalized Haar measure \( \mu \), 
the group integration of both sides \textit{w.r.t.} the normalized Haar measure \( \mu \) yields:
\begin{equation}
\label{eq:integral both sides}
\int_{G} f(\phi(g)(x)) \, d\mu(g) = \int_{G} \rho(g)(f(x)) \, d\mu(g)
\end{equation}
When \( G \) is a large or continuous group, as is the case in our work, the integrals over \( G \) may not be computable in closed form. Therefore, we approximate the integrals using a Monte Carlo approach with samples \( \{g_i\}_{i=1}^M \) from \( G \):
\begin{equation}
\label{eq:monte carlo 1}
\int_{G} f(\phi(g)(x)) \, d\mu(g)
 \approx \frac{1}{M} \sum_{i=1}^M f(\phi(g_i)(x)) 
\end{equation}
\begin{equation}
\label{eq:monte carlo 2}
\int_{G} \rho(g)(f(x)) \, d\mu(g) 
 \approx  \frac{1}{M} \sum_{i=1}^M \rho(g_i)(f(x))
\end{equation}

where $M$ is a large number of samples from $G$. 

Given the group averages, we can then define the equivariance error \( E(f, G) \) as the average norm of the difference between these two averages over the data distribution \( \mathcal{D} \):
\begin{equation}
\label{proof_eq:equi_measure_1}
    E(f, G) = \frac{1}{|\mathcal{D}|} \sum_{x \in \mathcal{D}} \left\| \frac{1}{M} \sum_{i=1}^M \rho(g_i)(f(x)) - \frac{1}{M} \sum_{i=1}^M f(\phi(g_i)(x)) \right\|_2
\end{equation}
with $\| \cdot \|_2$ denotes an $L_2$ norm (for non-scalar function). 
% This error indicates the average deviation of a function \( f \) from perfect equivariance across the data distribution \( D \) (lower value means more equivariant function). 

% indicate a practical metric for evaluating how closely the function $f$ approximates perfect equivariance throughout a data distribution $D$ (which should be zero for a perfect equivariance function). In practice, we use $M = 100$ samples from the group and noticed this was sufficient to obtain stable results. We also observed that both measures have very similar behavior in our experiments, where $E$ and $E'$ are near zero for equivariant models. We also demonstrate that increasing the value of $\beta$ in \hyperref[eq_total_loss]{Equation \ref*{eq_total_loss}} results in a less equivariant error for $E$ and $E'$.
\end{proof}
\section{Implementation Details}
\label{appendix: implementation_details}
\subsection{Equivariance Loss}
The empirical equivariance loss defined in Eq.~\ref{eq:equivariant_loss}, $\hatLequi(f_\theta) = \tfrac1n\sum \E_{g \sim G}[\lossell(f_\theta(\phig x_i), \rhog y_i)]$, measures the consistency of the model's predictions on transformed inputs against the correspondingly transformed ground truth labels. 
It is distinct from a direct measure of functional equivariance, which compare $f_\theta(\phig x_i)$ with $\rhog f_\theta(x_i)$ (the transformed prediction of the original input). While the latter directly assesses the equivariance of the function $f_\theta$ itself, our choice of $\hatLequi$ offers a crucial advantage: it continuously anchors the learning process to the ground truth. 
To see this, let $f_\theta(x) = y(x) + \gamma(x)$, where $y(x)$ is the true label for input $x$ and $\gamma(x)$ is the model's prediction error. If we assume the ground truth data itself is perfectly equivariant, i.e., $y(\phig x) = \rhog y(x)$, then the term minimized by $\hatLequi$ (for a single instance, taking $\lossell$ as an $L_p$ norm) becomes:
\[ \|\underbrace{f_\theta(\phig x_i)}_{y(\phig x_i) + \gamma(\phig x_i)} - \underbrace{\rhog y_i}_{y(\phig x_i)}\|_p = \| \gamma(\phig x_i) \|_p. \]
Thus, minimizing $\hatLequi$ directly minimizes the magnitude of the prediction error on transformed inputs. This helps prevent the model from "drifting" into solutions that might be equivariant but incorrect (i.e., $f_\theta(\phig x_i) \approx \rhog f_\theta(x_i)$ but both are far from $\rhog y_i$).
In contrast, a loss term based on functional equivariance, $\|f_\theta(\phig x_i) - \rhog f_\theta(x_i)\|_p$, would simplify to $\|\gamma(\phig x_i) - \rhog \gamma(x_i)\|_p$. While this term directly encourages the *error itself* to be equivariant, minimizing it alone does not guarantee that the error magnitude $\|\gamma(\cdot)\|_p$ is small.
Our \methodabbr{} objective, by combining $\alpha\hatLobj(f_\theta)$ (which minimizes $\|\gamma(x_i)\|_p$ on original data) with $\beta\hatLequi(f_\theta)$ (which minimizes $\|\gamma(\phig x_i)\|_p$ on transformed data, given ideal data equivariance), aims for both accuracy and consistency under transformations. The degree to which this also induces functional equivariance in $f_\theta$ (i.e., making $\| \gamma(\phig x_i) - \rhog \gamma(x_i) \|_p$ small) is then assessed empirically using the equivariance metrics $E$ and $E'$  as shown in our experiments.
\subsection{N-Body Dynamical System}
\label{appendix: n_body_system}
Following the methodology outlined in \cite{brehmer2023geometric}, the dataset for the N-body system simulation encompasses four objects per sample. The center object is assigned a mass ranging from $1$ to $10$, whereas the other objects are uniformly positioned at a radius from $0.1$ to $1.0$ with masses between $0.01$ and $0.1$. 
We structured the datasets into two setups: in-distribution and out-of-distribution (OOD). Each sample in the in-distribution dataset is  subjected to a random rotation within the range $[-10^\circ, 10^\circ]$.  
REMUL and data augmentation are trained with random rotations in the same range. Conversely, the OOD dataset is designed to evaluate the model’s generalization capabilities by incorporating extreme rotational perturbations, specifically with angles set within the ranges $[-180^\circ, -90^\circ]$ and $[90^\circ, 180^\circ]$. We trained on $100$ samples, and each of the validation, test, and OOD datasets contains $5000$ samples. For models hyperparameters and training, we follow the same settings in \cite{brehmer2023geometric}, summarized in Table \ref{tab:N body model parameters}. For REMUL, initial $\alpha =1$. 
\begin{table}[ht]
\centering
\caption{Hyperparameters settings for N-body dynamical system.}
\label{tab:N body model parameters}
\begin{tabular}{@{}lccc@{}}
\toprule
Hyperparameters & Geometric Algebra Transformer & SE(3)-Transformer & Transformer \\ \midrule
\#\text{attention blocks} & $10$ & - & $10$ \\
\#\text{channels} & $128$ & $8$ & $384$ \\
\#\text{attention heads} & $8$ & $1$ & $8$ \\
\#\text{multivector} & $16$ & - & - \\
\#\text{layers} & - & $4$ & - \\
\#\text{degrees} & - & $4$ & - \\
\#\text{training steps} & $50000$ & $50000$ & $50000$ \\
\#\text{optimizer} & Adam & Adam & Adam \\
\#\text{batch size} & $64$ & $64$ & $64$ \\
\#\text{lr} &  $3 \times 10^{-4}$ &  $3 \times 10^{-4}$ & $3 \times 10^{-4}$ \\
\bottomrule
\end{tabular}
\end{table}
\subsection{Motion Capture}
\label{appendix: Motion Capture}
Motion Capture dataset by \cite{cmumocap2003} features 3D trajectory data that records a range of human motions, and the task involves predicting the final trajectory based on initial positions and velocities. We have reported results for two types of motion: Walking (Subject \#$35$) and Running (Subject \#$9$).

\begin{table}[t!]
\centering
\caption{Hyperparameters settings for Motion Capture dataset.}
\label{tab:motion capture parameters}
\begin{tabular}{@{}lccc@{}}
\toprule
Hyperparameters & Geometric Algebra Transformer & SE(3)-Transformer & Transformer \\ \midrule
\#\text{attention blocks} & $12$ & - & $10$ \\
\#\text{channels} & $128$ & $8$ & $384$ \\
\#\text{attention heads} & $8$ & $1$ & $8$ \\
\#\text{multivector} & $16$ & - & - \\
\#\text{layers} & - & $4$ & - \\
\#\text{degrees} & - & $4$ & - \\
\#\text{epochs} & $2000$ & $2000$ & $2000$ \\
\#\text{optimizer} & Adam & Adam & Adam \\
\#\text{batch size} & $12$ & $12$ & $12$ \\
\#\text{lr} &  $3 \times 10^{-4}$ &  $3 \times 10^{-4}$ & $3 \times 10^{-4}$ \\
\bottomrule
\end{tabular}
\begin{tabular}{@{}lcccc@{}}
\toprule
Hyperparameters & Equivariant MLP & RPP & PER & standard MLP \\ \midrule
\#\text{hidden dim} & $532$ & $348$ & $532$ & $680$ \\
\#\text{layers} & $3$ & $3$ & $3$ & $3$\\
% Residual Pathway Priors
% Projection Equivariance
\bottomrule
\end{tabular}
\end{table}

% \newpage
Following the standard experimental setup in the literature on this task \citep{han2022equivariantgraphhierarchybasedneural, huang2022equivariantgraphmechanicsnetworks, xu2024equivariantgraphneuraloperator},
we apply a train/validation/test split of $200$/$600$/$600$ for Walking and $200$/$240$/$240$ for Running. The interval between trajectories, $\Delta T = 30$ for both tasks. 
For model hyperparameters, we fine-tuned around the same in Table \ref{tab:N body model parameters} and found it the best for each model except for the Geometric Algebra Transformer we increased the attention blocks to $12$.
We train each model for $2000$ epochs with batch size $=12$. For MLP comparisons, all models and baselines have the same number of layers and parameters. More details in Table \ref{tab:motion capture parameters}. 
For REMUL, $\alpha =1$.
\subsection{Molecular Dynamics}
\label{appendix: Molecular Dynamics}
MD17 dataset \citep{Chmiela_2017} is a molecular dynamics benchmark that contains the trajectories of eight small molecules (Aspirin, Benzene, Ethanol, Malonaldehyde Naphthalene, Salicylic, Toluene, Uraci). We use the same dataset split in \cite{huang2022equivariantgraphmechanicsnetworks, xu2024equivariantgraphneuraloperator}, allocating $500$ samples for train, $2000$ for validation, and $2000$ for test. The interval between trajectories, $\Delta T = 5000$. We selected the Equivariant Graph Neural Networks (EGNN) architecture and its non-equivariant version GNN, as introduced by \cite{satorras2022en}. The input for GNN architecture is the initial positions along with atom types.
Both architectures have the same hyperparameters, details in Table \ref{tab:hyperparameters_settings_for_MD17_dataset}.
For REMUL, $\alpha =1$.
\begin{table}[ht]
\centering
\caption{Hyperparameters settings for MD17 dataset.}
\label{tab:hyperparameters_settings_for_MD17_dataset}
\begin{tabular}{@{}lc@{}}
\toprule
Hyperparameters          &     \\ \midrule
\#\text{layers} & $4$        \\
\#\text{hidden dim}         & $64$       \\
\#\text{epochs}           & $500$     \\
\#\text{optimizer}        & Adam      \\
\#\text{batch size}       & $200$       \\
\#\text{lr} & $5 \times 10^{-4}$ \\
\bottomrule
\end{tabular}
\end{table}
\newpage
\subsection{Computational Complexity}
\label{appendix: Time Complexity}
In the computational experiment of Geometric Algebra Transformer (GATr) and Transformer, we selected models with an equivalent number of blocks and parameters. GATr incorporates a unique design that includes a multivector parameter; we adjusted the Transformer architecture to match the parameter count of GATr. Both models have around $2.6$M parameters, detailed configurations are provided in Table \ref{tab:Hyperparameters settings for time complexit}. 
SE(3)-Transformer gives out of memory for this setting.
We selected a uniformly random Gaussian input with $20$ nodes and $7$ features dimension. 
We measured the computational efficiency of each model by recording the time taken for both forward and backward passes during training, as well as the inference time as a function of batch size. 
For each value, we took the average over $10$ runs with Nvidia A$10$ GPU.
\begin{table}[ht]
\centering
\caption{Hyperparameters settings for Computational Complexity.}
\label{tab:Hyperparameters settings for time complexit}
\begin{tabular}{@{}lccc@{}}
\toprule
Hyperparameters & Geometric Algebra Transformer & Transformer \\ \midrule
\#\text{attention blocks} & $12$ & $12$ \\
\#\text{channels} & $128$ &  $168$ \\
\#\text{attention heads} & $8$ &  $8$ \\
\#\text{multivector} & $16$ & - \\
\bottomrule
\end{tabular}
\end{table}
\subsection{Compute Resources}
\label{appendix: Compute Resources}
We ran all the experiments using a single Nvidia A10 GPU.
% \newpage
\section{Additional Experiments}
\label{appendix: Additional Experiments}
% In this section, we include additional results on the three tasks (N-Body Dynamical System, Motion Capture, and Molecular Dynamics), using the equivariance measure defined in \hyperref[eq:equi_measure_2]{Equation \ref*{eq:equi_measure_2}} which is consistent with our results in the paper. 
% We also provide ablation studies on the number of samples required from the symmetry group during training for REMUL and data augmentation.
% Finally, we include molecules from the MD17 dataset, along with visualizations of their structures in both 2D and 3D.

% In this section, we present additional results on the three tasks: N-body dynamical system, Motion Capture, and Molecular Dynamics using the standard equivariance measure defined in Eq. \ref{eq:equi_measure_2}, which is consistent with our findings in the paper. We also include ablation studies on the number of samples required from the symmetry group during training for REMUL and data augmentation. Finally, we showcase molecules from the MD17 dataset, along with 2D and 3D visualizations of their structures.

This section provides additional experimental results to further validate our training procedure \methodabbr{}. We include:
\begin{itemize}
    \item \textbf{Additional evaluations on the three tasks}: For the N-body dynamical system, Motion Capture, and Molecular Dynamics (MD17), we include the standard equivariance error $E'$ (defined in Eq.~\ref{eq:equi_measure_2} of the main paper). These results are consistent with our findings in the paper, using the $E$ metric.
    \item \textbf{Performance on MD17:} We include detailed results showing performance and equivariance error trade-offs for \methodabbr{} applied to GNN architecture on individual molecules from the MD17 dataset (complementing Table~\ref{tab:performance:MD17} in the main paper).
    \item \textbf{Ablation on Group Sampling:} We conduct an ablation study investigating the impact of the number of group samples used during training for \methodabbr{} and data augmentation.
    \item \textbf{Convergence Speed Analysis:} We compare the convergence speed of \methodabbr{} against data augmentation by tracking training and validation MSE as a function of training steps.
    \item \textbf{Axis-Specific Equivariance Error on Motion Capture:} To further investigate the nature of symmetries in the Motion Capture dataset, we report the equivariance error around individual $X$, $Y$, and $Z$ axes, separately. 
    \item \textbf{Additional N-Body System Benchmark:} We evaluate \methodabbr{} (applied to a GNN model) on an additional N-body system benchmark, comparing it against several equivariant architectures.
    \item \textbf{Molecular Structures:} We provide 2D and 3D visualizations of the molecules from the MD17 dataset.
\end{itemize}
\newpage
\subsection{N-Body Dynamical System}
\label{Additional Experiments: N body system}
\begin{figure}[ht]
    \centering
    \subfigure[REMUL: Gradual penalty]{
        \includegraphics[width=0.3\textwidth]{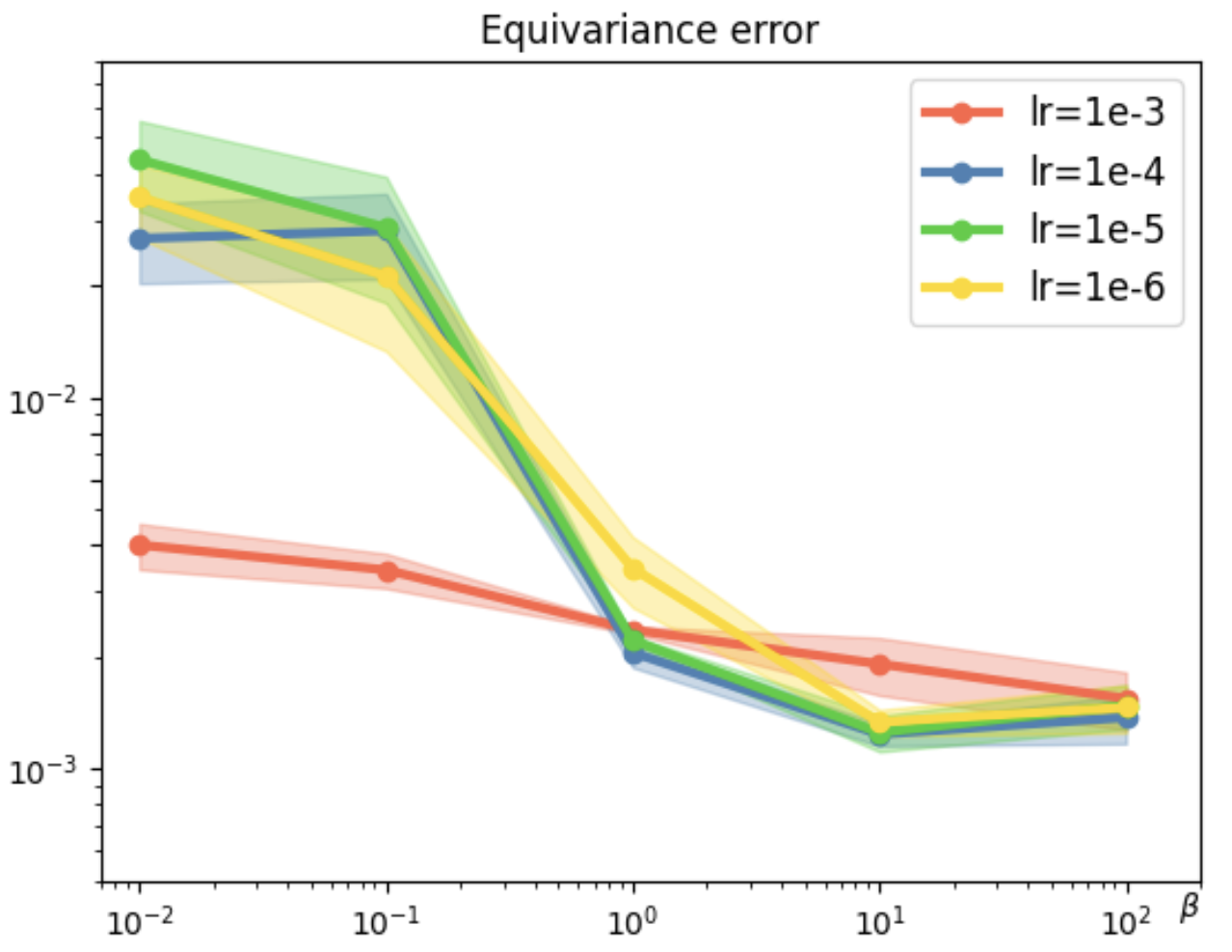}
    }
    \hspace{-0.01cm}
    \subfigure[REMUL: Constant penalty]{
      \begin{tikzpicture}[baseline={(0,0)}, scale=0.95]
      \node[inner sep=0, anchor=base] {\includegraphics[width=0.3\textwidth]{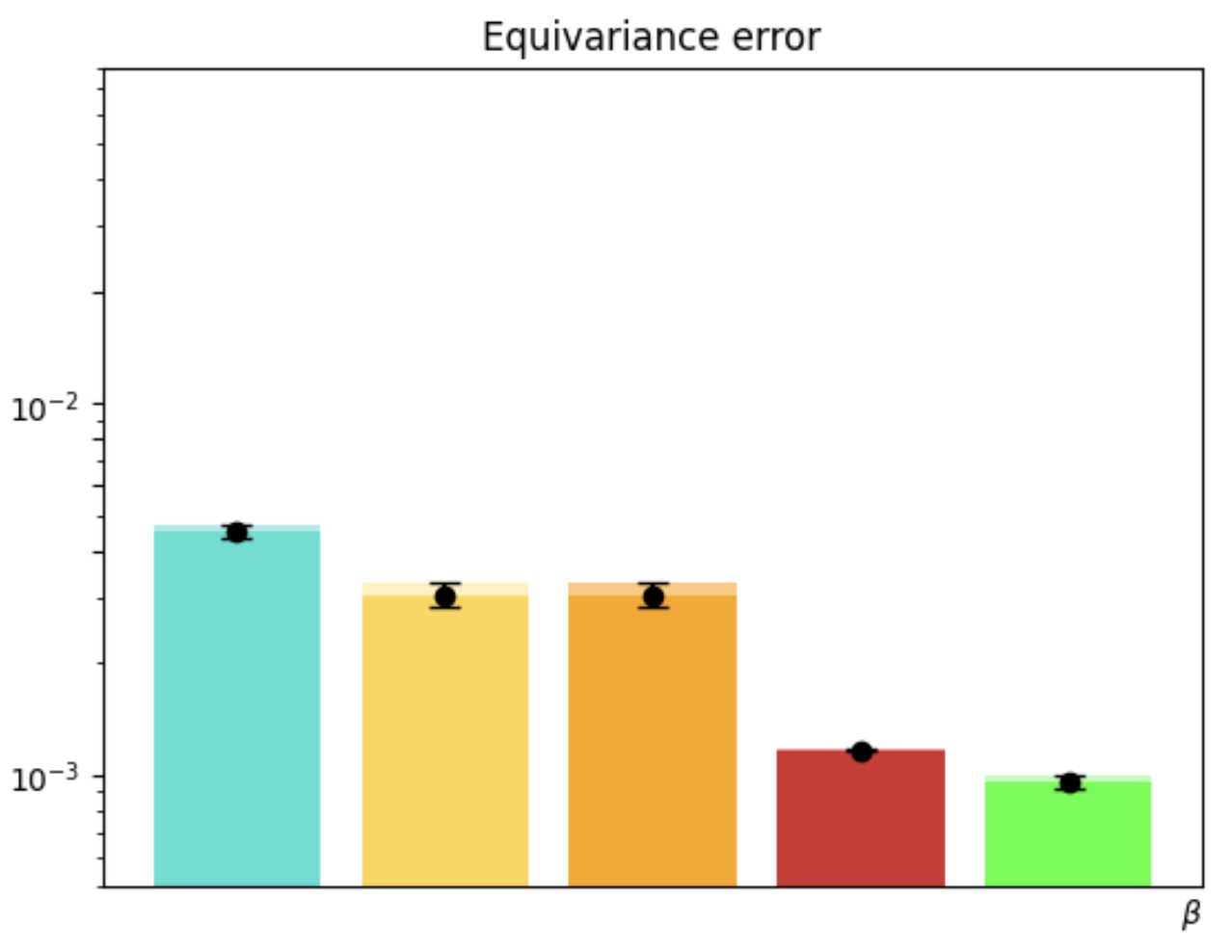}};
      \node at (-1.36, 0.1) {\scalebox{0.46}{$0.01$}};
      \node at (-0.6, 0.1) {\scalebox{0.46}{$0.1$}};
      \node at (0.1, 0.1) {\scalebox{0.46}{$1.0$}};
      \node at (0.88, 0.1) {\scalebox{0.46}{$10.0$}};
      \node at (1.65, 0.1) {\scalebox{0.46}{$100.0$}};
    \end{tikzpicture}%
    }
    \hspace{-0.01cm}
    \subfigure{
    \begin{tikzpicture}[baseline={(0,0)}, scale=0.95]
        \node[inner sep=0, anchor=base] {\includegraphics[width=0.3\textwidth]{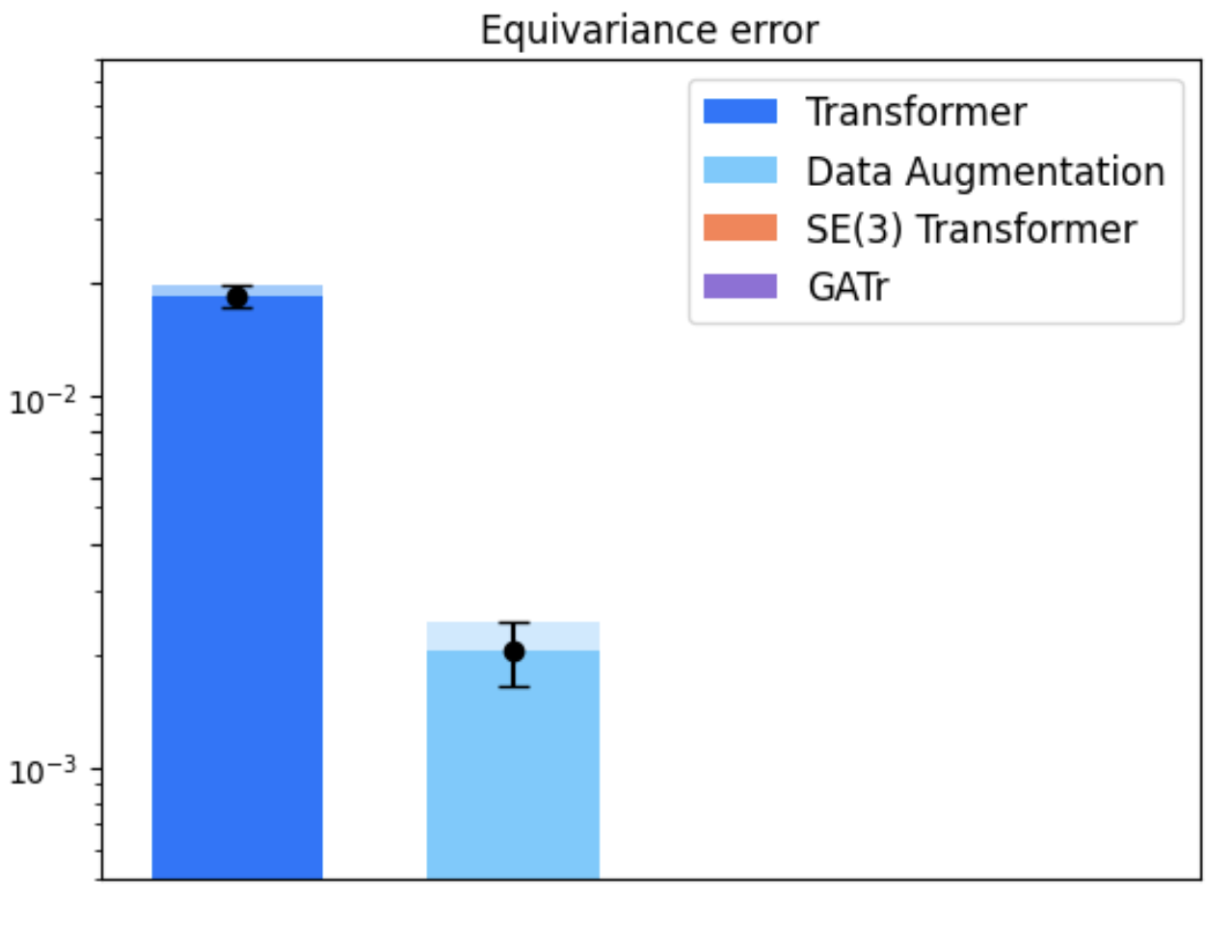}};
        \fill[customorange] (0.34,0.08) rectangle (0.9,0.18);  
        \fill[custompurple] (1.27,0.08) rectangle (1.8,0.18);
        \node at (0.0, -0.3) {\scalebox{0.9}{(c) Baselines}};
        \end{tikzpicture}
    }
    \caption{N-body dynamical system. The second equivariance measure $E'$.
    (a) Transformer trained with REMUL (gradual penalty), (b) Transformer trained with REMUL (constant penalty), and (c) Baselines: Equivariant models, standard Transformer, and data augmentation.}
    \label{fig:N body system: second equivariant measure}
\end{figure}

\subsection{Number of Samples from the Symmetry Group}
\begin{figure}[t!]
    \centering
    % First figure on the left
    \subfigure[MSE: In-distribution]{
    \label{fig:group_samples_id}
        \includegraphics[width=0.42\textwidth]{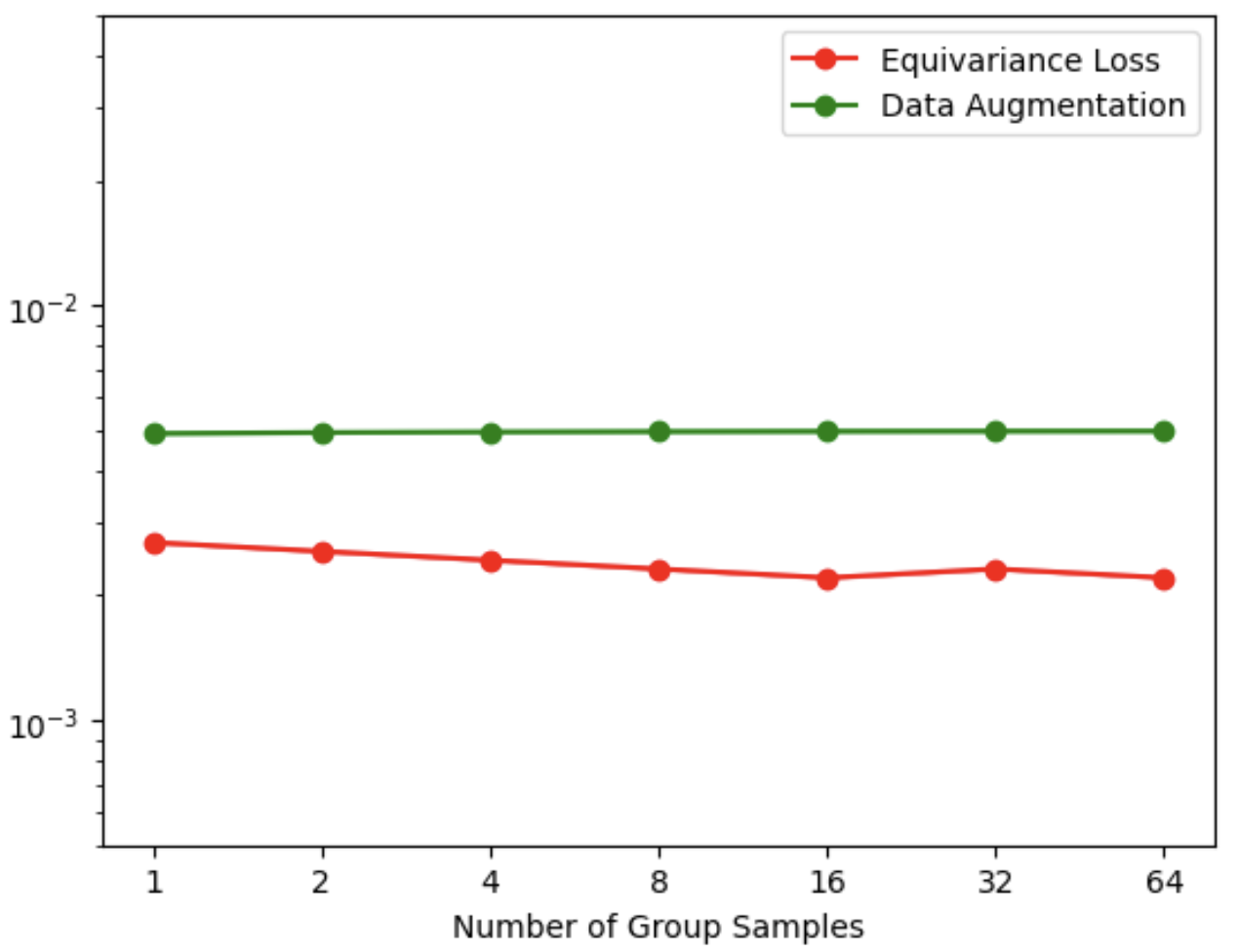} % Adjust path to your image
    }
    \hspace{-0.01cm}
    \subfigure[MSE: Out-of-distribution]{
     \label{fig:group_samples_ood}
        \includegraphics[width=0.42\textwidth]{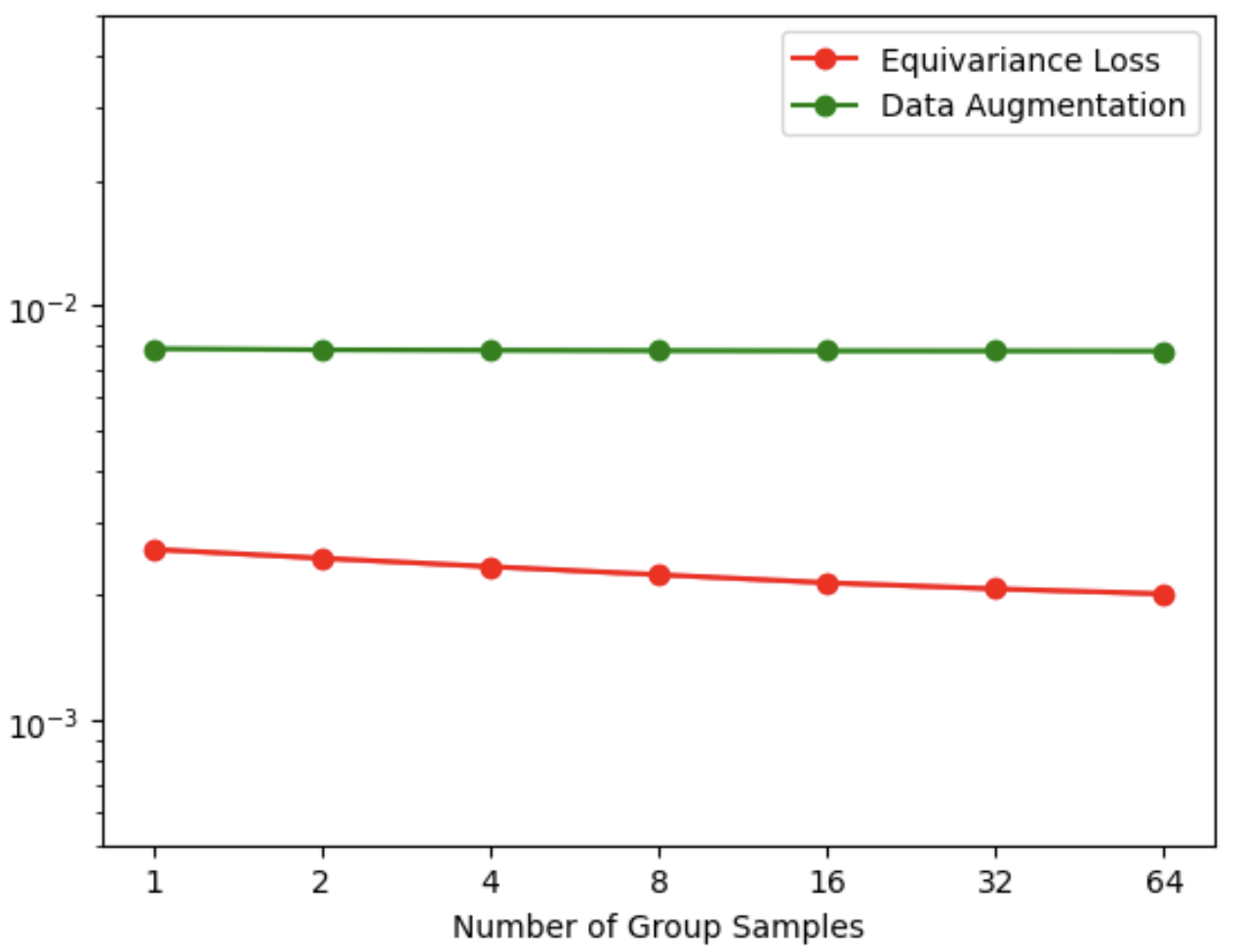} % Adjust path to your image
    }
    \caption{Performance comparison of REMUL and data augmentation on N-body dynamical system, using different numbers of samples from the symmetry group. 
    % Left: In-distribution performance and right: Out-of-distribution performance (N-body dynamical system). 
    }
    \label{fig:number of group samples}
\end{figure}

We conduct ablation studies on the number of samples required from the symmetry group during training. We compare our training procedure with data augmentation method. We selected the N-body dynamical system with the same training details and hyperparameters indicated in  Appendix \ref{appendix: n_body_system}. As shown in Figure \ref{fig:number of group samples}, REMUL achieves better performance using fewer samples from the symmetry group compared to data augmentation.
% \begin{paracol}{2}
% \sloppy 
\subsection{Convergence Speed Analysis}
\begin{figure}[t!]
    \centering
    % First figure on the left
    \subfigure[MSE: Training samples]{
        \includegraphics[width=0.42\textwidth]{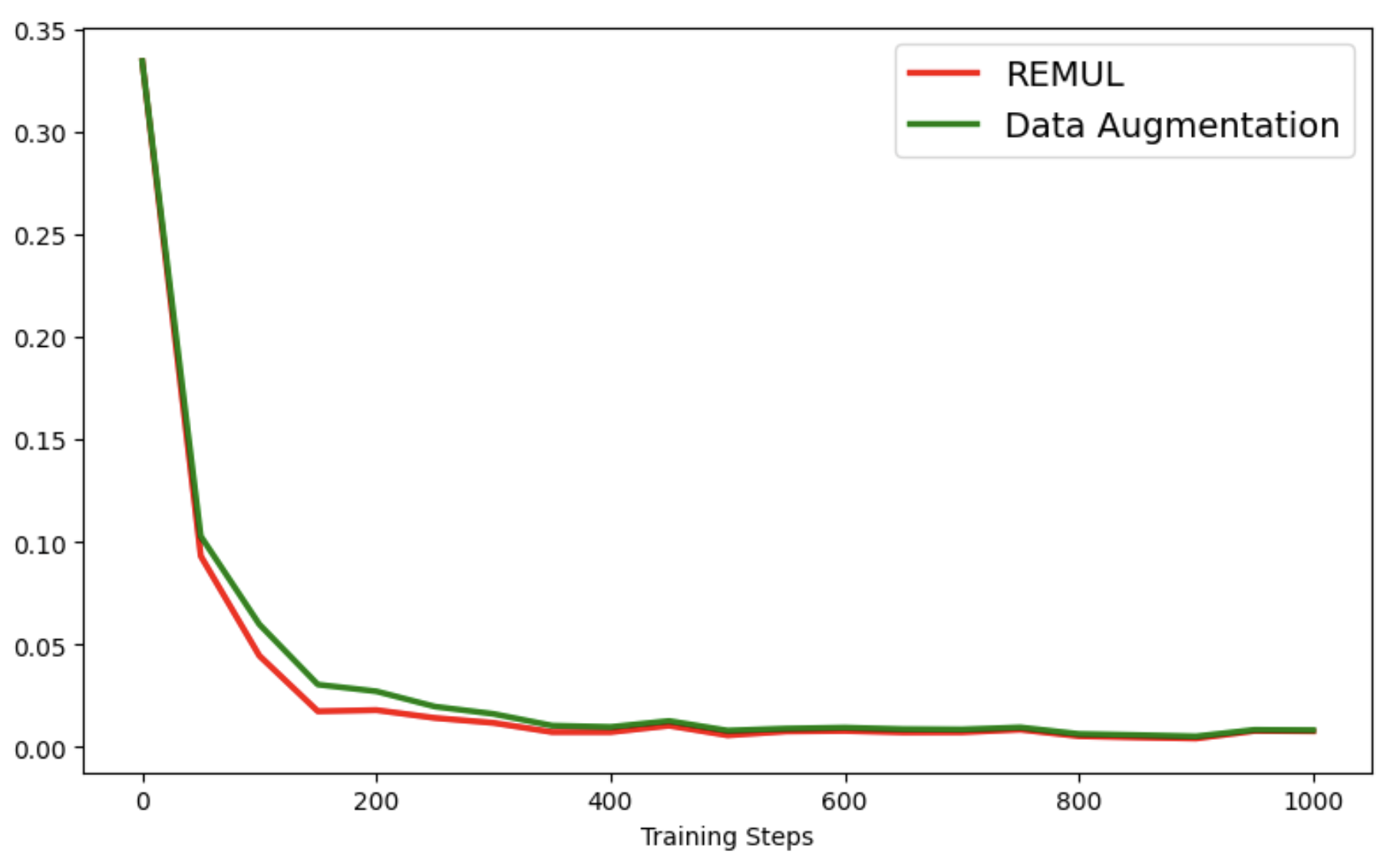} % Adjust path to your image
    }
    \hspace{-0.01cm}
    \subfigure[MSE: Validation samples]{
        \includegraphics[width=0.42\textwidth]{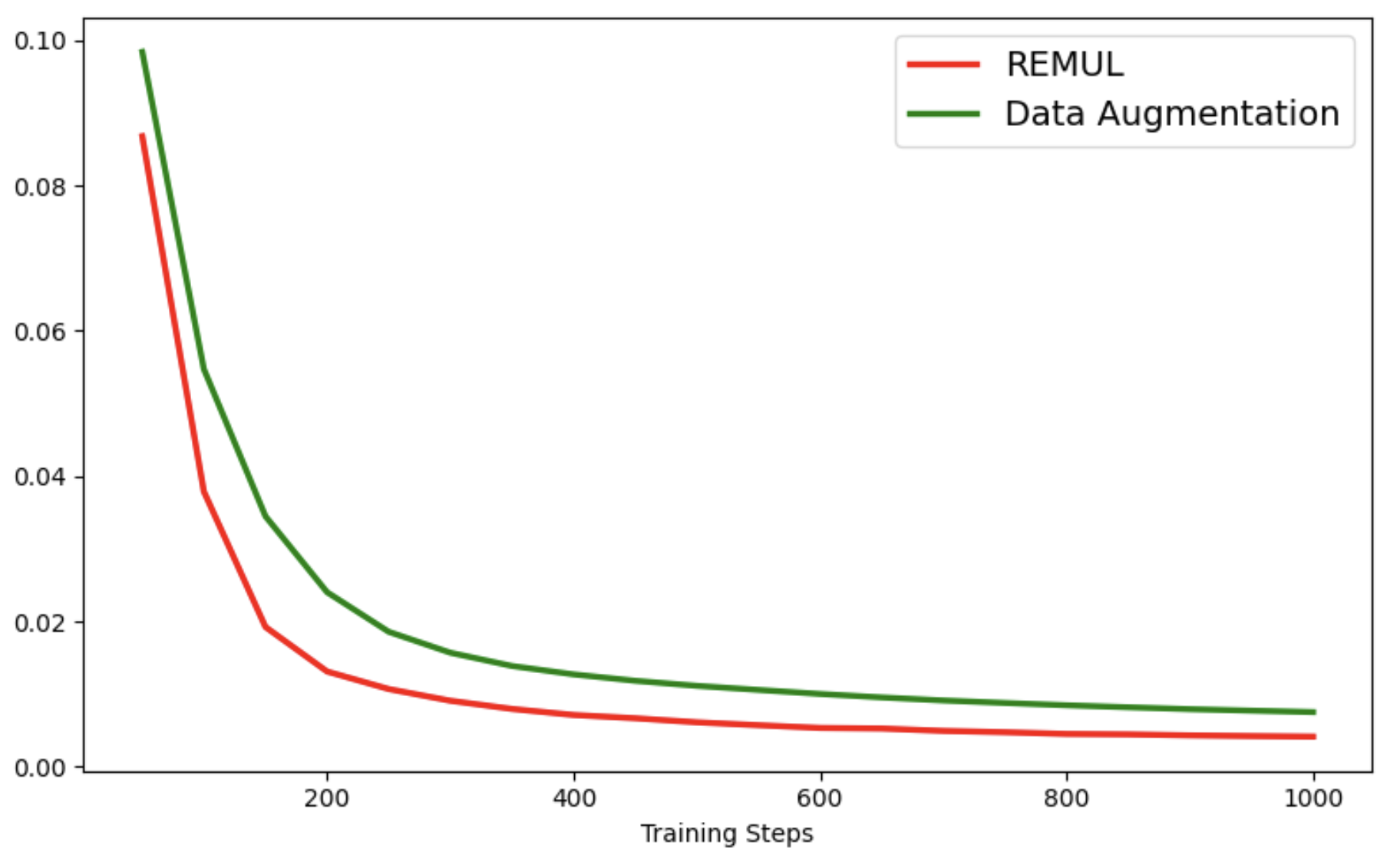} % Adjust path to your image
    }
    \caption{Comparison of convergence speed between \methodabbr{} and data augmentation on the N-body system.
    % Left: In-distribution performance and right: Out-of-distribution performance (N-body dynamical system). 
    }
    \label{fig:conv time}
\end{figure}

To assess the training efficiency of \methodabbr{} relative to data augmentation, we analyzed their convergence behavior on the N-body dynamical system. Both \methodabbr{} and DA were applied to a standard Transformer using the same experimental settings described in Appendix~\ref{appendix: n_body_system}. We report the Mean Squared Error (MSE) of the training and validation samples as a function of the training steps. 
The results, presented in Figure~\ref{fig:conv time}, indicate that \methodabbr{} achieves lower training and validation errors compared to data augmentation.
\subsection{Additional Benchmark}
% To further assess \methodabbr{}'s capabilities, we evaluated it on an additional N-body system benchmark, mainly the one in \citep{satorras2022en}. We applied \methodabbr{} to a Graph Neural Network (GNN) architecture, utilizing the same hyperparameter configurations in \citep{satorras2022en} for a fair comparison. We compare against several equivariant models, including EGNN, SEGNN \cite{brandstetter2022geometric}, FA-GNN \cite{puny2022frameaveraginginvariantequivariant}, and TFN \cite{thomas2018tensorfieldnetworksrotation}. As shown in Table~\ref{tab:additional_nbody_benchmark}, \methodabbr{} shows strong performance, outperforms EGNN and FA-GNN, and competes with SEGNN, despite the fact that the latter incorporates more specialized geometric features.
To further assess \methodabbr{}'s capabilities, we evaluated it on the N-body system benchmark from \cite{satorras2022en}. We applied \methodabbr{} to a Graph Neural Network (GNN) architecture, using the same hyperparameter configurations in \citep{satorras2022en} for a fair comparison. Table~\ref{tab:additional_nbody_benchmark} compares our approach against several equivariant models, including EGNN \citep{satorras2022en}, SEGNN \citep{brandstetter2022geometric}, FA-GNN \citep{puny2022frameaveraginginvariantequivariant}, and TFN \citep{thomas2018tensorfieldnetworksrotation}. The results demonstrate that \methodabbr{} achieves strong performance, outperforming EGNN and FA-GNN while being competitive with SEGNN, despite the latter incorporates more specialized geometric features.
% \switchcolumn
\begin{table}[ht]
\centering
% \small
\caption{Additional benchmark on N Body system.}
\label{tab:additional_nbody_benchmark}
\renewcommand{\arraystretch}{1.2}  % More spacing between rows
\begin{tabular}{lc}
\hline
& MSE \\
\hline
SE(3)-Tr & 0.0244 \\
TFN & 0.0155 \\
MPNN & 0.0107 \\
% Radial Field & 0.0104 \\
EGNN & 0.0071 \\
SEGNN & 0.0043 \\
% CGENN & 0.0039 \\
% EGNO & 0.0054 \\
FA-GNN & 0.0057 \\
REMUL-GNN & 0.0046 \\
\hline
\end{tabular}
\end{table}
% \end{paracol}
\subsection{Motion Capture}
\label{Additional Experiments: Motion Capture}
\begin{figure}[ht]
    \centering
    \subfigure[Equiv. error: Walking]{
    \label{fig:equi_error_2_walk}
        \includegraphics[width=0.42\textwidth]{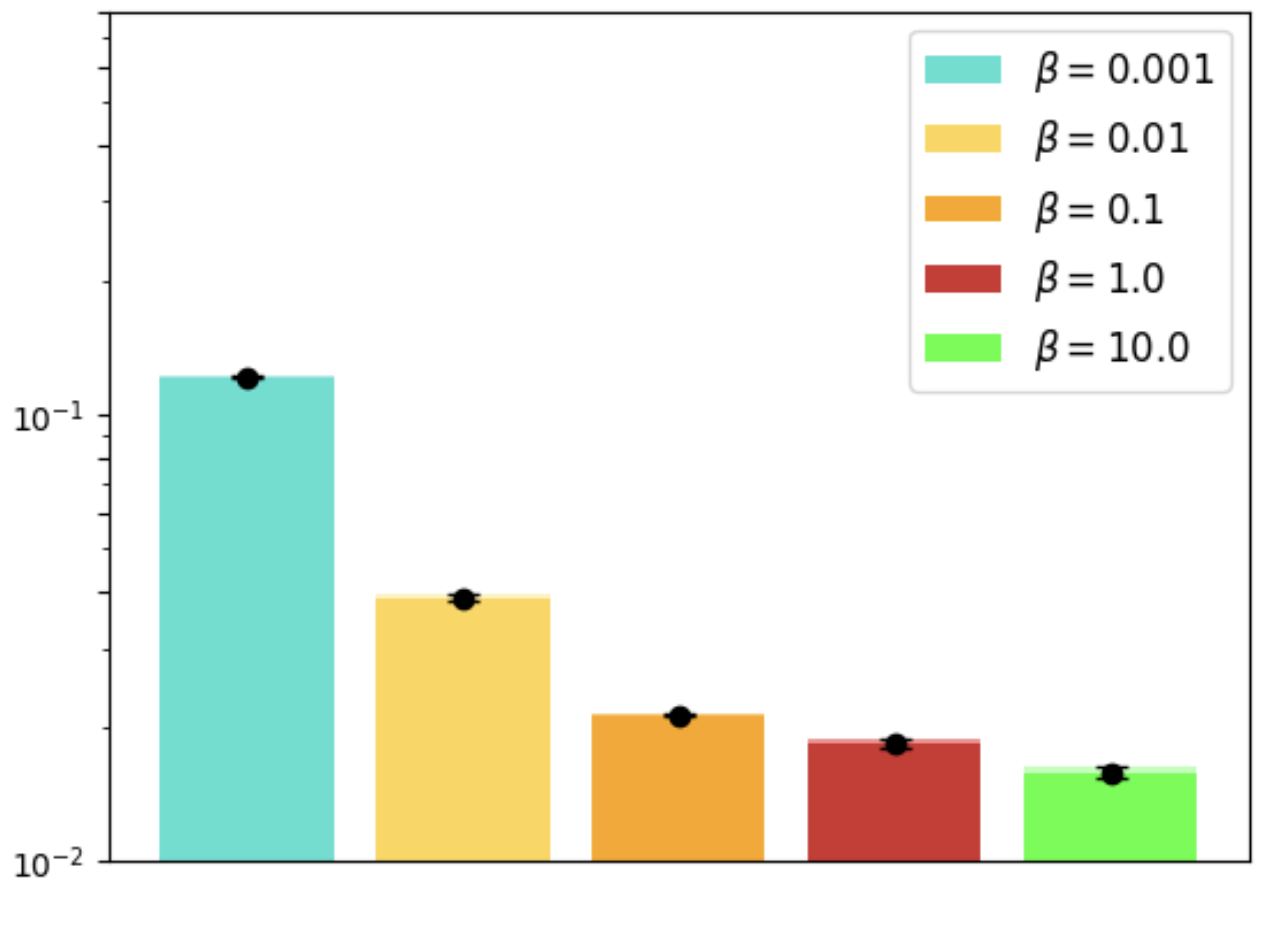}  
    }
    \hspace{-0.01cm}
    \subfigure[Equiv. error: Running]{
    \label{fig:equi_error_2_run}
        \includegraphics[width=0.42\textwidth]{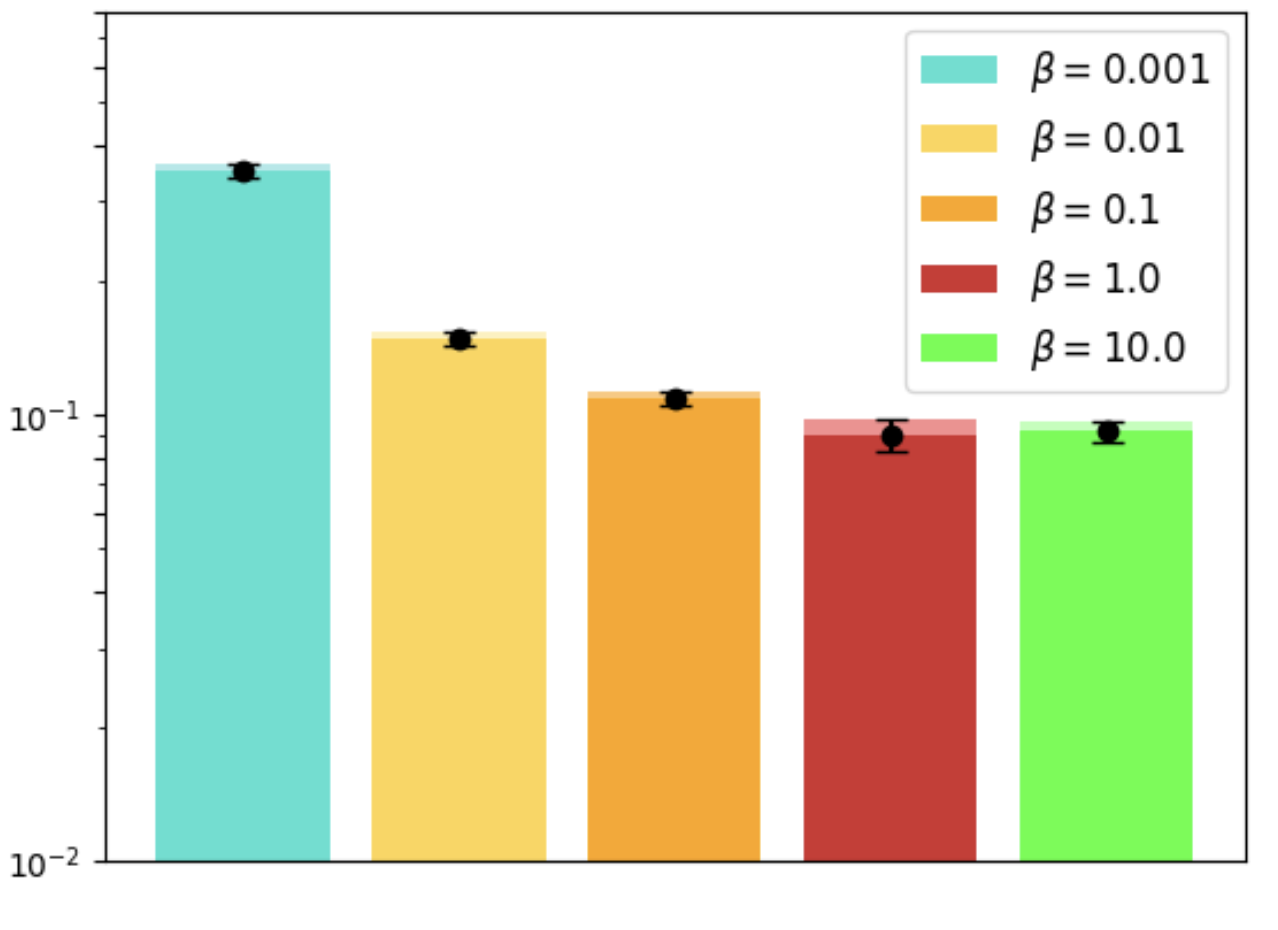} 
    }
    \caption{Motion Capture: Transformer trained with REMUL. The second equivariance measure $E'$.
    % Equation \ref{eq:equi_measure_2}.
    % Left:  Walking task (Subject \#$35$) and right: Running task (Subject \#$9$).
    }
    \label{fig:Motion capture dataset: Second equivariance measure}
\end{figure}
In the main paper (Section~\ref{sec: motion capture}) we note that human motion may lack full SO(3) symmetry, particularly along the vertical (gravity) axis. To investigate this further, we measured the equivariance error $E$ separately for rotations applied around $X$, $Y$, and $Z$ axes. We use the best performing \methodabbr{}-Transformer models on Motion Capture dataset (specifically, $\beta=0.1$ for the Walking task and $\beta=0.01$ for the Running task). 
The results are presented in 
Table~\ref{tab:seperate equi error motion_capture}. For both Walking and Running tasks, the equivariance error associated with rotations around the Z-axis is higher than the errors for $X$-axis and $Y$-axis, which supports that the underlying dynamics in the Motion Capture exhibit a lesser degree of symmetry \textit{w.r.t.} Z-axis, and aligns with our observation that models with relaxed equivariance (intermediate $\beta$) perform best on this task.
\begin{table}[ht]
\centering
\caption{Motion Capture: Equivariance error around different $X$, $Y$, and $Z$ axis separately.}
\label{tab:seperate equi error motion_capture}
\begin{tabular}{ccc}
\noalign{\hrule height 0.8pt}
& Walking & Running \\
\noalign{\hrule height 0.8pt}
X & 0.0047 & 0.026 \\
Y & 0.0034 & 0.031 \\
Z & 0.0084 & 0.042 \\
\noalign{\hrule height 0.8pt}
\end{tabular}
\end{table}

\newpage
\subsection{Molecular Dynamics}
\begin{figure}[ht]
\centering
\subfigure{
    \includegraphics[width=0.4\textwidth]{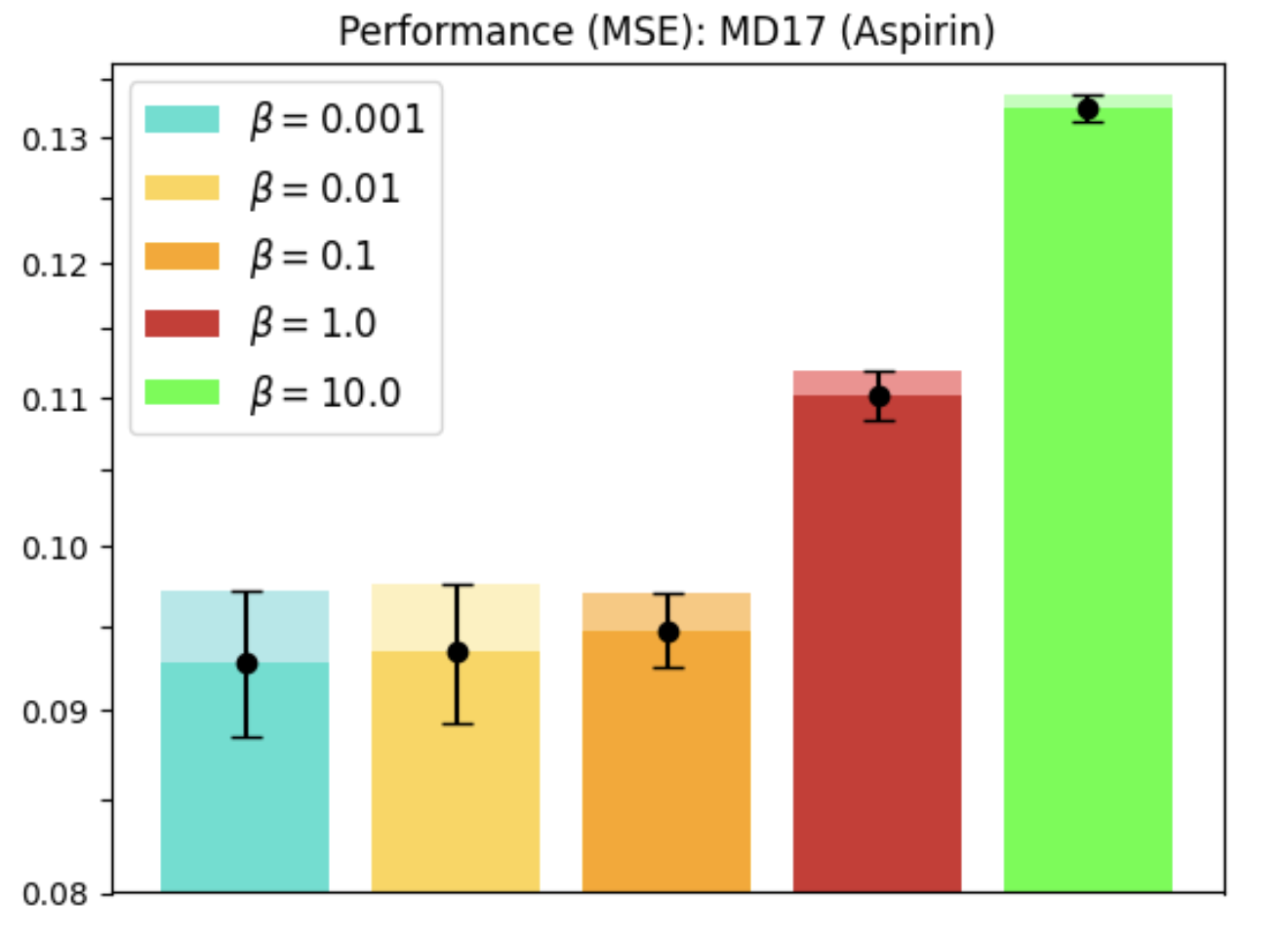}
}
\hspace{-0.01cm}
\subfigure{
    \includegraphics[width=0.4\textwidth]{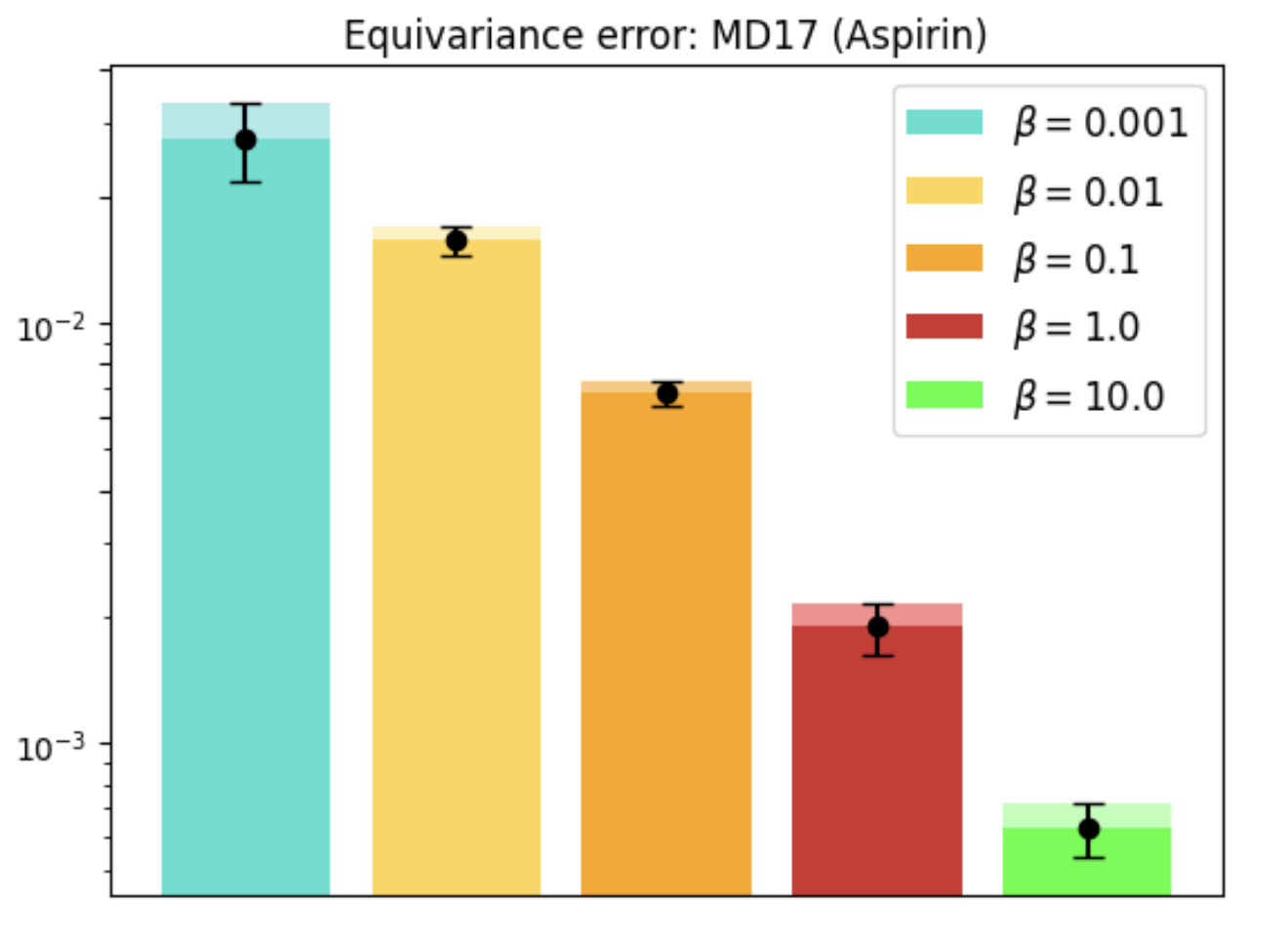}
}
% \hspace{-0.01cm}
% mse_ethanol

\subfigure{
\hspace{-0.012\linewidth}
    \includegraphics[width=0.4\textwidth]{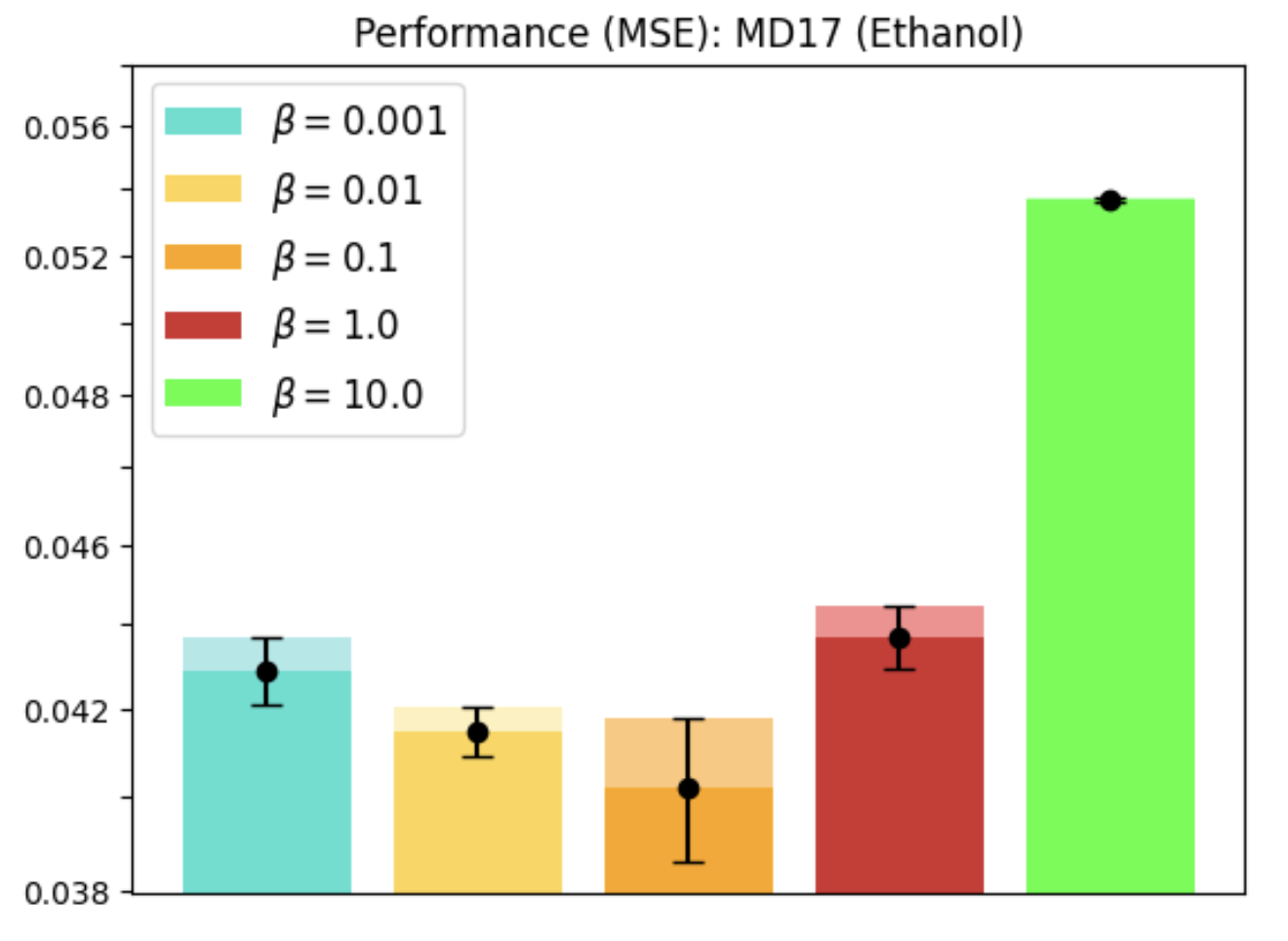}
}
% \hspace{-0.01cm}
\subfigure{
\hspace{0.003\linewidth}
    \includegraphics[width=0.4\textwidth]{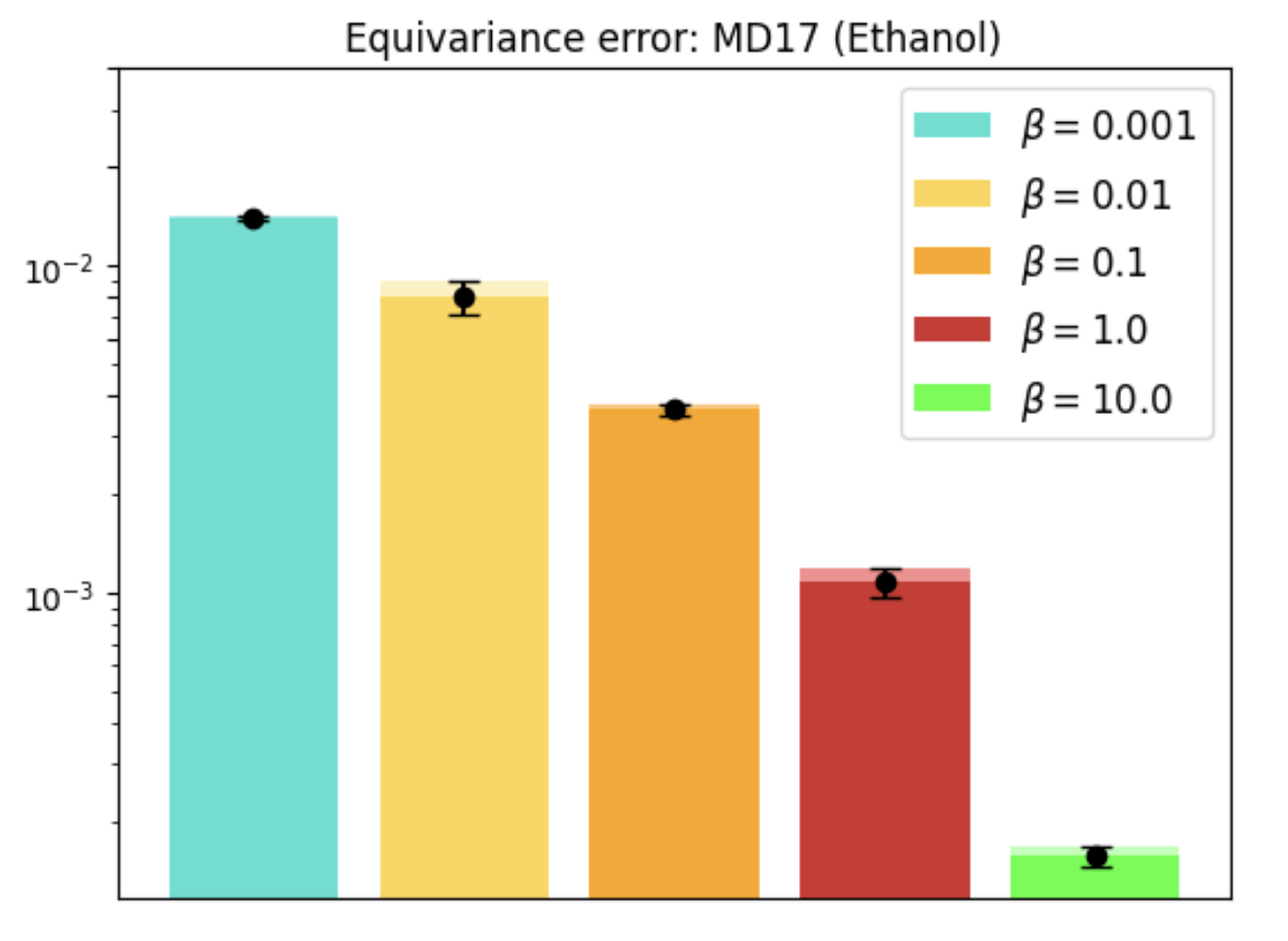}
}
\subfigure{
    \includegraphics[width=0.4\textwidth]{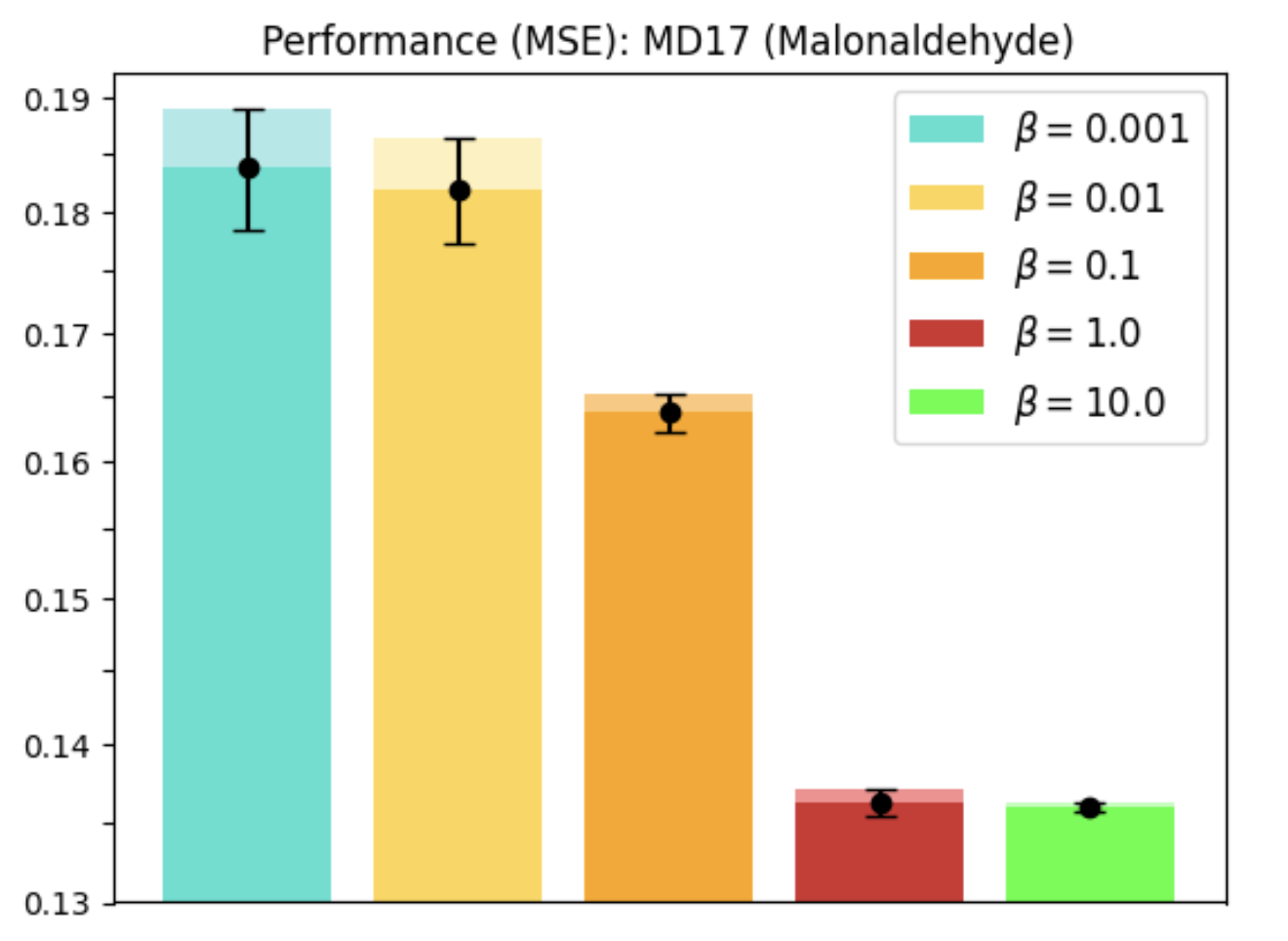}
}
\hspace{-0.01cm}
\subfigure{
    \includegraphics[width=0.4\textwidth]{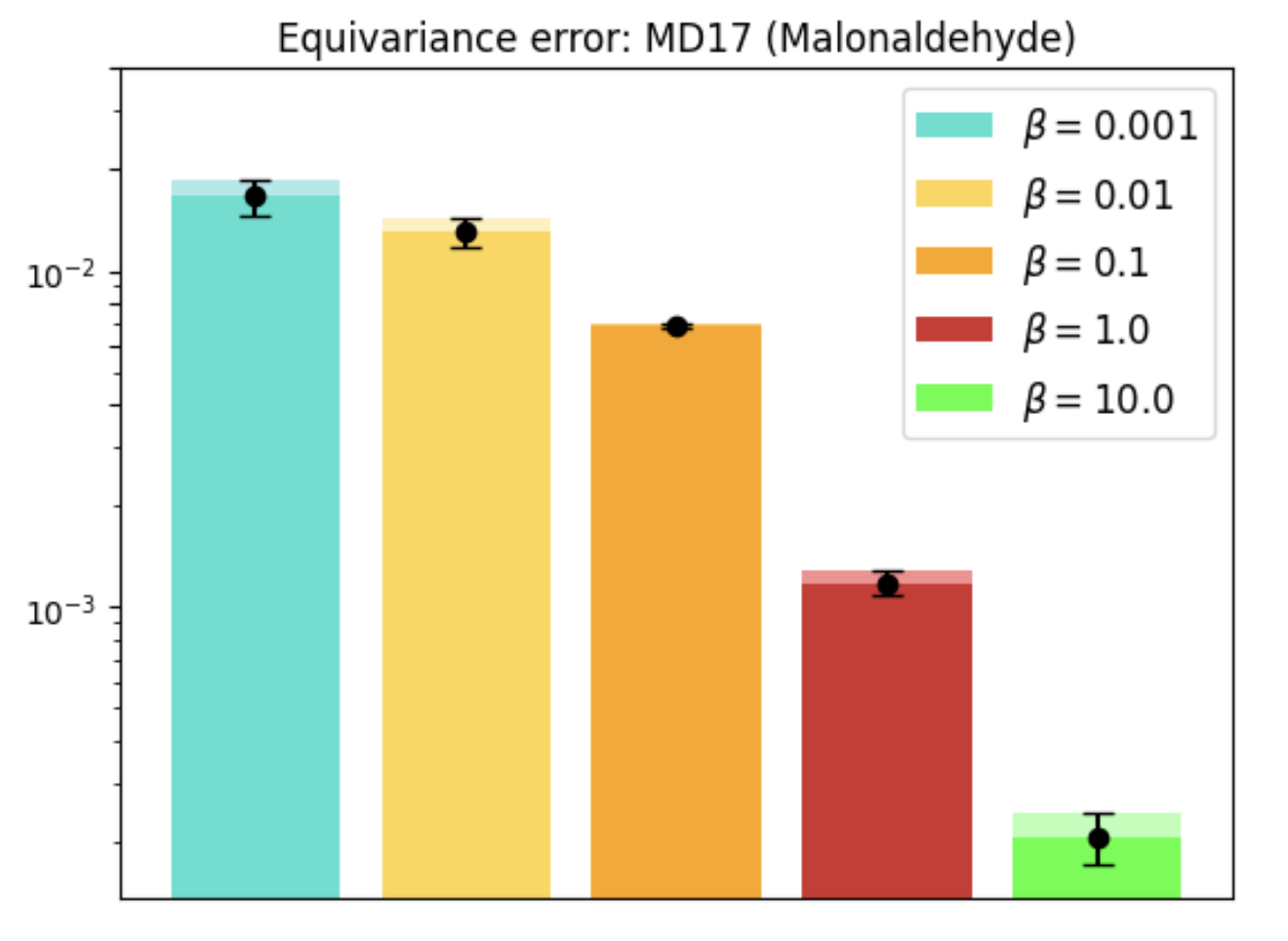}
}
% \hspace{-0.01cm}
\subfigure{
\hspace{-0.004\linewidth}
    \includegraphics[width=0.4\textwidth]{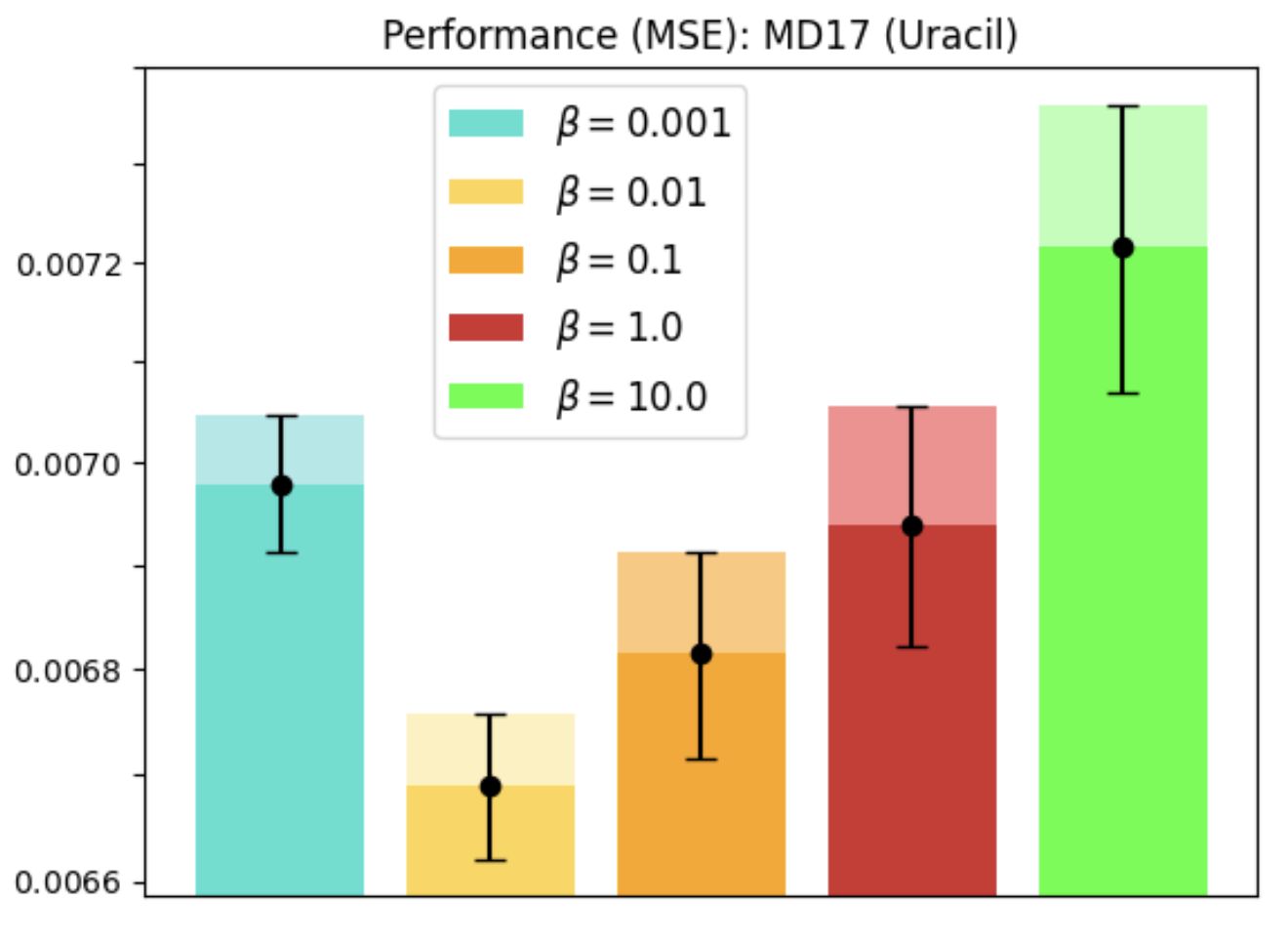}
}
\hspace{0.008\linewidth}
\subfigure{
    \includegraphics[width=0.4\textwidth]{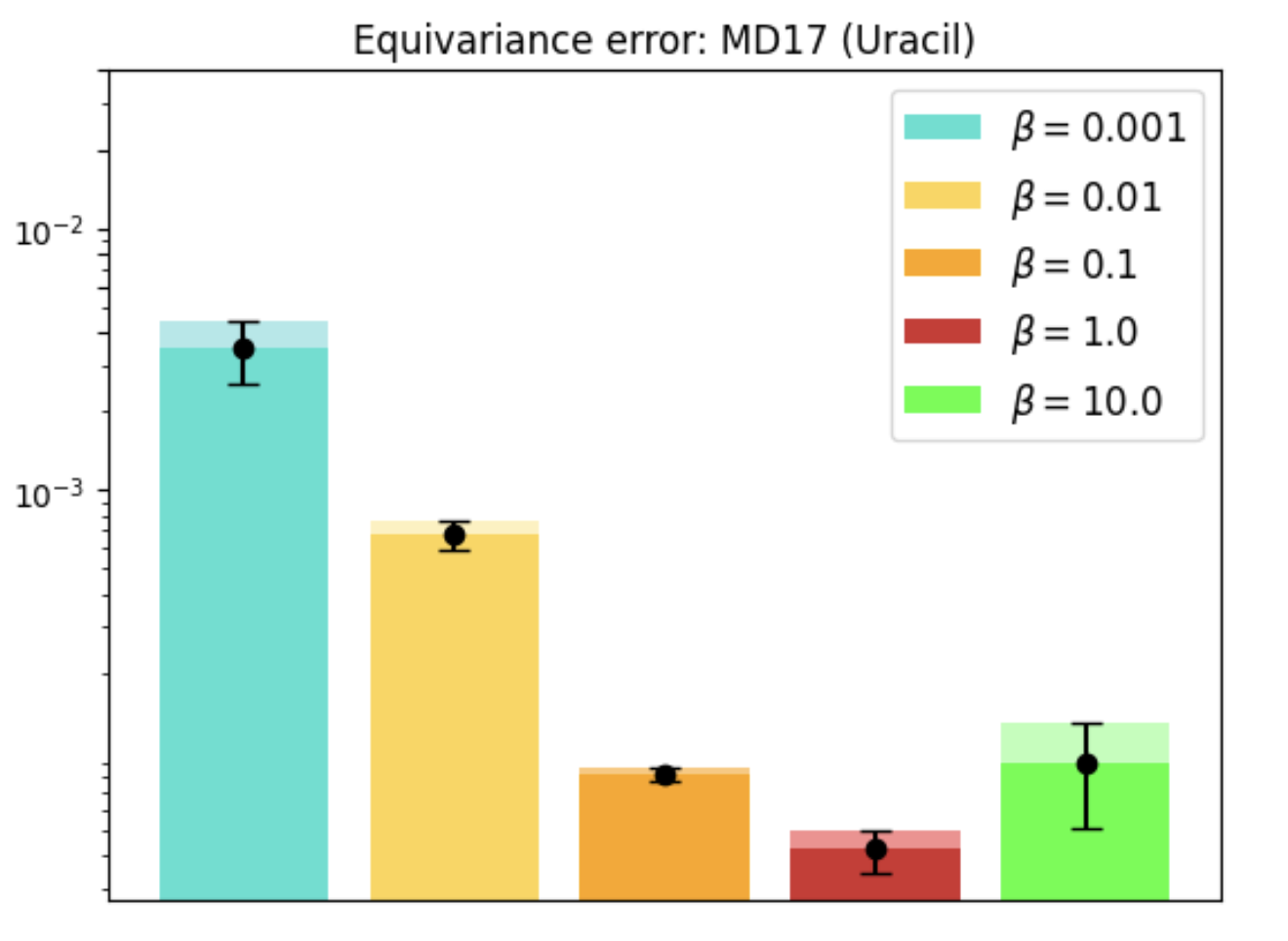}
}
\caption{MD17 dataset: GNN trained with REMUL. The first column is model performance (MSE), and the second column is equivariance error $E$. Rows from top to bottom represent Aspirin, Ethanol, Malonaldehyde, and Uracil, respectively.
The equivariance error decreases on all molecules with a higher value of $\beta$. In contrast, the required equivariance for best model performance varies for each molecule.}
\label{fig:md17}
\end{figure}

\begin{figure}[ht]
\centering
% Each subfigure set to take up half the text width
\subfigure{
    \includegraphics[width=0.4\textwidth]{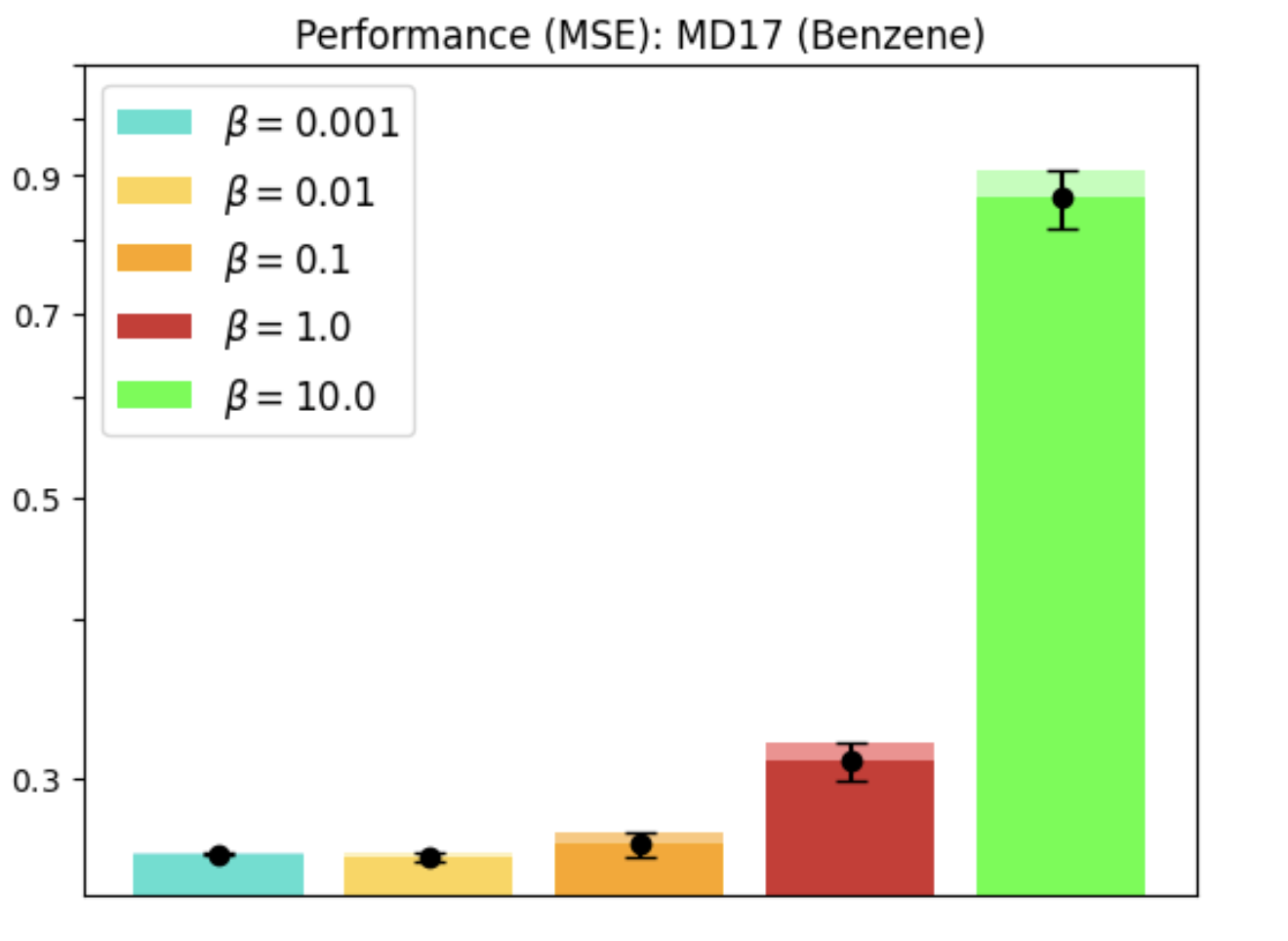}
    % \caption{mse_benzene}
    % \label{fig:mse_benzene}
}
\hspace{-0.01cm}
\subfigure{
    \includegraphics[width=0.4\textwidth]{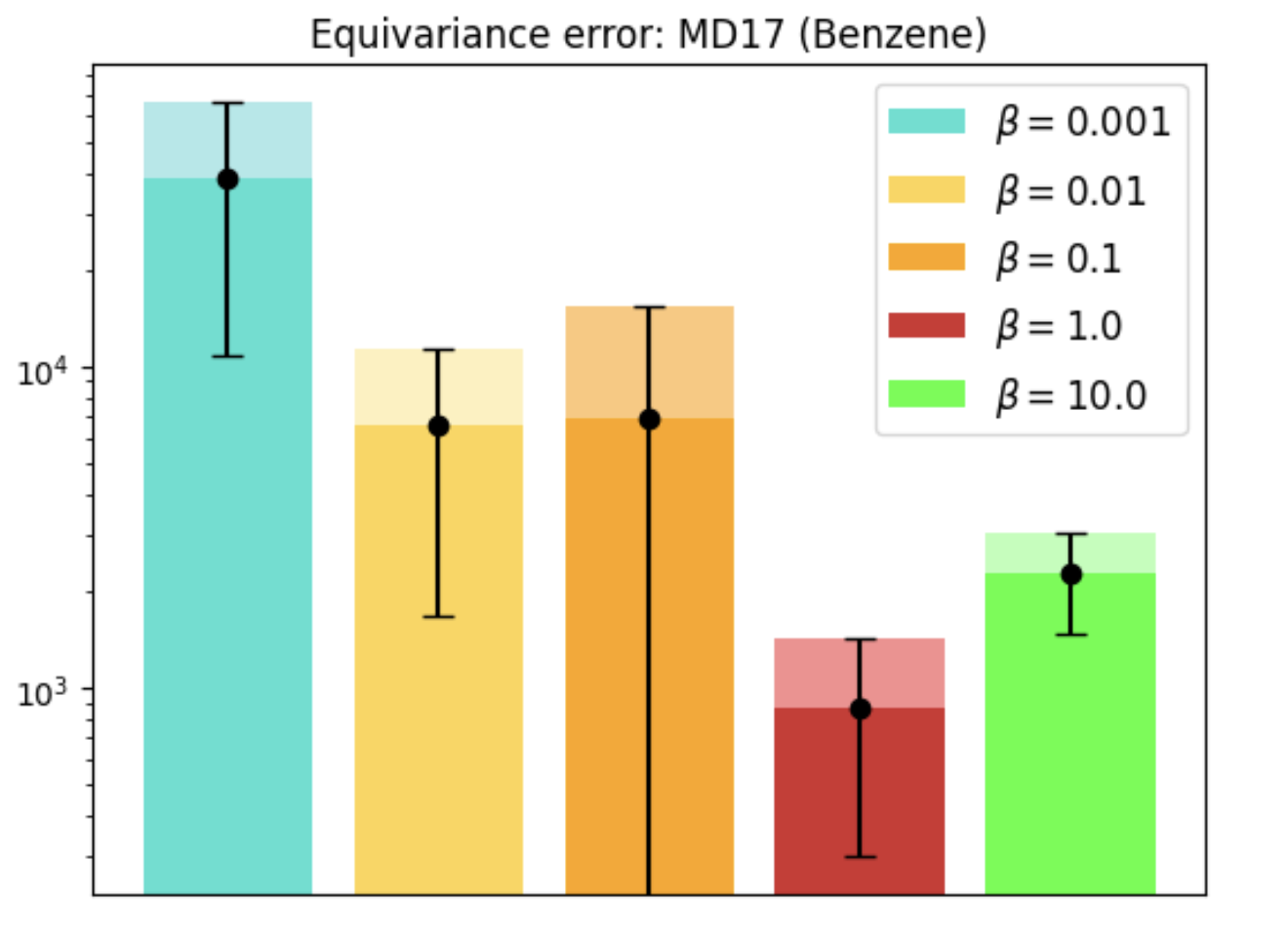}
    % \caption{m1_benzene}
    % \label{fig:m1_benzene}
}

\subfigure{
    \hspace{-0.028\linewidth} % or any other dimension that looks good
    \includegraphics[width=0.4\textwidth]{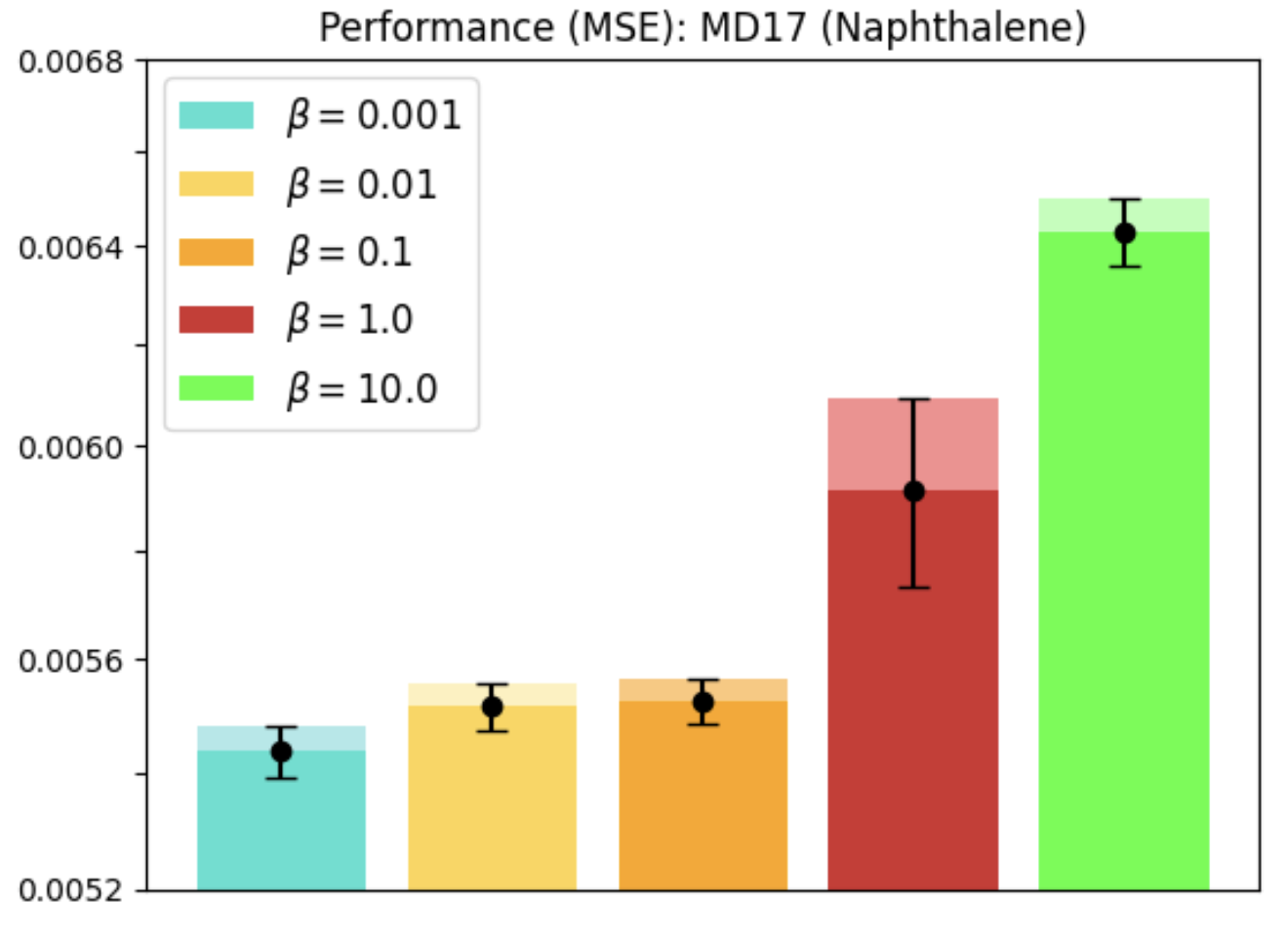}
    % \caption{mse_naphthalene}
    % \label{fig:mse_naphthalene}
}
\subfigure{
    \hspace{0.01\linewidth} % or any other dimension that looks good
    \includegraphics[width=0.4\textwidth]{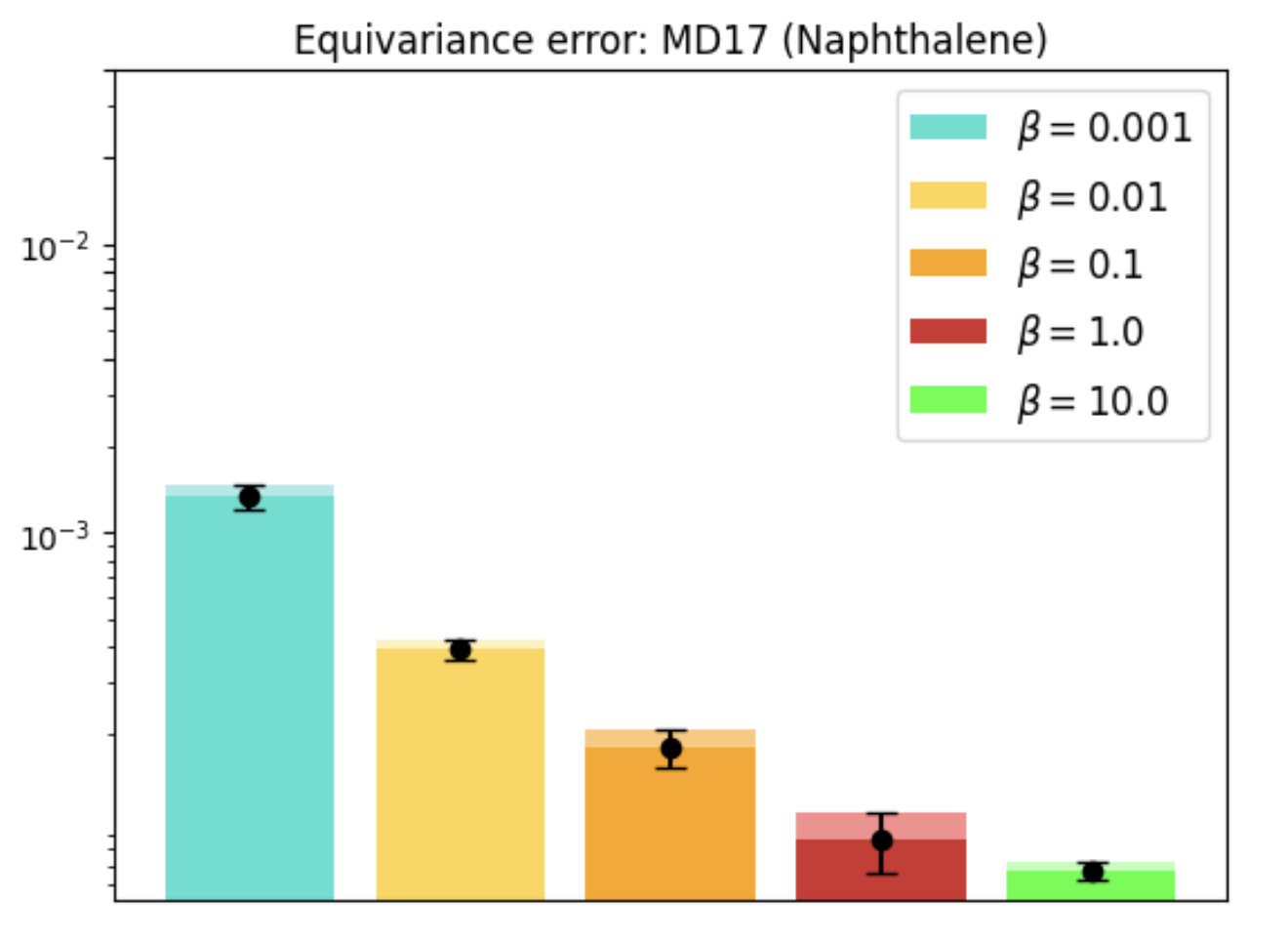}
    % \caption{mse_naphthalene}
    % \label{fig:mse_naphthalene}
}

\subfigure{
\hspace{-0.028\linewidth}
    \includegraphics[width=0.4\textwidth]{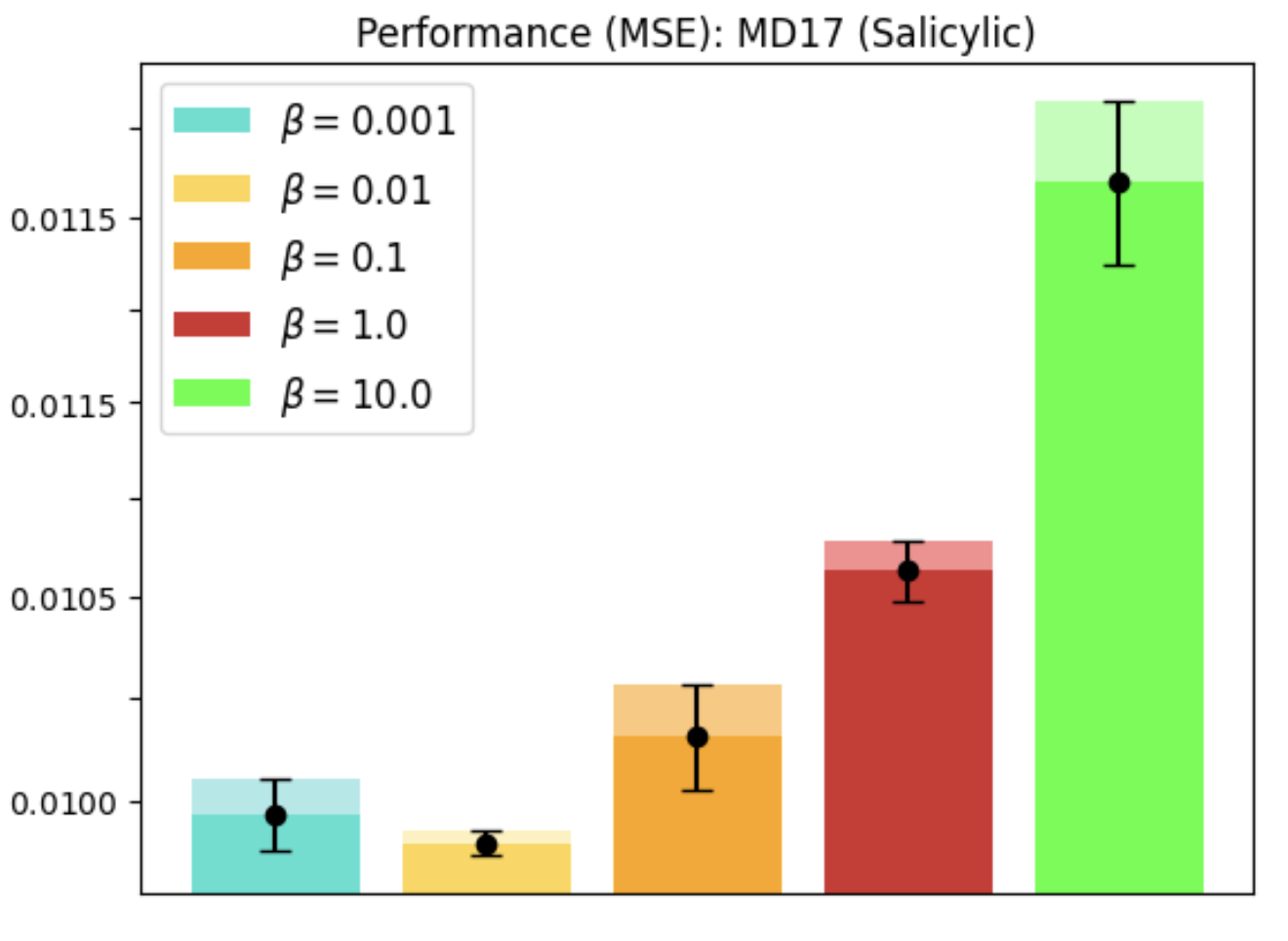}
    % \caption{mse_salicylic}
    % \label{fig:mse_salicylic}
}
% \hspace{-0.01cm}
\subfigure{
\hspace{0.014\linewidth}
    \includegraphics[width=0.4\textwidth]{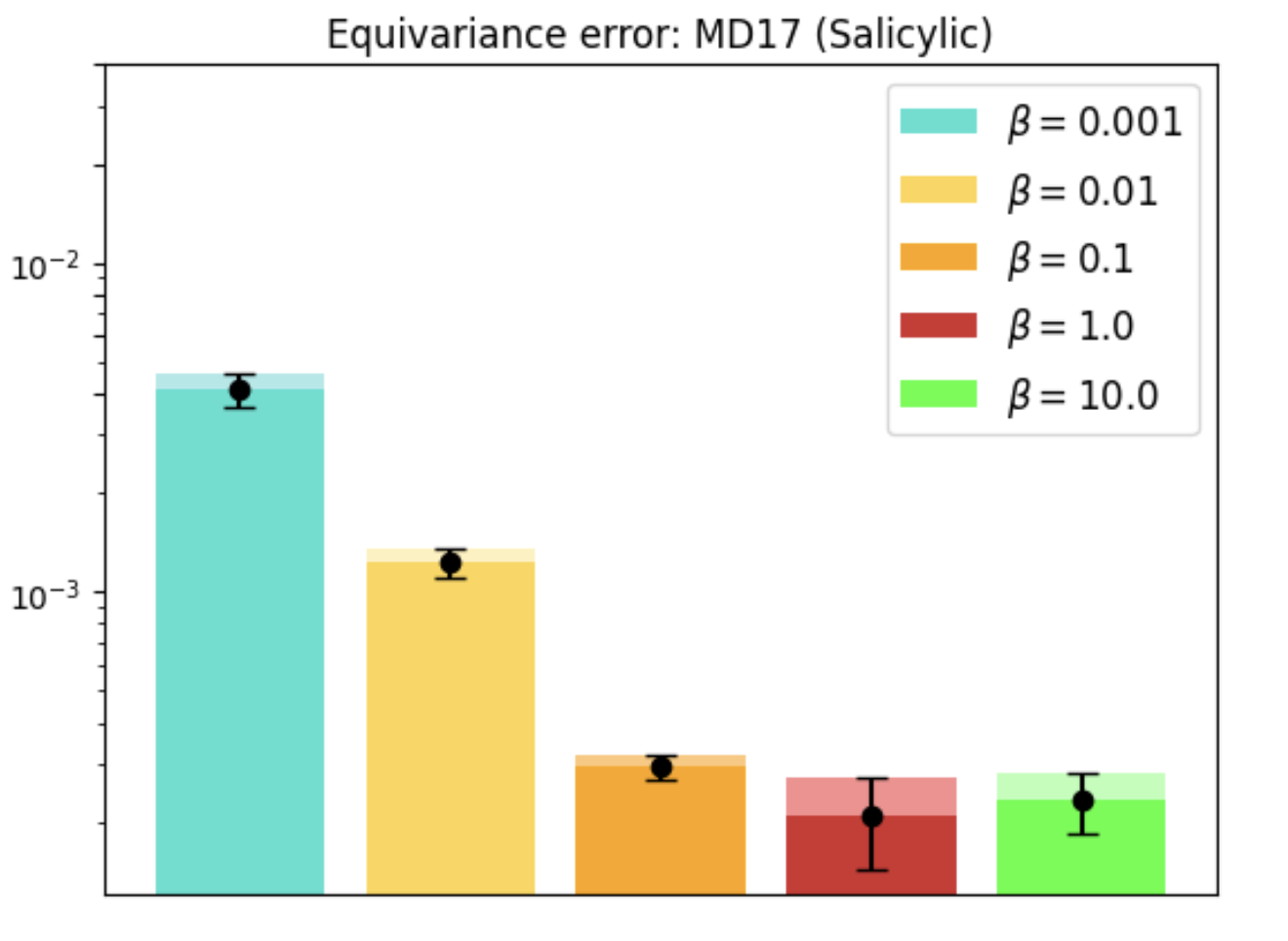}
    % \caption{m1_salicylic}
    % \label{fig:m1_salicylic}
}
\subfigure{
\hspace{-0.018\linewidth}
% \hspace{0.1cm}
    \includegraphics[width=0.4\textwidth]{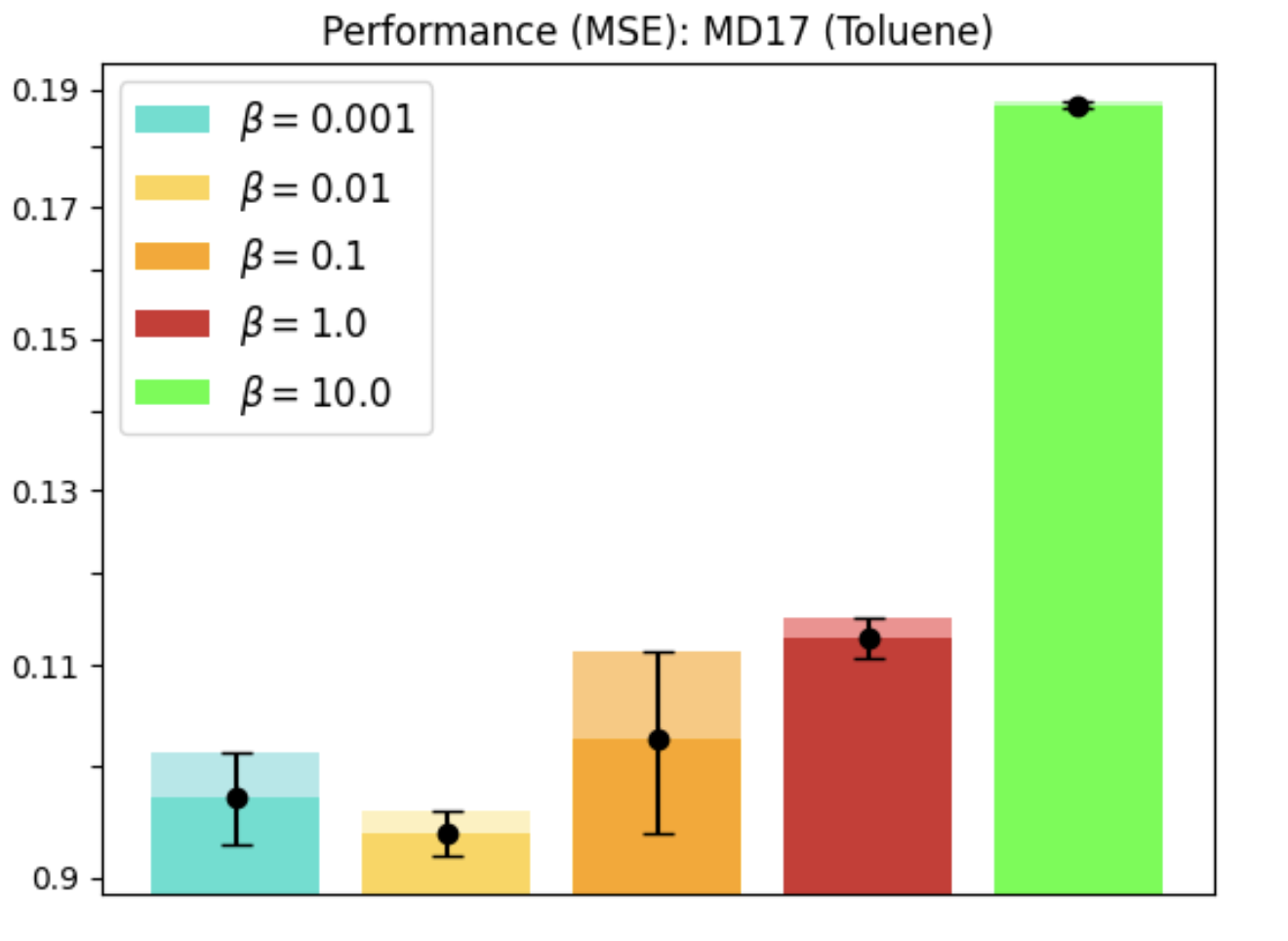}
    % \caption{mse_toluene}
    % \label{fig:mse_toluene}
}
% \hspace{-0.01cm}
\subfigure{
\hspace{0.003\linewidth}
    \includegraphics[width=0.4\textwidth]{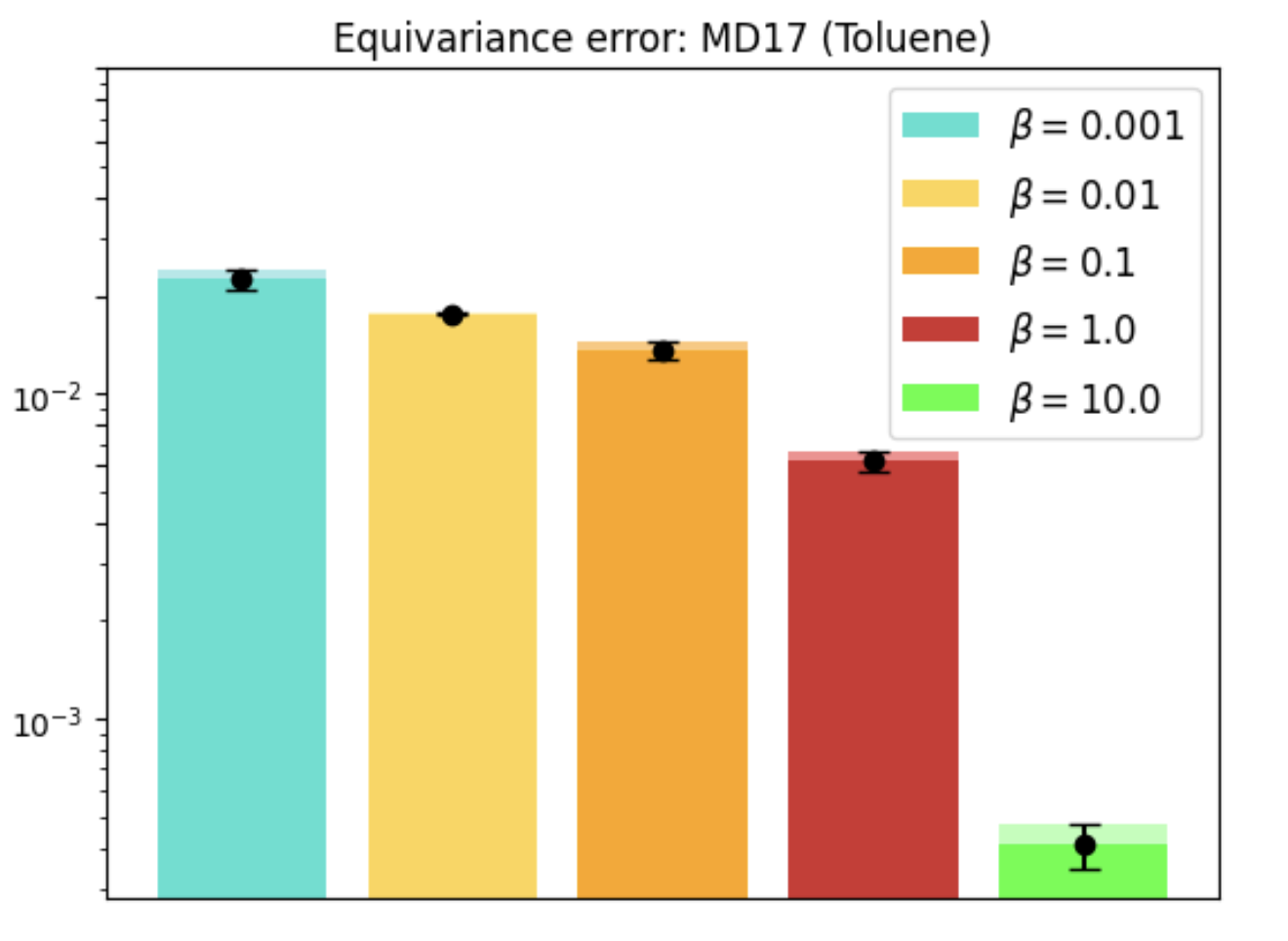}
    % \caption{m1_toluene}
    % \label{fig:m1_toluene}
}

\caption{MD17 dataset: GNN trained with REMUL. The first column is model performance (MSE), and the second column is equivariance error $E$.
Rows from top to bottom represent Benzene, Naphthalene, Salicylic, and Toluene, respectively.}
\label{fig:Additional MD17}
\end{figure}

\begin{figure}[ht]
\centering
% Each subfigure set to take up half the text width
\subfigure{
    \includegraphics[width=0.4\textwidth]{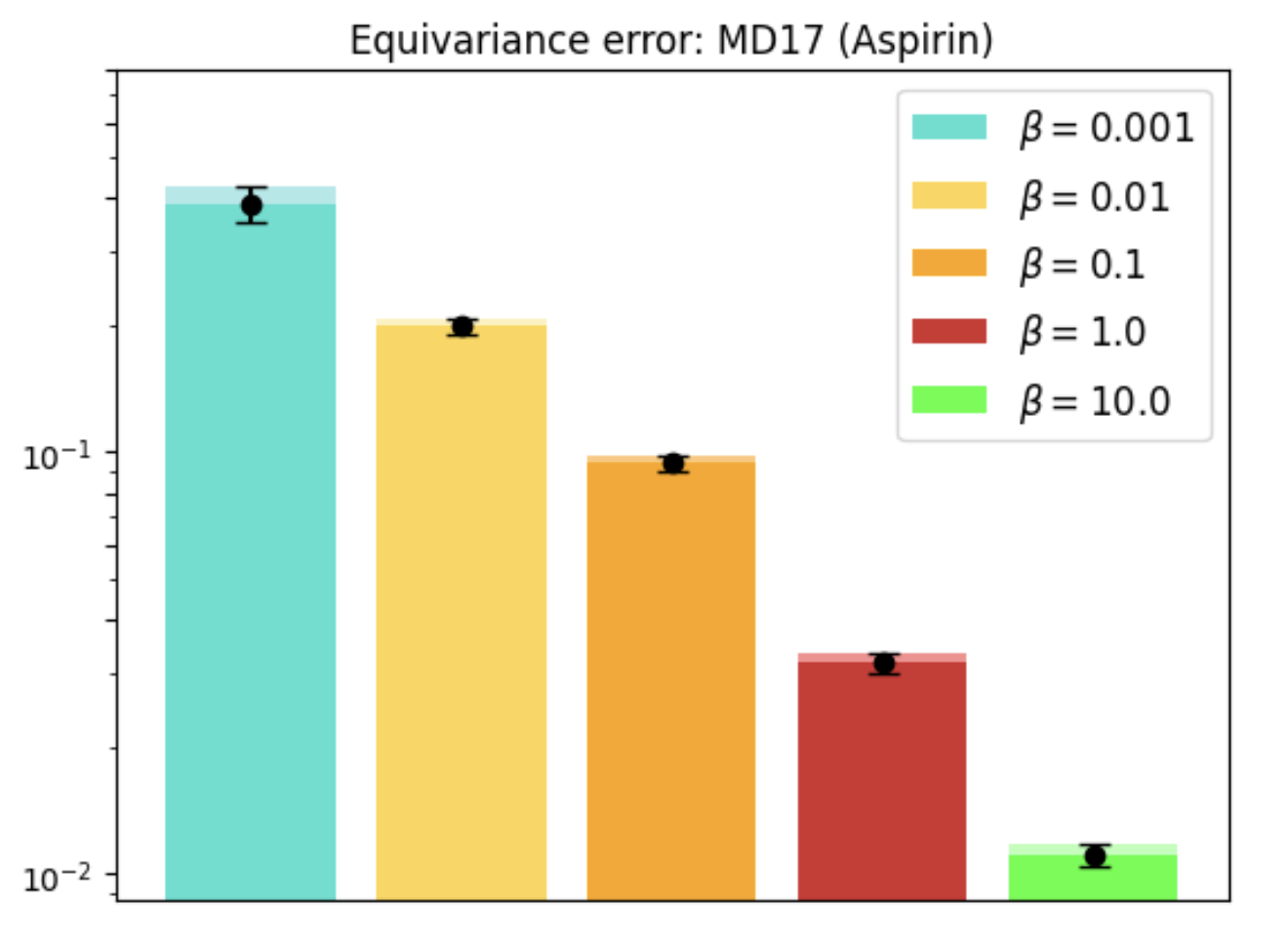}
    % \caption{m2_aspirin}
    % \label{fig:m2_aspirin}
}
\hspace{-0.01cm}
\subfigure{
    \includegraphics[width=0.4\textwidth]{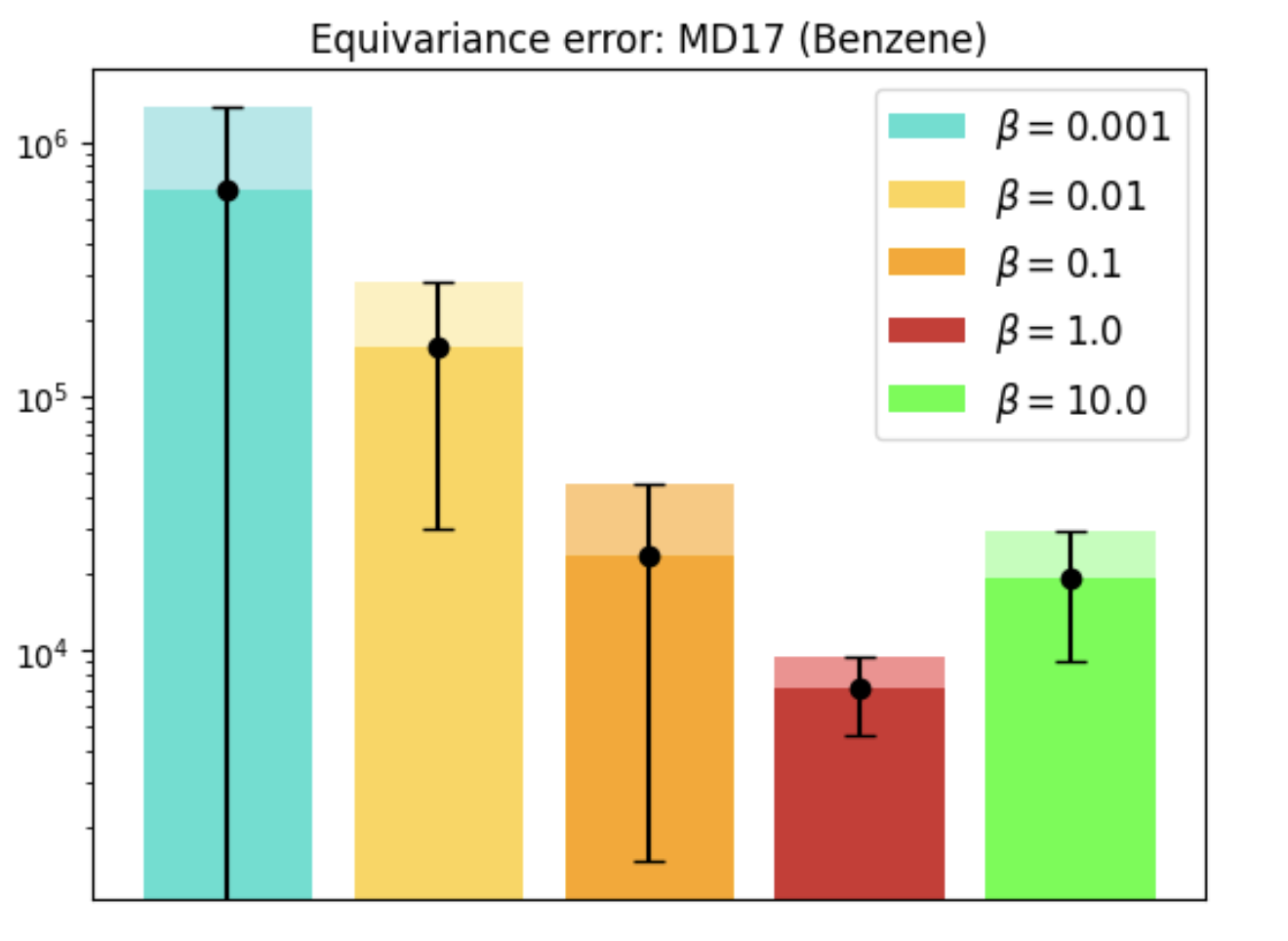}
    % \caption{m2_benzene}
    % \label{fig:m2_benzene}
}

\subfigure{
    \includegraphics[width=0.4\textwidth]{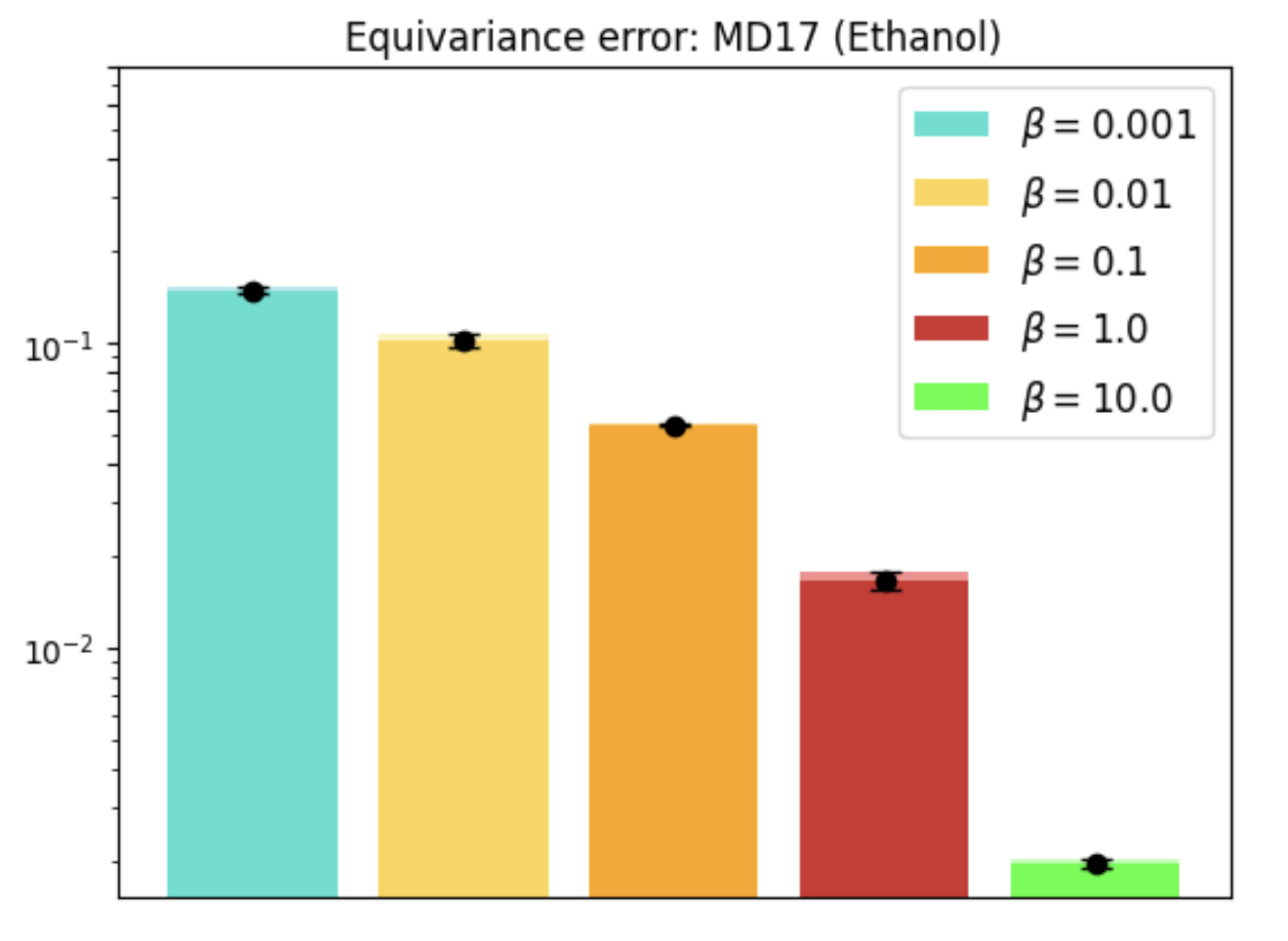}
    % \caption{m2_ethanol}
    % \label{fig:m2_ethanol}
}
\hspace{-0.01cm}
\subfigure{
    \includegraphics[width=0.4\textwidth]{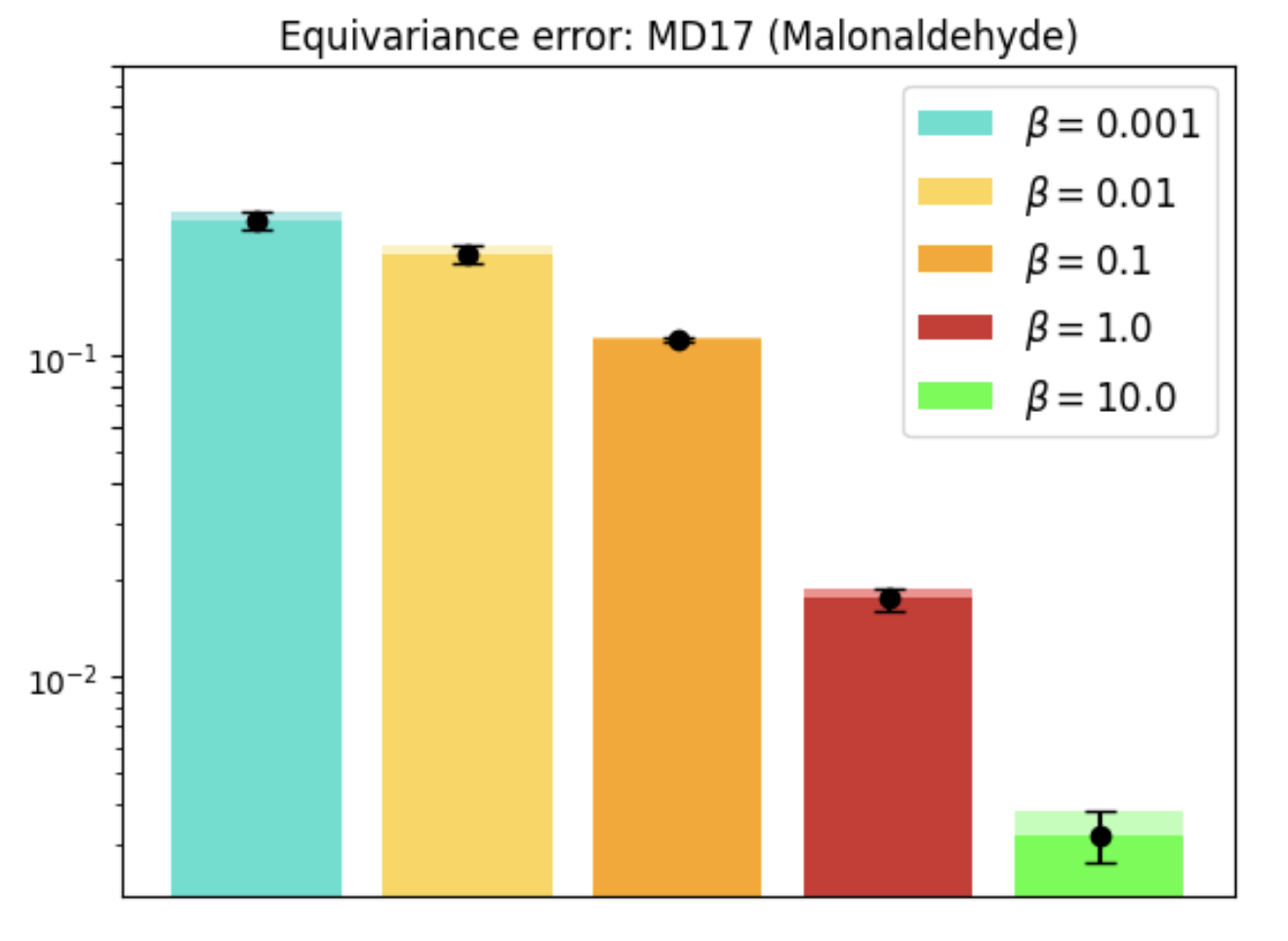}
    % \caption{m2_malonaldehyde}
    % \label{fig:m2_malonaldehyde}
}

\subfigure{
    \includegraphics[width=0.4\textwidth]{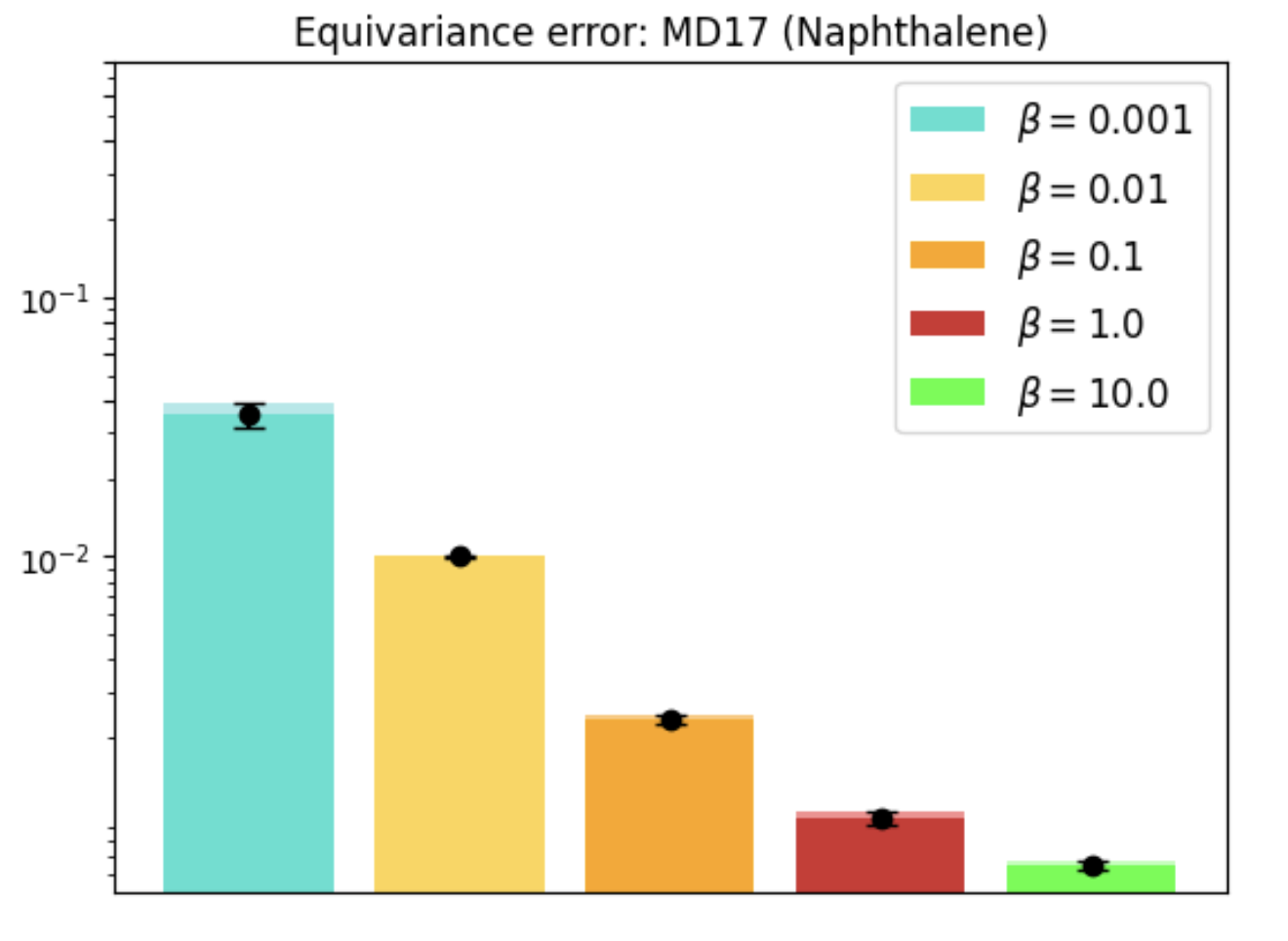}
    % \caption{m2_naphthalene}
    % \label{fig:m2_naphthalene}
}
\hspace{-0.01cm}
\subfigure{
    \includegraphics[width=0.4\textwidth]{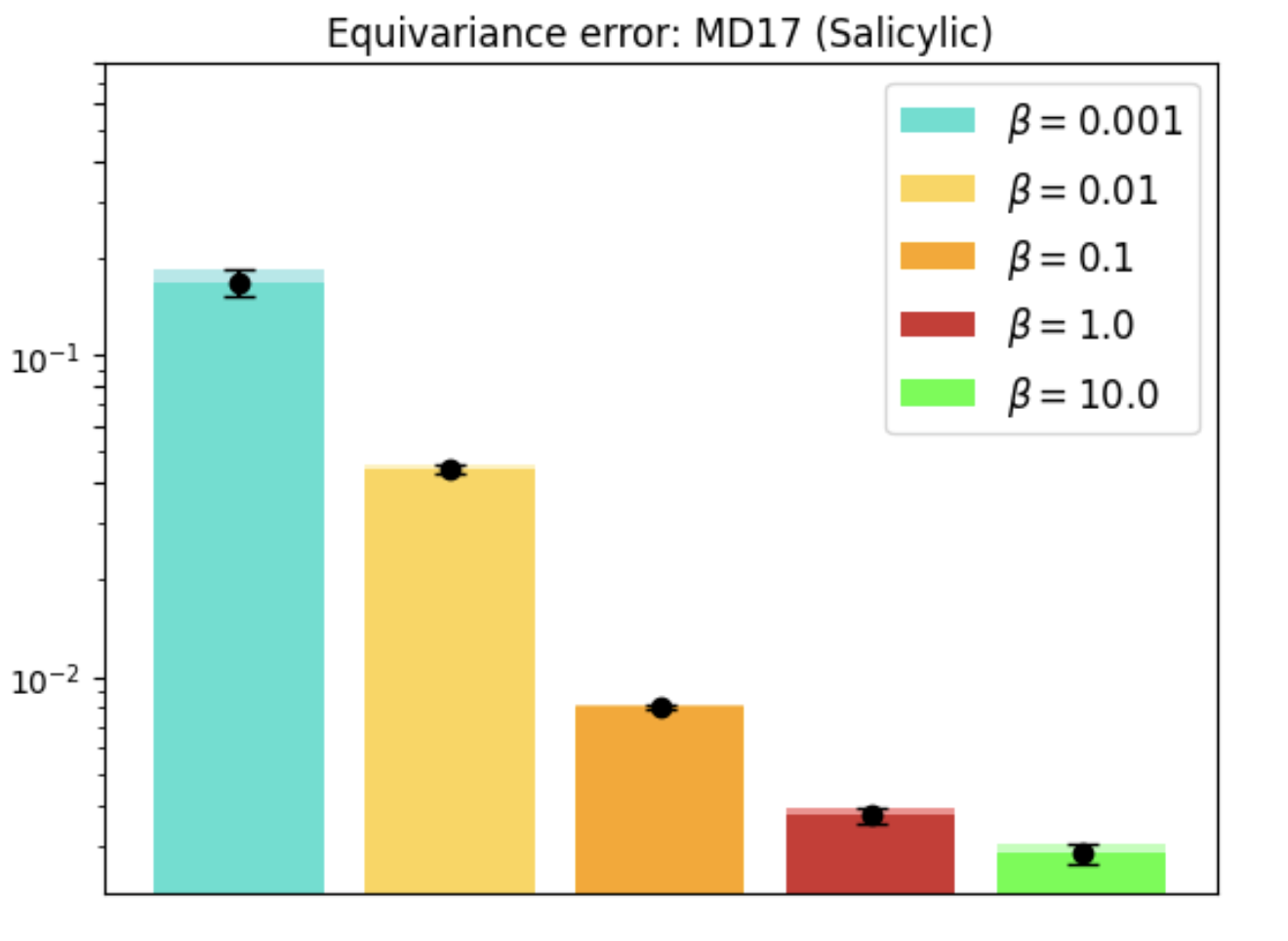}
    % \caption{m2_salicylic}
    % \label{fig:m2_salicylic}
}
\subfigure{
    \includegraphics[width=0.4\textwidth]{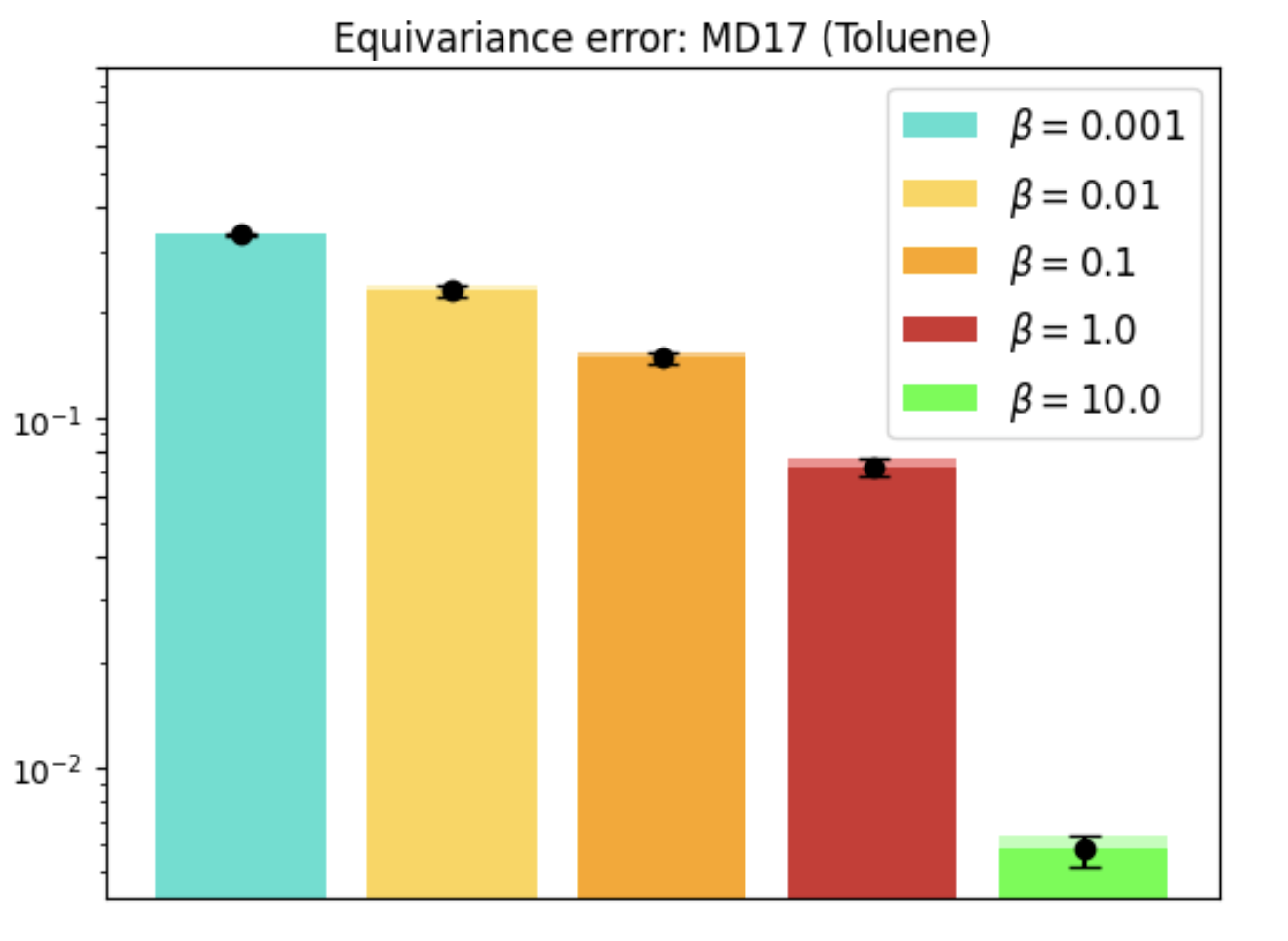}
    % \caption{m2_toluene}
    % \label{fig:m2_toluene}
}
\hspace{-0.01cm}
\subfigure{
    \includegraphics[width=0.4\textwidth]{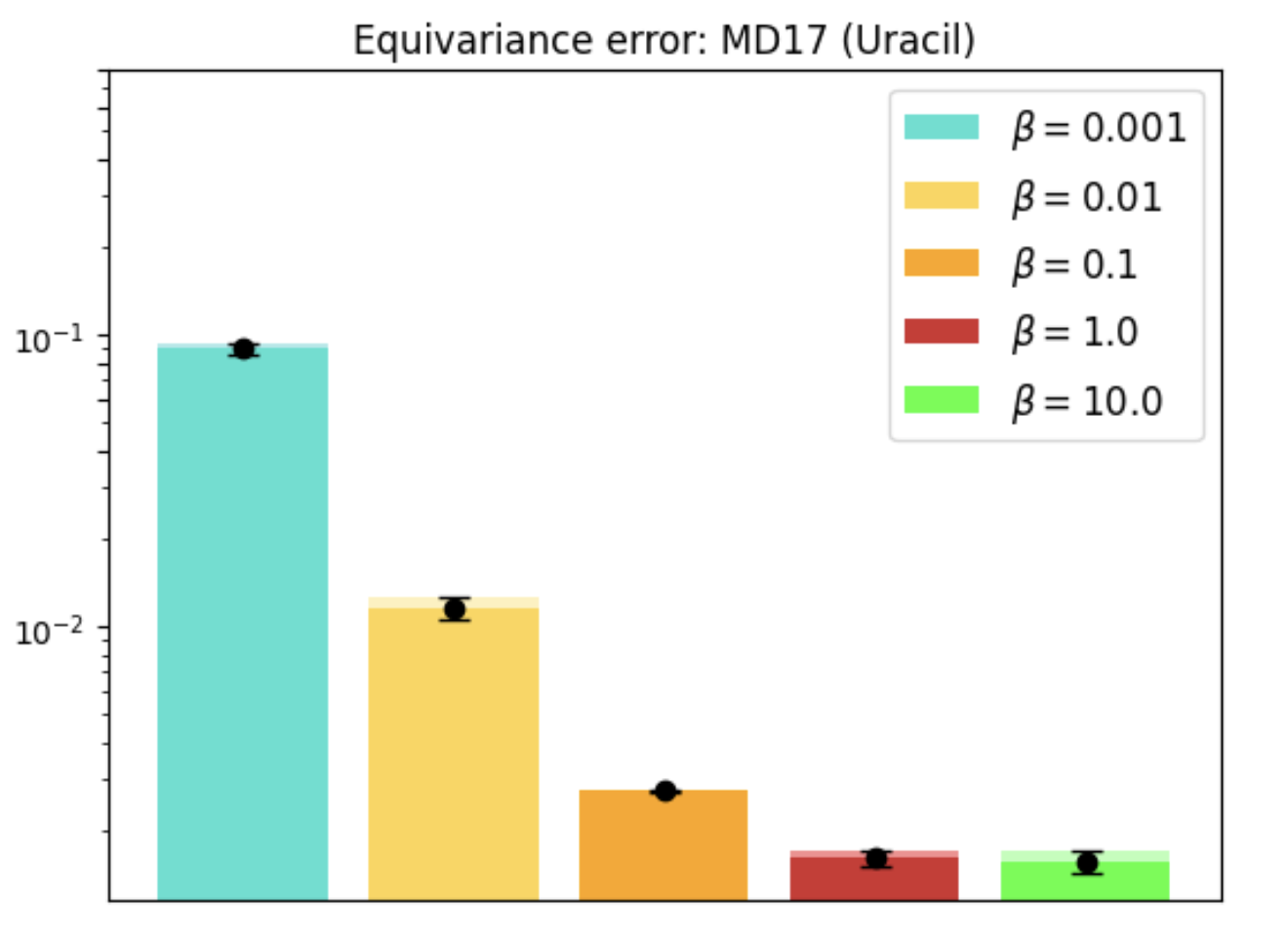}
    % \caption{m2_uracil}
    % \label{fig:m2_uracil}
}

\caption{MD17 dataset: GNN trained with REMUL. The second equivariance measure $E'$.
}
\label{fig:Additional MD17 second measure}
\end{figure}

In the main paper (Section~\ref{sec: Molecular Dynamics} and Table~\ref{tab:performance:MD17}), we present the performance of \methodabbr{} applied to GNN architecture on the MD17 dataset. To provide more insights into how \methodabbr{} behaves across different molecular structures and how the equivariance penalty $\beta$ affects task performance and equivariance error, we illustrate these relationships in Figures~\ref{fig:md17}--\ref{fig:Additional MD17 second measure}. For each molecule in the MD17 dataset, we trained a standard GNN using \methodabbr{} procedure with varying values of $\beta$. All experiments use the same training settings detailed in Appendix~\ref{appendix: Molecular Dynamics}.

\label{Additional Experiments: Molecular Dynamics}
\begin{figure}[ht]
    \centering
    % First row of images
    \subfigure[Aspirin (2D)]{
    \label{fig:aspirin_2d}
        \includegraphics[width=0.2\textwidth]{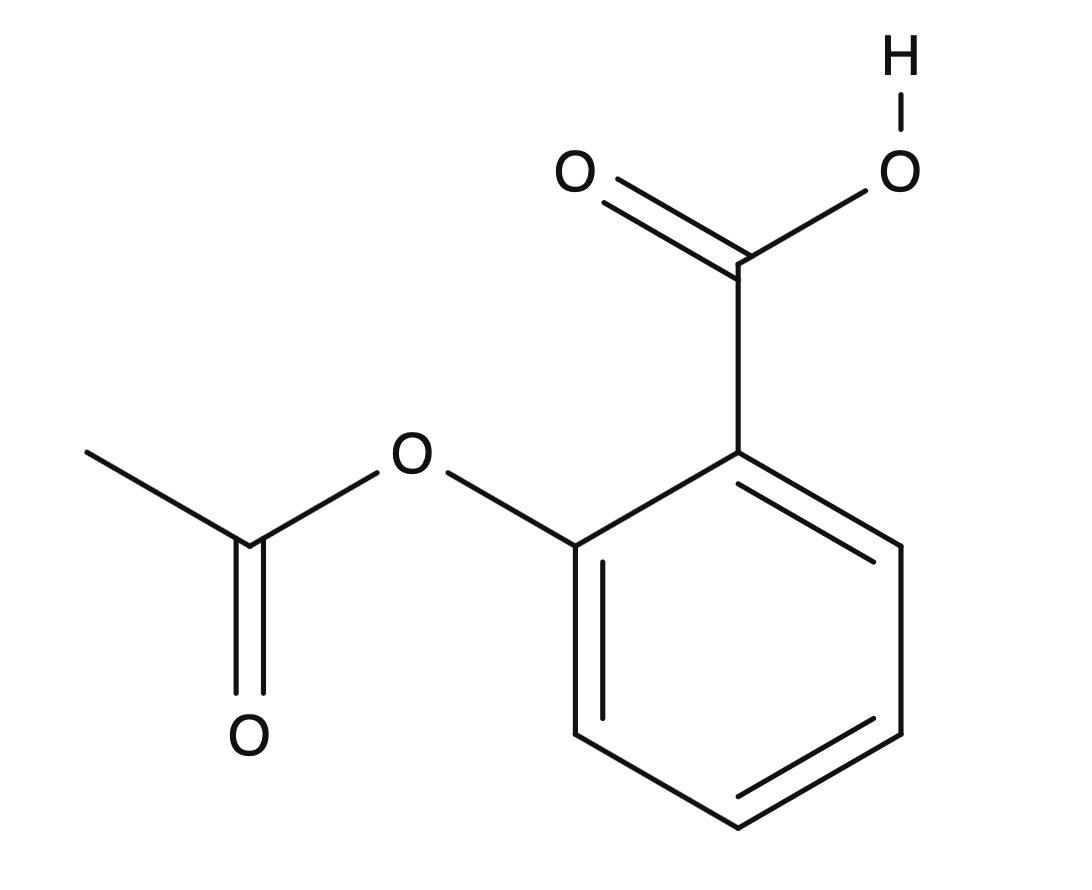}
        }
    \hspace{-0.01cm}
    \subfigure[Aspirin (3D)]{
    \label{fig:aspirin_3d}
        \includegraphics[width=0.2\textwidth]{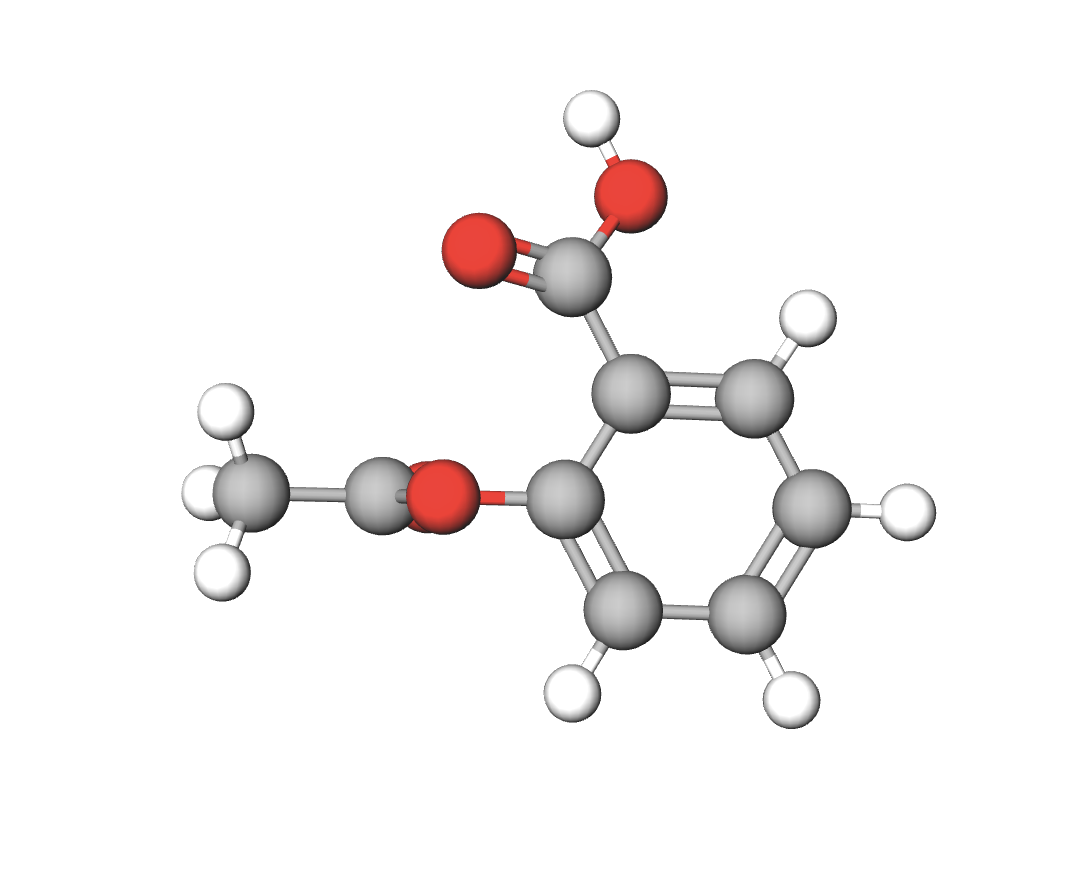}
        }
    \hspace{-0.01cm}
    \subfigure[Ethanol (2D)]{
    \label{fig:ethanol_2d}
        \includegraphics[width=0.2\textwidth]{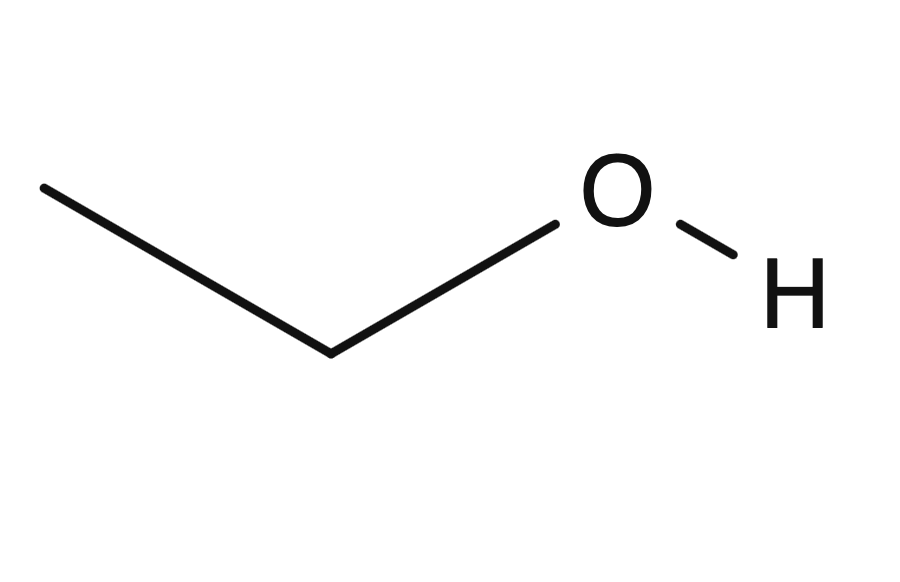}
        }
    \hspace{-0.01cm}
    \subfigure[Ethanol (3D)]{
     \label{fig:ethanol_3d}
        \includegraphics[width=0.2\textwidth]{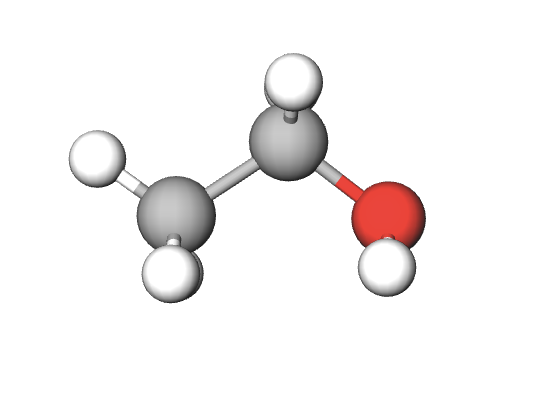}
    }
    % Second row of images
    \subfigure[Benzene (2D)]{
    \label{fig:benzene_2d}
        \includegraphics[width=0.2\textwidth]{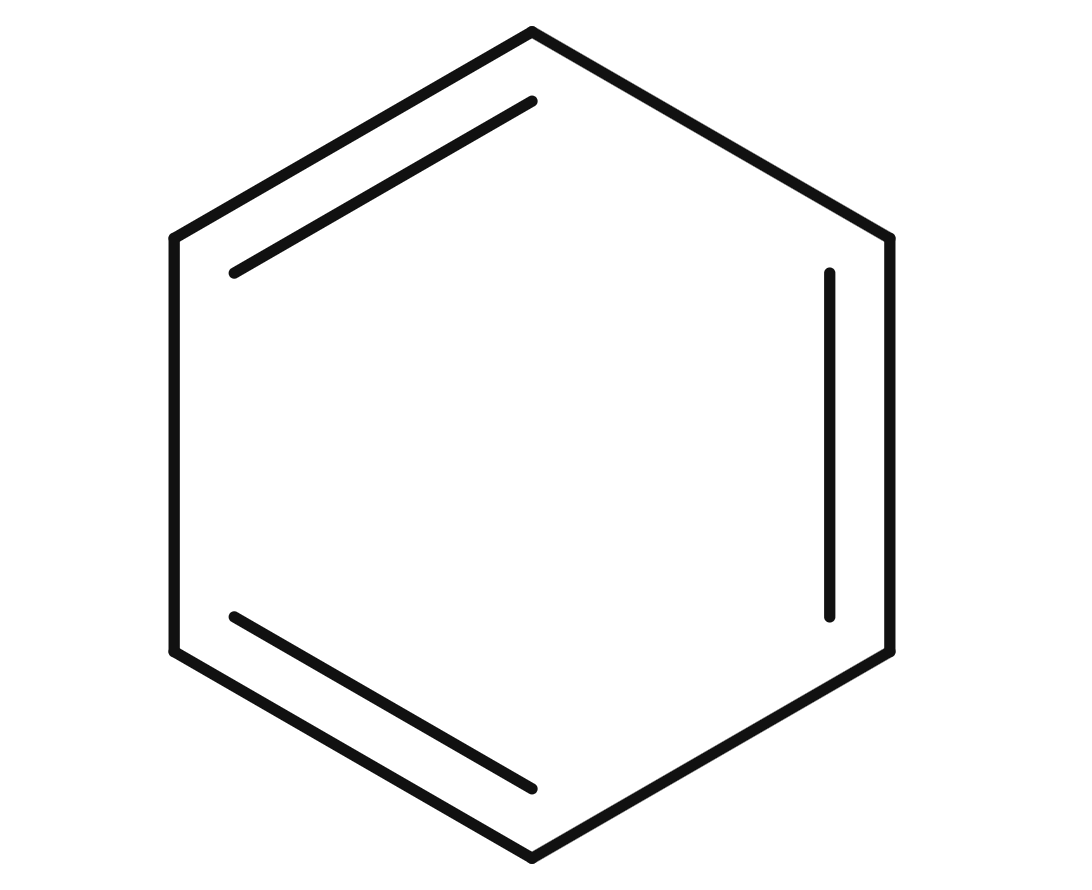}
    }
    \hspace{-0.01cm}
    \subfigure[Benzene (3D)]{
    \label{fig:benzene_3d}
        \includegraphics[width=0.2\textwidth]{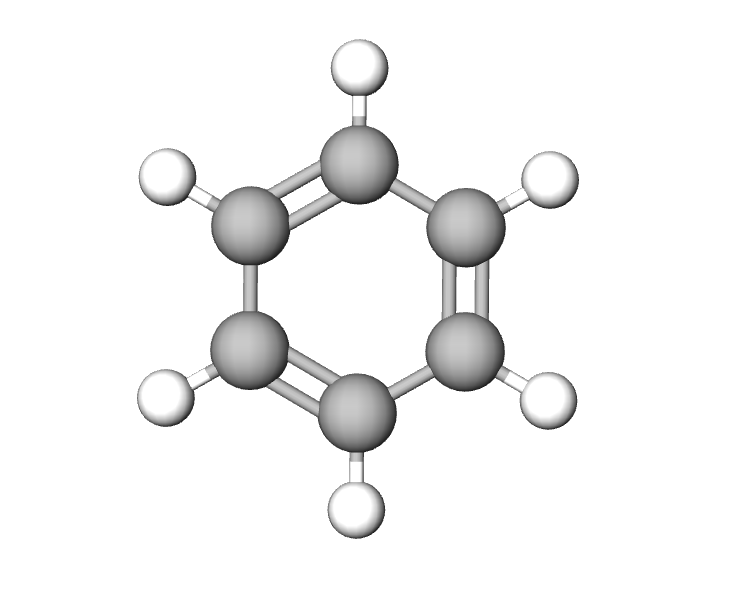}
    }
    \hspace{-0.01cm}
    \subfigure[Malonaldehyde (2D)]{
     \label{fig:Malonaldehyde_2d}
        \includegraphics[width=0.2\textwidth]{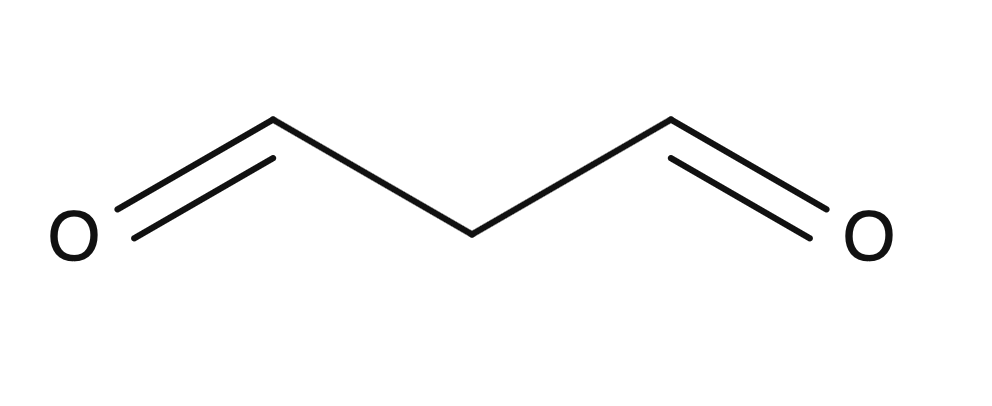}
    }
    \hspace{-0.01cm}
    \subfigure[Malonaldehyde (3D)]{
    \label{fig:Malonaldehyde_3d}
        \includegraphics[width=0.2\textwidth]{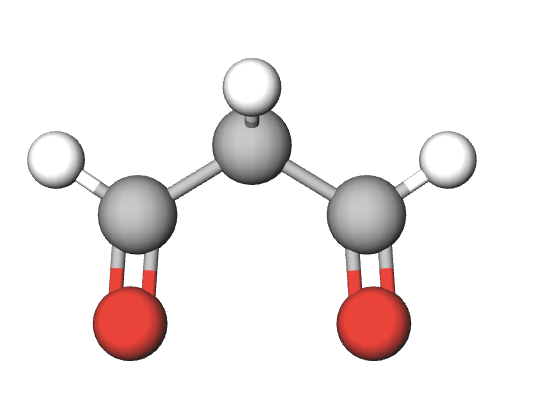}
    }
    % Third row of images
    \subfigure[Naphthalene (2D)]{
        \label{fig:Naphthalene_2d}
        \includegraphics[width=0.2\textwidth]{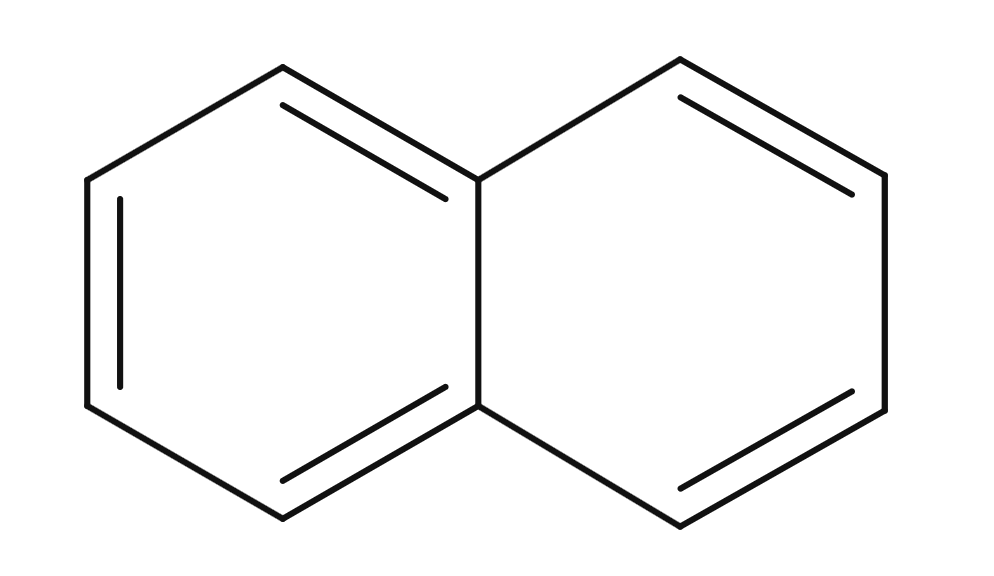}
    }
    \hspace{-0.01cm}
    \subfigure[Naphthalene (3D)]{
     \label{fig:Naphthalene_3d}
        \includegraphics[width=0.2\textwidth]{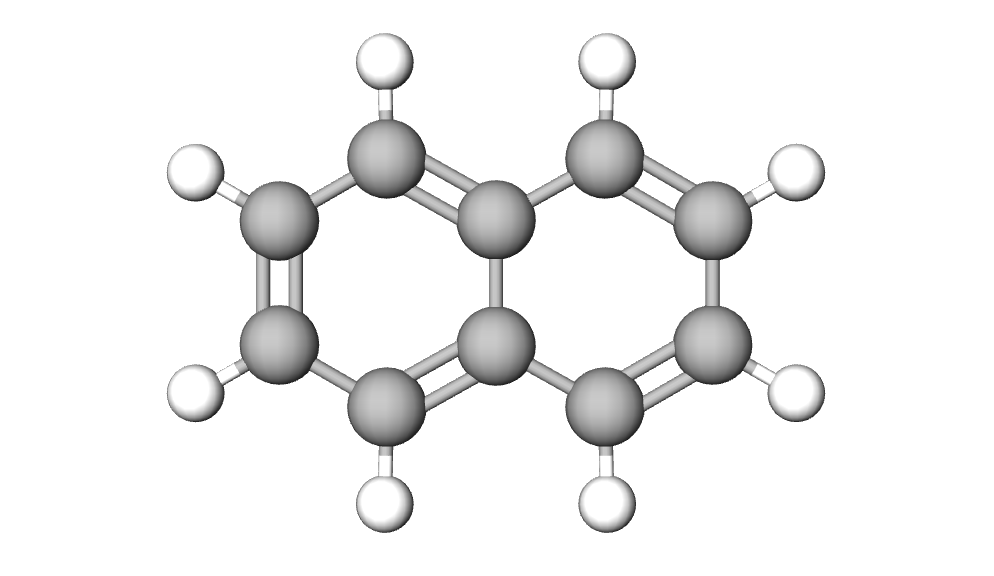}
    }
    \hspace{-0.01cm}
    \subfigure[Salicylic (2D)]{
    \label{fig:Salicylic_2d}
        \includegraphics[width=0.2\textwidth]{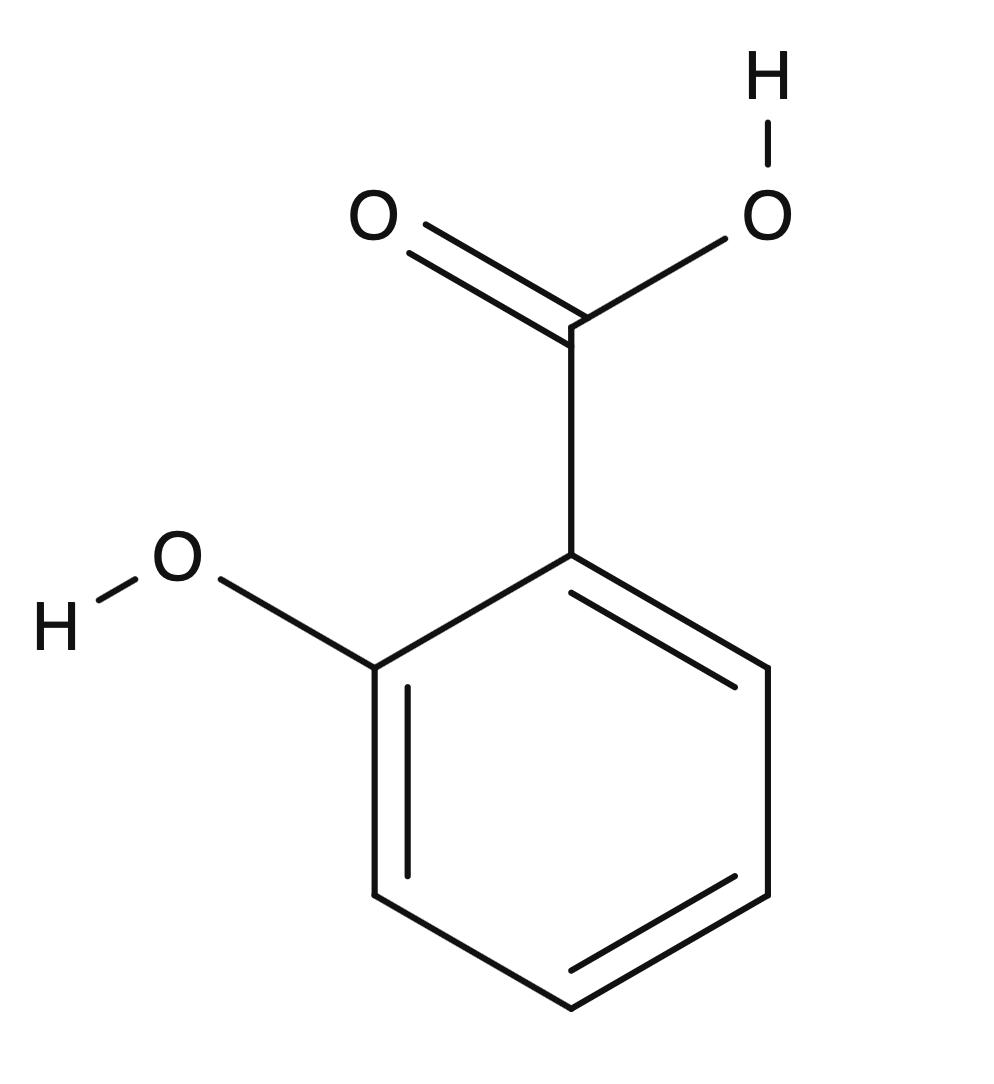}
    }
    \hspace{-0.01cm}
    \subfigure[Salicylic (3D)]{
        \label{fig:Salicylic_3d}
        \includegraphics[width=0.2\textwidth]{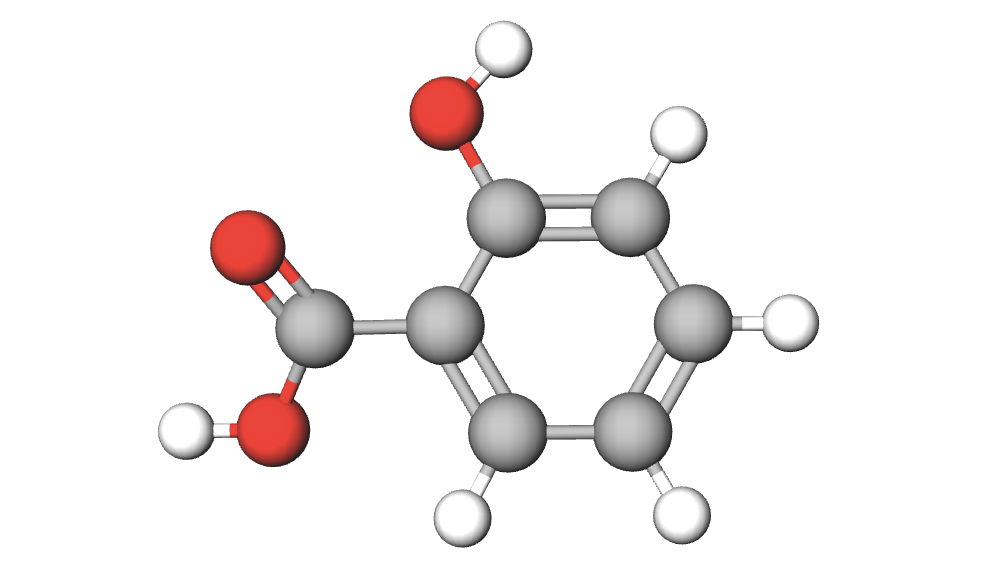}
    }
    % Fourth row of images
    \subfigure[Toluene (2D)]{
    \label{fig:Toluene_2d}
        \includegraphics[width=0.2\textwidth]{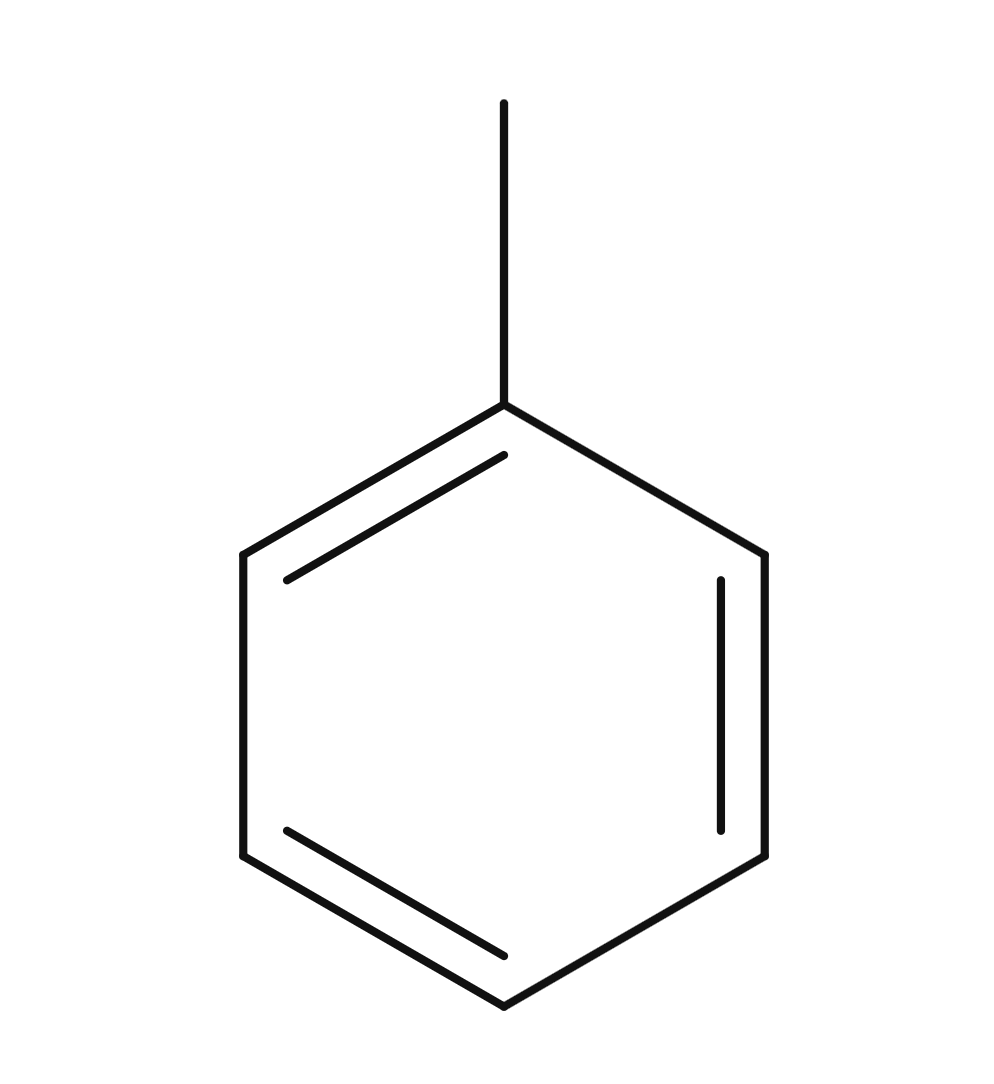}
    }
    \hspace{-0.01cm}
    \subfigure[Toluene (3D)]{
     \label{fig:figure14}
        \includegraphics[width=0.2\textwidth]{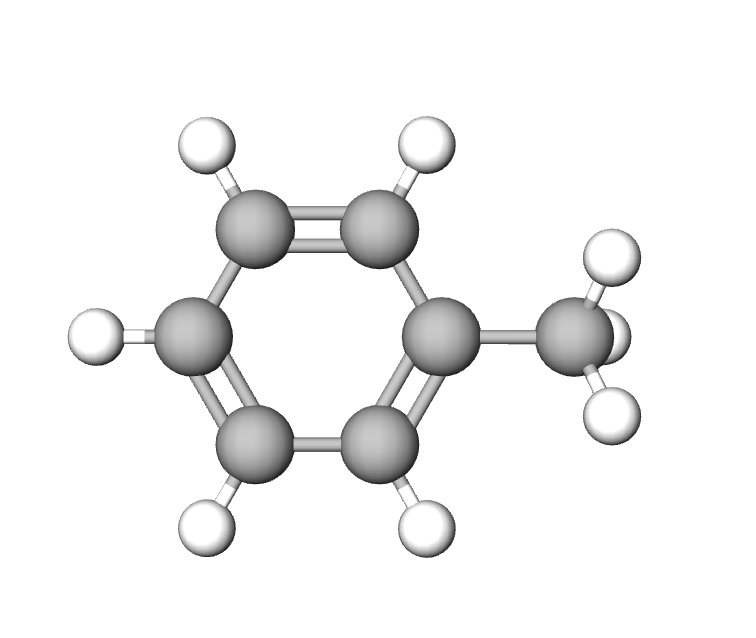}
    }
    \hspace{-0.01cm}
    \subfigure[Uracil (2D)]{
    \label{fig:Uracil_2d}
        \includegraphics[width=0.2\textwidth]{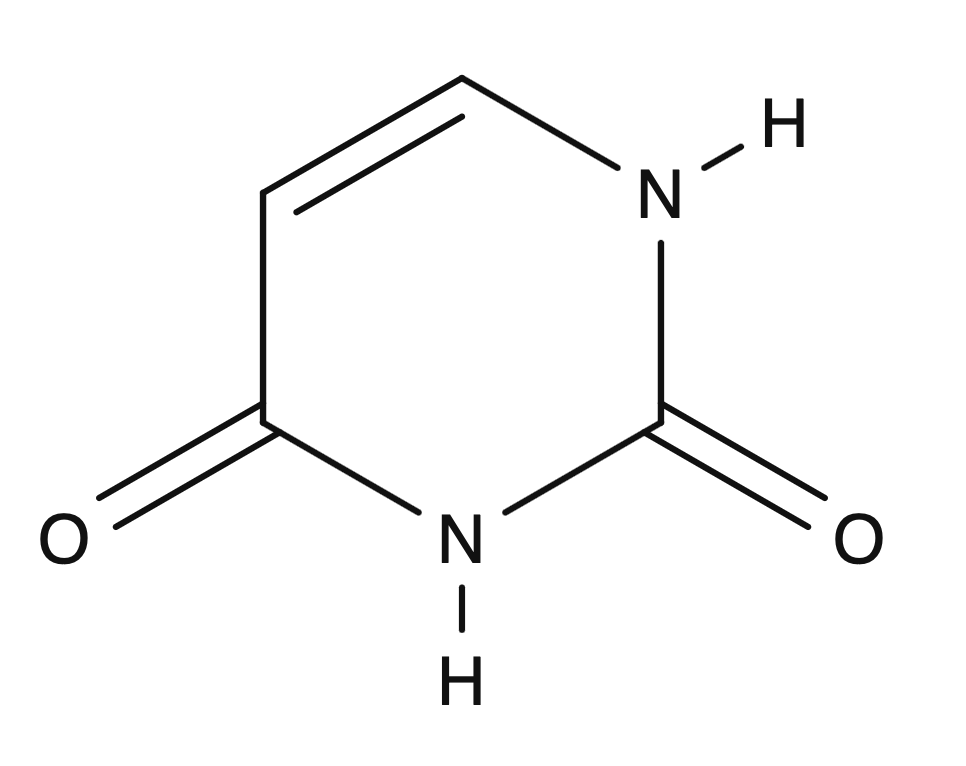}
    }
    \hspace{-0.01cm}
    \subfigure[Uracil (3D)]{
    \label{fig:Uracil_3d}
        \includegraphics[width=0.2\textwidth]{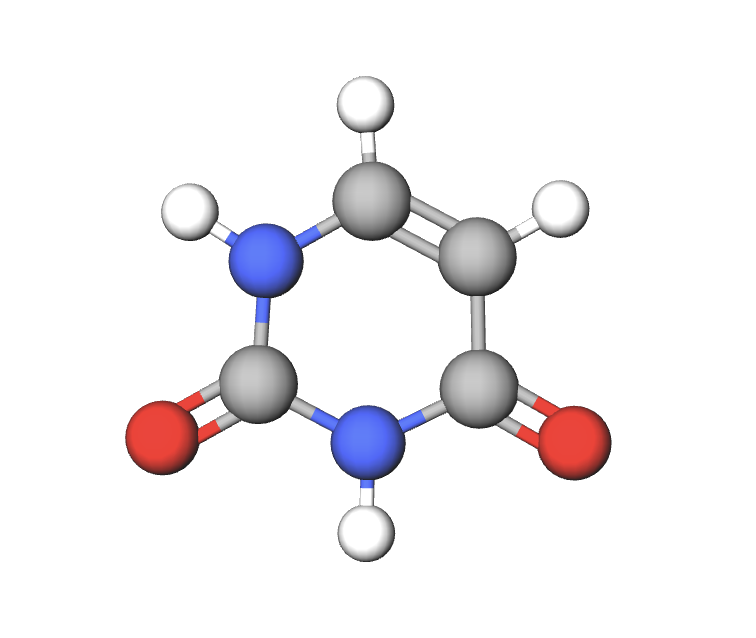}
    }
    \caption{MD17 molecules structures.}
    \label{fig:MD17 molecules structures}
\end{figure}

% \newpage
\clearpage 

\end{document}